\documentclass[sn-basic]{sn-jnl}% Basic Springer Nature Reference Style/Chemistry Reference Style
\usepackage{amsmath,amssymb,amsfonts,pifont,bbding}
\usepackage{graphicx}
\usepackage{subfig}
\usepackage{float}
\usepackage{color}
\usepackage{multirow}
\jyear{2023}%

\theoremstyle{thmstyleone}%
\theoremstyle{thmstyletwo}%

\theoremstyle{thmstylethree}%
\renewcommand{\tablename}{Table}
\newcommand{\figref}[1]{Fig.~\ref{#1}}
\newcommand{\tabref}[1]{Tab.~\ref{#1}}
\newcommand{\equref}[1]{Eq.~\ref{#1}}

\begin{document}

\title[ ]{Rethinking PRL: A Multiscale Progressively Residual Learning Network for Inverse Halftoning}

%%=============================================================%%
%% Prefix	-> \pfx{Dr}
%% GivenName	-> \fnm{Joergen W.}
%% Particle	-> \spfx{van der} -> surname prefix
%% FamilyName	-> \sur{Ploeg}
%% Suffix	-> \sfx{IV}
%% NatureName	-> \tanm{Poet Laureate} -> Title after name
%% Degrees	-> \dgr{MSc, PhD}
%% \author*[1,2]{\pfx{Dr} \fnm{Joergen W.} \spfx{van der} \sur{Ploeg} \sfx{IV} \tanm{Poet Laureate} 
%%                 \dgr{MSc, PhD}}\email{iauthor@gmail.com}
%%=============================================================%%

\author[1,2]{\fnm{Feiyu} \sur{Li}}\email{lifeiyu.cs@foxmail.com}
\author*[2]{\fnm{Jun} \sur{Yang}}\email{yangj95@mail2.sysu.edu.cn}

\affil[1]{\orgname{Zhejiang Sci-Tech University}, \orgaddress{\city{Hangzhou}, \country{China}}}
\affil[2]{\orgname{Jiaxing University}, \orgaddress{\city{Jiaxing}, \country{China}}}

%%==================================%%
%% sample for unstructured abstract %%
%%==================================%%

\abstract{Image inverse halftoning is a classic image restoration task, aiming to recover continuous-tone images from halftone images with only bilevel pixels. Because the halftone images lose much of the original image content, inverse halftoning is a classic ill-problem. Although existing inverse halftoning algorithms achieve good performance, their results lose image details and features. Therefore, it is still a challenge to recover high-quality continuous-tone images. In this paper, we propose an end-to-end multiscale progressively residual learning network (MSPRL), which has a UNet architecture and takes multiscale input images. To make full use of different input image information, we design a shallow feature extraction module to capture similar features between images of different scales. We systematically study the performance of different methods and compare them with our proposed method. In addition, we employ different training strategies to optimize the model, which is important for optimizing the training process and improving performance. Extensive experiments demonstrate that our MSPRL model obtains considerable performance gains in detail restoration.}


%%\pacs[JEL Classification]{D8, H51}

%%\pacs[MSC Classification]{35A01, 65L10, 65L12, 65L20, 65L70}

\maketitle

\section{Introduction}
\begin{figure}[th]
% \vspace{0.5em}
\begin{center}
    \begin{minipage}{0.3\linewidth}
        \subfloat[]{\includegraphics[width=1\linewidth]{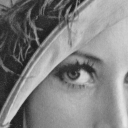}}
    \end{minipage}
    \begin{minipage}{0.3\linewidth}
        \subfloat[]{\includegraphics[width=1\linewidth]{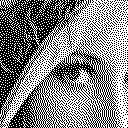}}
    \end{minipage}
    \begin{minipage}{0.3\linewidth}
        \subfloat[]{\includegraphics[width=1\linewidth]{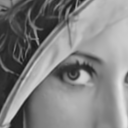}}
    \end{minipage}
    \\ \vspace{0.5em}
\caption{Examples of halftone and inverse halftoning images of the Lena image: original grayscale image (a), error diffusion halftone image (b) from (a) and MSPRL inverse halftone image (c) from (b).}
\label{fig:1}
\end{center}
% \vspace{-1em}
\end{figure}

The halftoning method represents continuous-tone images with two levels of color, namely black and white, due to cost considerations and is commonly used in digital image printing, publishing and displaying applications \citep{mulligan1992principled}. There are various methods for halftoning algorithms, such as error diffusion, dot diffusion, ordered dithering and direct binary search \citep{floyd1976adaptive, eschbach1991error, knuth1987digital, bayer1973optimum, seldowitz1987synthesis}. Because the halftone image has only two values, it can save considerable storage space and network transfer bandwidth compared to continuous-tone images. It is also a feasible and important image compression method. \figref{fig:1} illustrates an original grayscale image, corresponding to the halftone images and the inverse halftoning images.

Image inverse halftoning is an image restoration task that reconstructs continuous-tone images with 255 or more levels from its corresponding halftone images. The purpose is to convert a binary image at $\{0, 1\}^{H \times W}$ into a continuous image in $\mathbb{R}^{H \times W}$ space, where $H$ and $W$ are the image height and width respectively. Because the halftone image loses many detailed features during the halftoning process, it is a challenging and ill-posed problem. In the past several decades numerous image inverse halftoning approaches have been explored to achieve good inverse halftoning performance \citep{kite2000fast, analoui1992new, mese2001look, liu2010inverse, wong1995inverse}.

Owing to the success of deep convolutional neural networks (CNNs) in vision tasks, CNN-based image restoration methods have been extensively studied and have shown amazing performance. Several inverse halftoning methods based on deep learning have also achieved the significant advancements \citep{hou2017image, xiao2017deep, yuan2019gradient, xia2018deep}. These methods mainly use the typical UNet architecture to build their CNN models. The UNet architecture is a multilevel design that aims to recover detailed features by extracting different information at multiple scales of the image. Therefore, it is widely used as a baseline in many vision models.

However, there is still a certain gap in detail restoration. Although most of the existing methods use the UNet architecture, they cannot effectively extract features at different image scales, because of which the quality of image reconstruction still has much room for improvement. Previous studies \citep{hou2017image, xiao2017deep, yuan2019gradient} did not effectively extract image textures and features of multilayer downsampled images, and failed to restore high-quality continuous-tone images. In addition, \cite{shao2021efficient} added an attention mechanism to enhance the detail extraction, but, despite the increased complexity of the model, there was no obvious performance improvement.

In this paper, we present a novel multiscale progressively learning network architecture that inspired by precious progressively learning UNet architectures \citep{zamir2022restormer, chen2022simple, cho2021rethinking}. Our model takes multiscale input images and uses a shallow feature extraction module to extract similar features from multiscale images. The encoder and decoder are composed of multiple residual block modules. Then, the feature fusion module fuses the output images of different stages of the encoder as the input images of the decoder, and finally outputs the continuous-tone images via progressively learning, which can ensure the efficiency of the model's learning ability. We conduct experiments on the VOC2012 dataset, which is widely used in other vision tasks, such as image classification, object detection and instance segmentation. The main contributions of this paper are as follows.

1) Our MSPRL contains encoder and decoder stages. The encoder is mainly responsible for restoring image information and removing noise that affects image quality. The decoder aims to recover the texture details of different feature maps from the encoding stage, and outputs continuous-tone grayscale images. Meanwhile, we compare some common feature extraction blocks in the encoder and decoder.

2) We propose a computationally inexpensive shallow feature extraction module (SFE) to extract attention information between images to recover content feature representation and use a feature fusion module (FF) to fuse different stages of feature information.

3) Compared with other researchers who are keen on designing model architectures, we delve into the optimization of training strategies. Good training strategies for the performance improvement of the model are obvious and are used in our model training process, such as data augmentation methods and compound loss functions, which bring considerable improvements to model training and optimization.

\section{Related Work}

\subsection{Conventional Inverse Halftoning}
During the last decade, many approaches have been proposed for image inverse halftoning. Simple approaches use low-pass filtering to remove halftone noise \citep{wong1995inverse, catte1992image}. Although these methods can remove most of the halftone noise, they also remove high-frequency edge information. Thus, \cite{kite2000fast} proposed gradient-based spatially varying filtering for error diffused images to better recover high-frequency details of images. \cite{unal2001restoration}, and \cite{analoui1992new} proposed projection onto the convex sets method (POCS) for inverse halftoning. In addition, some researchers used wavelet-based methods to separate halftoned image noises and then reconstruct the original image by wavelet shrinkage. Based on the Bayesian approach, \cite{liu2010inverse} built a correlation map between adjacent points for inverse halftoning. Dictionary-based learning has also been widely and successfully applied to inverse halftoning \citep{zhang2018sparsity}. \cite{son2014local} proposed an edge-oriented local learned dictionaries (LLD) method to enhance the edge details of the restored image. Considering computational efficiency, Mese and Guo further proposed a precomputed look-up table (LUT) \citep{mese2001look, guo2013efficient} to improve performance and utilize efficiency. \cite{huang2008neural} used a hybrid neural network method to process halftone and inverse halftoning images.

\subsection{Deep Convolutional Neural Networks}
Deep convolutional neural networks (CNNs) have become the dominant method for solving various image reconstruction problems and have achieved state-of-the-art performance on a wide variety of vision datasets. SRCNN \citep{dong2014learning} first introduced CNNs to the image super-resolution (SR) task, which focuses on reconstructing high-resolution (HR) details from corresponding low-resolution (LR) images, and obtains superior performance against previous conventional SR methods. ResNet \citep{he2016deep} introduced an identity skip connection that alleviates the difficulty of model degradation in deep neural networks and allows networks to learn deeper feature representations. VDSR \citep{kim2016accurate} achieved a good recovery effect using a residual learning architecture for super resolution. EDSR \citep{lim2017enhanced} built a very wide network using residual blocks. DnCNN \citep{zhang2017beyond} used CNNs to remove the white Gaussian noise of images. MIMO-UNet \citep{cho2021rethinking}, NAFNet \citep{chen2022simple} and Restormer \citep{zamir2022restormer} presented the multi-input fusion UNet architecture to aggregate multiscale feature information for image deblurring and image deraining.

Image inverse halftoning is similar to many image restoration tasks. Thus, \cite{hou2017image}, and \cite{xiao2017deep} applied CNNs to inverse halftoning by building a UNet network as the restoration network. \cite{xia2018deep} proposed a progressively residual learning network (PRL) including two main stages: the content aggregation stage, which restores the content map, and the detail enhancement stage, which restores the extract texture and details. \cite{yuan2019gradient} proposed gradient-guided residual learning CNNs (GGRL) for inverse halftoning. The same subnetworks are used to learn gradient maps of different Sobel orientations from the input halftone image, and a coarse map is output that is used to restore the continuous-tone images. \cite{shao2021efficient} presented an attention model for inverse halftoning by using residual channel attention blocks (RCAB) \citep{zhang2018image}. \cite{xia2021deep} and \cite{yen2021inverse} combined inverse halftoning with image colorization methods to recover color continuous-tone images with better visual quality from corresponding halftone grayscale images.

\subsection{The importance of training strategies}
Better training strategies can increase the performance of a model and effectively decrease the training time \citep{goyal2017accurate, he2019bag, qian2022pointnext, lin2022revisiting}. Data augmentation is one of the most important strategies to boost the performance of a neural network \citep{cubuk2020randaugment}. It can provide more learning samples and improve model generalization through various random changes for training images. Many researchers use cosine annealing \citep{loshchilov2016sgdr} decay to boost performance. Furthermore, the warm-up method \citep{goyal2017accurate, he2019bag} is used to alleviate the instability of the model in the early training stage. In many vision tasks, removing batch normalization (BN) layers can increase performance and reduce computational complexity such as SR and deblurring \citep{lim2017enhanced, wang2018esrgan}. \cite{zhao2016loss} showed that L$_1$ loss has a better convergence effect and image perceptual quality than L$_2$ loss. In this paper, we adopt suitable training strategies for our inverse halftoning task to improve the visual quality of restored continuous-tone images.

\begin{figure*}[ht]
\begin{center}
    \begin{minipage}{1\linewidth}
        \includegraphics[width=1\linewidth]{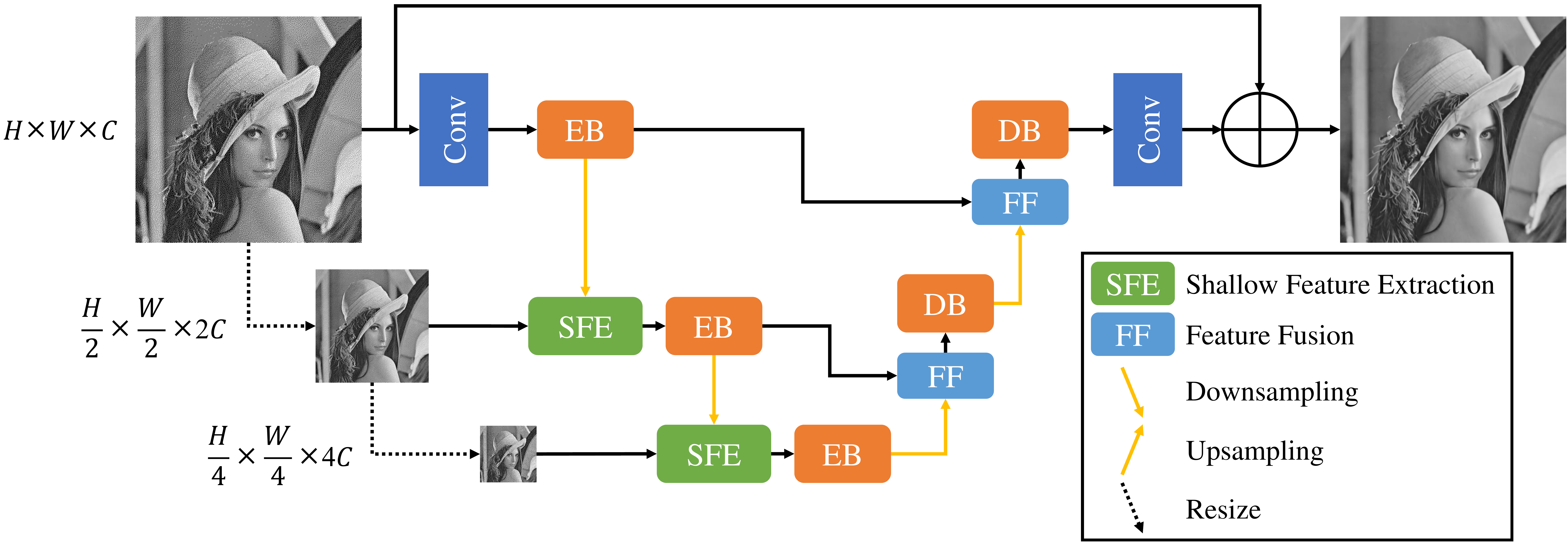}
    \end{minipage}
    \\ \vspace{0.5em}
\caption{Architecture of MSPRL for inverse halftoning. Our MSPRL consists of multiscale stage and different modules. The core of MSPRL is encoder block (EB) and decoder block (DB), which consist of residual block groups. The Conv is a 3$\times$3 convolution layer. The shallow feature extraction module (SFE) extracts multiscale features and the feature fusion module (FF) fuses different features from the encoders. H and W represent the height and width of the image, and C represents the number of feature map channels.}
\label{fig:2}
\end{center}
% \vspace{-1em}
\end{figure*}

\section{Methodology}
In this section, we first introduce our MSPRL model based on the UNet architecture and propose the shallow feature extraction module (SFE) in Sec. \ref{Architecture}. The overall architecture of MSPRL is shown in \figref{fig:2}. Then we describe our loss function in Sec. \ref{Loss}. Last, we use an overall different training strategy compared with PRL in Sec. \ref{Strategies}.

\subsection{model architecture}\label{Architecture}
As shown in \figref{fig:2}, given a halftone input image $X \in\{0, 1\}^{H \times W \times 1}$, the goal of our method is to restore the clear and continuous-tone grayscale image $Y \in\mathbb{R}^{H \times W \times 1}$ by progressively learning. Our model is mainly divided into two stages: the left encoder (EC) stage and the right decoder (DC) stage, and three levels from top to bottom.

\noindent\textbf{Overall Pipeline.} In the encoder stage, we first use a 3$\times$3 convolution layer to obtain a low-level feature map F$_{k}^{EB}\in\mathbb{R}^{H \times W \times C}$, where $H$$\times$$W$ denotes the spatial dimension, $C$ is the number of feature map channels that we set to 48, $k$ represents the $k^{th}$ level and EB represents encoder block that consist of 8 residual blocks (RBs). Then, F$_k^{EB}$ passes through an encoder EB$_k$ which transforms it into deep feature maps at level 1. EB$_k$ obtains an output EB$_k^{down}$ by downsampling, where the number of channels is doubled and the image size is halved. For the downsampling and upsampling modules, we apply pixel-unshuffle and pixel-shuffle operations respectively. To extract the similar information of multiscale images, we use the shallow feature extraction module (SFE) to exploit the attention features of EB$_k^{down}$ and X$_{k-1}^{resize}$ in the second and third levels, respectively, and output the fusion attention feature maps of SFE$_k$, where $resize$ uses linear interpolation downsampling from the corresponding $k$-1 level of the input image $X$. Then SFE$_k$ passes through EB to obtain deep features. The left encoding stage process is defined as:
\begin{equation}
    EB_k=\left\{
    \begin{array}{ll}
     EB_k(Conv_3(X_k)) & k=1,\\
     EB_k(SFE_k(EB{_{k=1}})) & k=2, 3,
    \end{array}
    \right.
    \label{eq:1}
\end{equation}
where $X$ is the input image, $Conv_3$ represents a 3$\times$3 convolutional layer, and EB$_k$ and SFE$_k$ represent the outputs of the $k^{th}$ level EB and SFE respectively.

In MSPRL, the decoder takes the encoder features EB as input and progressively recovers the continuous-tone representations. First, the feature fusion module (FF) aggregates the feature maps of different encoder stages EB$_k$ and EB$_{k+1}^{up}$ and outputs the aggregated features FF$_k$. Then, we use the decoder block (DB) DB$_k$ to reconstruct the image details, where DB is also composed of 8 residual blocks (RBs). Through a series of decoding and reconstruction, we obtain F$_k^{DB}$. Finally, we apply a 3$\times$3 convolution and residual connection to obtain the final continuous-tone image $Y$. The overall process is progressively learning. The right decoding stage process is defined as:
\begin{equation}
    DB_k=DB_k(FF_k(EB_k,EB_{k+1}^{up})), \label{eq:2}
\end{equation}
\begin{equation}
    Y=Conv_3(DB_1)+X, \label{eq:3}
\end{equation}
where $X$ and $Y$ are the input and output images, respectively, $Conv_3$ represents a 3$\times$3 convolutional layer, and DB$_k$ and FF$_k$ represent the outputs of the $k^{th}$ DB and FF respectively, and $k$ = 1,2 in \equref{eq:2}.

\begin{figure}[ht]
\begin{center}
    \begin{minipage}{1\linewidth}
        \begin{center}
        \subfloat[]{\includegraphics[width=0.6\linewidth]{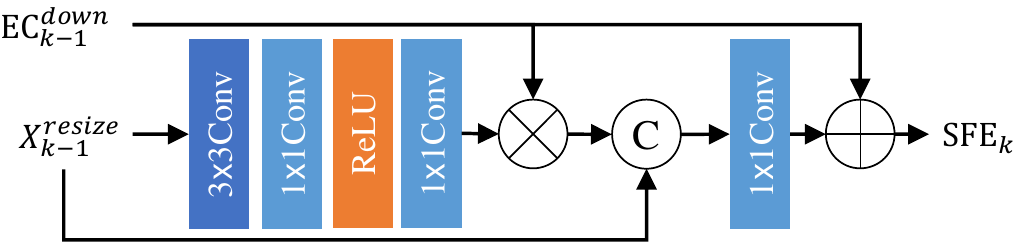}\label{fig:3_1}}
        \hspace{1.2em}
        \subfloat[]{\includegraphics[width=0.24\linewidth]{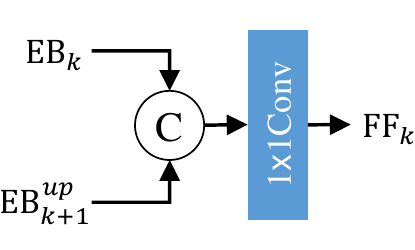}\label{fig:3_2}}
        \end{center}
    \end{minipage}
    \\ \vspace{0.5em}
\caption{The structures of submodules: (a) SFE and (b) FF.}
\label{fig:3}
\end{center}
% \vspace{-1.0em}
\end{figure}

\noindent\textbf{Shallow Feature Extraction and Feature Fusion.} Inspired by the shallow convolutional module (SCM) in MIMO-UNet \citep{cho2021rethinking}, our shallow feature extraction module (SFE) is shown in \figref{fig:3}\subref{fig:3_1}. The $X_{k-1}^{resize}$ passes through a 3$\times$3 convolutional layer and two stacks of 1$\times$1 point-wise convolution to output a low-level feature map Conv$_k^{stack}$. Then, we use element-wise multiplication to obtain attention features between the previous Conv$_k^{stack}$ and EB$_{k-1}^{down}$. A 1$\times$1 point-wise convolution is used to aggregate the attention features of previous $X_{k-1}^{resize}$ and EB$_{k-1}^{down}$, which are shown in \figref{fig:3}\subref{fig:3_2}. The SFE is formulated as:
\begin{equation}
     SFE_k^{att}=Conv_k^{stack}(X_{k-1}^{resize}) \otimes EC_{k-1}^{down}, \label{eq:4}
\end{equation}
\begin{equation}
     SFE_k=Conv_1(Concat(X_{k-1}^{resize}, SFE_k^{att}))+EC_{k-1}^{down}, \label{eq:5}
\end{equation}
where $k$ = 2,3 represents the $k^{th}$ level, $Conv_{stack}$, $Conv_1$ and $\otimes$ represent multiple stacked convolutional layers, a 1$\times$1 convolutional layer and element-wise multiplication respectively.

For the feature fusion module (FF), the FF aggregates the feature maps of EB$_k$ and EB$_{k+1}^{up}$ and is formulated as:
\begin{equation}
     FF_k=Conv_1(Concat(EB_{k}, EB_{k+1}^{up})), \label{eq:6}
\end{equation}
where $k$ = 1,2 represents $k^{th}$ level and $Conv_1$ represents a 1$\times$1 convolutional layer.

\noindent\textbf{Downsampling and Upsampling.} We use PixelShuffle as the upsampling and downsampling operation. Compared with convolution upsampling and downsampling, PixelShuffle can obtain better visual quality \citep{shi2016real}.

\begin{figure}[ht]
\begin{center}
    \begin{minipage}{0.3\linewidth}
        \subfloat[]{\includegraphics[width=1\linewidth]{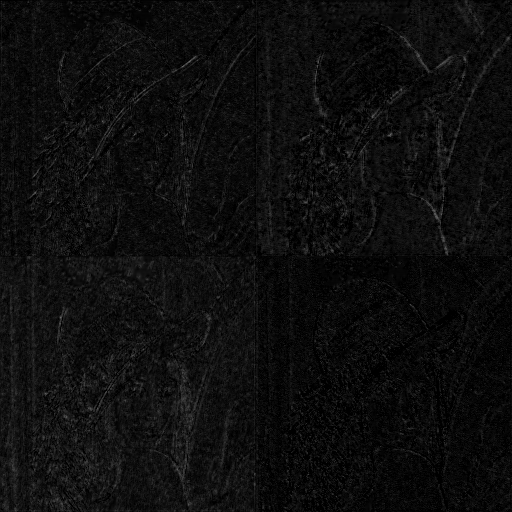}}
    \end{minipage}
    \begin{minipage}{0.3\linewidth}
        \subfloat[]{\includegraphics[width=1\linewidth]{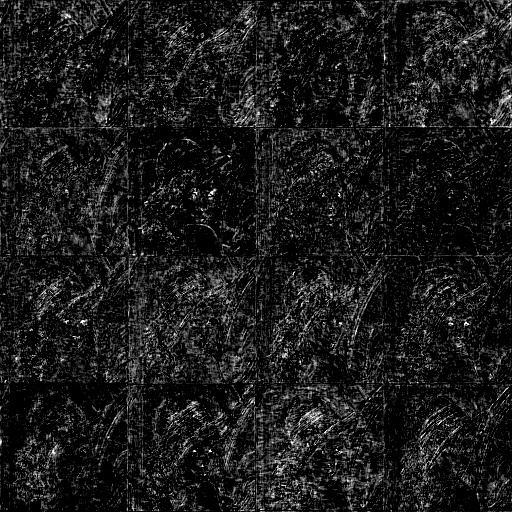}}
    \end{minipage}
    \begin{minipage}{0.3\linewidth}
        \subfloat[]{\includegraphics[width=1\linewidth]{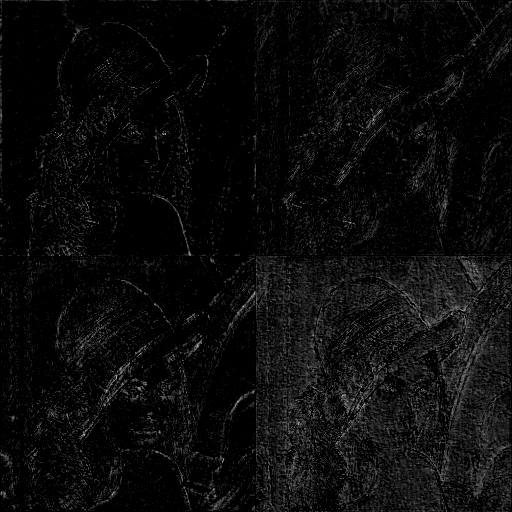}}
    \end{minipage}
    \\ \vspace{0.5em}
\caption{Examples of the feature map images of the Lena image in different stages: feature maps of 7th layer EB$_{k=2}$ (a), and the corresponding DB$_{k=2}$ feature maps (c), and EB$_{k=3}$ feature maps (b). The encoder's feature maps tend to recover image content, and the decoder's feature maps focus on detail extraction.}
\label{fig:5}
\end{center}
\vspace{-1.5em}
\end{figure}

\noindent\textbf{Progressively Learning.} Progressively learning allows the network to learn local and global features, which makes full use of semantic information of different scale images. In addition, it can also greatly reduce the convolution operation time under the small image patches. The feature maps of different stages are shown in \figref{fig:5}.

\subsection{Loss Function}\label{Loss}
Although L$_1$ loss, MSE loss and perceptual loss are used in PRL, we experimentally found that perceptual loss is added with a very large penalty coefficient, which has little effect on model convergence, and that MSE loss has a smooth ing effect. In this paper, we only use L$_1$ loss as follows:
\begin{equation}
    L_{pixel}={\|X_{gt}-Y\|}_1,\label{eq:7}
\end{equation}
where $L_{pixel}$ is the pixel-wise loss that evaluates the L$_1$ distance between recovered image $Y$ and the ground-truth gray image $X_{gt}$. Some studies have shown that composite loss functions can improve performance. Inspired by \citep{cho2021rethinking}, we add the fast Fourier transform (FFT) \citep{cochran1967fast} loss function to strengthen the high-frequency extraction as follows:
\begin{equation}
    L_{FFT}={\|FFT(X_{gt})-FFT(Y)\|}_1,\label{eq:8}
\end{equation}
where FFT represents the fast Fourier transform that transfers the image signal to the frequency domain and uses L$_1$ loss to evaluate the distance between recovered image $Y$ and the ground-truth gray image $X_{gt}$. The final loss function for training our model is as follows:
\begin{equation}
    L_{total}=L_{pixel}+\lambda L_{FFT},\label{eq:9}
\end{equation}
where we set $\lambda$ = 0.1 in experiment.

\subsection{Training strategies}\label{Strategies}
We first show the different training strategy comparisons in \tabref{tab:5}. Then we illustrate the strategies that differ from PRL.

\noindent\textbf{Data augmentation.} We found that other researchers use the resize operation to scale the images to 256$\times$256. However, this resizing operation results in the loss of much of detail and texture information of the original image. During our training period, we use random cropping on the training data so that the model can learn image information of different regions. Data augmentation enables the model to learn richer feature representations and improves model generalization.

\noindent\textbf{Bigger batch-size.} The original PRL uses a minimum batch-size of 1. A smaller batch size will make the model training unstable and affect the convergence speed. We use the most commonly used batch-size 16.

\noindent\textbf{Optimizer and Schedule.} Unlike PRL, we utilize the AdamW optimizer \citep{loshchilov2017decoupled} instead of Adam \citep{kingma2014adam} and optimizer momentum with ($\beta_1=0.9,\beta_2=0.999$). For the learning rate decreasing strategy, we use the cosine annealing decay schedule \citep{loshchilov2016sgdr} instead of linear decay.

\section{Experiments}
In this section, we first describe the datasets, evaluation metrics and training details. We then show the impact of different training strategies in the same baseline PRL. We demonstrate the performance of the different models later.

\subsection{Datasets and Implementation Details}
\noindent\textbf{Datasets and Metrics.} Following the PRL, we use the VOC2012\footnote{\url{http://host.robots.ox.ac.uk/pascal/VOC/}.} dataset \citep{everingham2015pascal}, which includes over 17000 images. We randomly select 13841 images for training and 3000 nonoverlapping images for validation, where excluding images patch size smaller than 256$\times$256. We evaluate the model on the Place365\footnote{\url{http://places2.csail.mit.edu/}.} small test dataset \citep{zhou2017places}, In addition, some classic images, such as Lena, Barbara and Baboon, and the Kodak\footnote{\url{http://r0k.us/graphics/kodak/}.} dataset are added to the test dataset. We also test five standard SR benchmark datasets including Set5 \citep{bevilacqua2012low}, Set14 \citep{zeyde2010single}, BSD100 \citep{martin2001database}, Urban100 \citep{huang2015single} and Manga109 \citep{matsui2017sketch}, where some images are properly cropped to fit the original PRL \citep{xia2018deep} model. In the experiment, the halftone images for all datasets are generated by the Floyd Steinberg error diffusion algorithm \citep{floyd1976adaptive}. For the evaluation metrics, the peak signal to noise ratio (PSNR), and the structural similarity metric (SSIM) are used in all experiments. Our code and pre-trained models are available at \url{https://github.com/FeiyuLi-cs/MSPRL}.

\begin{table}[ht]
\footnotesize
\renewcommand\arraystretch{1.2}
\begin{center}
    \caption{Comparison of training strategies between PRL and MSPRL.}
    \label{tab:5}
    % \resizebox{\linewidth}{!}{
     \begin{tabular}{lcc}
     \toprule
     Training config & PRL & MSPRL \\
     \midrule
     System implement & TensorFlow & Pytorch \\ 
     Dataset size & 13K & 13K \\ 
     Data augment & \ding{55} & \ding{51} \\ 
     Batch size & 1 & 16 \\ 
     Image size & 256 & 128 \\ 
     Epochs & 150 & 347 \\ 
     Total iterations & 1950K & 300K \\ 
     Channel dimension & 64 & 48 \\ 
     Optimizer & Adam & AdamW \\ 
     Optimizer momentum & $\beta_1=0.9,\beta_2=0.999$ & $\beta_1=0.9,\beta_2=0.999$ \\ 
     Learning rate decay & $2e^{-4}\to2e^{-6}$ & $2e^{-4}\to1e^{-6}$ \\ 
     Learning rate schedule & Linearly decay & Cosine decay \\ 
     Loss function & L$_1$+MSE+Perceptual Loss & L$_1$+FFT Loss \\
     \botrule
    \end{tabular}
    % }
\end{center}
% \vspace{-1em}
\end{table}

\noindent\textbf{Training detail.} In the training process, the batch-size is set to 16, and then the sampled images are randomly cropped to 128$\times$128. For data augmentation, each image patch is horizontally flipped with a probability of 0.5. We use iterations instead of epochs to represent the training length. The model is trained by the AdamW optimizer \citep{loshchilov2017decoupled} ($\beta_1=0.9,\beta_2=0.999$) for 300K iterations. The initial learning rate is set to 0.0002, which gradually decays to $1e^{-6}$ with cosine annealing \citep{loshchilov2016sgdr}. The model training time is approximately 18 hours and runs on one Nvidia RTX 3090 GPU.

\subsection{Ablation study}
In this section, we conduct experiments to show the effects of different modules, activation functions and feature blocks with our method. Our MSPRL model employs 8 residual blocks for each encoder and decoder. First, we evaluate the effectiveness of MSPRL without SFE and FF. The experimental results are shown in \tabref{tab:2}. The FF improves PSNR by 0.02 dB compared with SFE in the Kodak datasets, and the performance gain is further increased to 0.05 dB when we combine FF with SFE. The results show that aggregating feature maps from different encoders is more important than computing attention feature maps for our model learning.

Many vision networks adopt ReLU \citep{nair2010rectified} or LeakyReLU \citep{maas2013rectifier} as the activation function. In recent years, GELU \citep{hendrycks2016gaussian} has gradually become the first choice. Therefore, we test three activation functions to explore the best performance for our method. The experimental results are shown in \tabref{tab:3}. The results show the effect of different activation functions on model performance. ReLU performs better overall on multiple datasets; the results of LeakyReLU and GELU are close to ReLU but add some training time. Thus, we choose ReLU as the activation function in our model.

\begin{table}[ht]
% \vspace{-1.0em}
\footnotesize
\begin{center}
    \caption{Ablation study of SFE and FF.}
    % \vspace{-0.5em}
    \label{tab:2}
    % \resizebox{\linewidth}{!}{
    \begin{tabular*}{0.62\textwidth}{@{\extracolsep{\fill}}ccccccc@{\extracolsep{\fill}}}
     \toprule
     \multirow{2}{*}{SFE} & \multirow{2}{*}{FF}
     & \multicolumn{2}{c}{Place365} & \multicolumn{2}{c}{Kodak}\\
     \cmidrule{3-4} \cmidrule{5-6}
     & & PSNR & SSIM & PSNR & SSIM \\
     \midrule
     \checkmark & & 30.76 & 0.9019 & 31.84 & 0.8897 \\
     & \checkmark & 30.76 & 0.9019 & 31.86 & 0.8897 \\
     \checkmark & \checkmark & 30.77 & 0.9020 & 31.89 & 0.8898 \\
     \botrule
    \end{tabular*}
    % }
\end{center}
\vspace{-2em}
\end{table}

\begin{table}[ht]
% \vspace{-1.0em}
% \footnotesize
% \renewcommand\arraystretch{1}
\begin{center}
% \begin{minipage}{0.8\textwidth}
    \caption{Performance comparison of different activation functions.}
    % \vspace{-0.5em}
    \label{tab:3}
    \begin{tabular*}{0.62\textwidth}{@{\extracolsep{\fill}}cccccc@{\extracolsep{\fill}}}
     \toprule
     \multirow{2}{*}{Method} & \multicolumn{2}{c}{Place365} & \multicolumn{2}{c}{Kodak} \\\cmidrule{2-3} \cmidrule{4-5}
     & PSNR & SSIM & PSNR & SSIM \\
     \midrule
     ReLU & 30.77 & 0.9020 & 31.89 & 0.8898 \\
     LeakyReLU & 30.76 & 0.9015 & 31.87 & 0.8894 \\
     GELU & 30.76 & 0.9017 & 31.85 & 0.8896 \\
     \botrule
    \end{tabular*}
% \end{minipage}
\end{center}
% \vspace{-1em}
\end{table}

Besides, we also compared three common feature blocks: residual block (RB) \citep{he2016deep}, residual channel attention block (RCAB) \citep{zhang2018image} and residual-in-residual dense block (RRDB) \citep{wang2018esrgan} to explore the performance of encoder and decoder of MSPRL. Both RCAB and RRDB will increase the computational complexity, and RRDB will greatly increase the model parameters, while RB can maintain model performance between low computational complexity and parameters. Their parameters and performance comparisons are shown in \tabref{tab:8}.

% different feature blocks
\begin{table}[ht]
\footnotesize
\begin{center}
    \caption{Comparison of PSNR performance of different feature blocks.}
    % \vspace{-0.5em}
    \label{tab:8}
    % \resizebox{\linewidth}{!}{
    \begin{tabular}{ccccc}
    \toprule
    Method & Amounts & Total parameters & Place365 & Kodak\\
     \midrule
     RB & 8 & 9681505 & 30.77 & 31.89 \\
     RCAB & 8 & 9745489 & 30.79 & 31.85 \\
     RRDB & 2 & 22082593 & 30.80 & 31.90 \\
     \botrule
    \end{tabular}
    % }
\end{center}
% \vspace{-2em}
\end{table}

\subsection{Impact of Training Strategies}\label{Impact}
\begin{figure}[ht]
% \vspace{-1.0em}
\begin{center}
% \resizebox{\linewidth}{!}{
\begin{minipage}{0.6\linewidth}
    \subfloat{\includegraphics[width=1\linewidth]{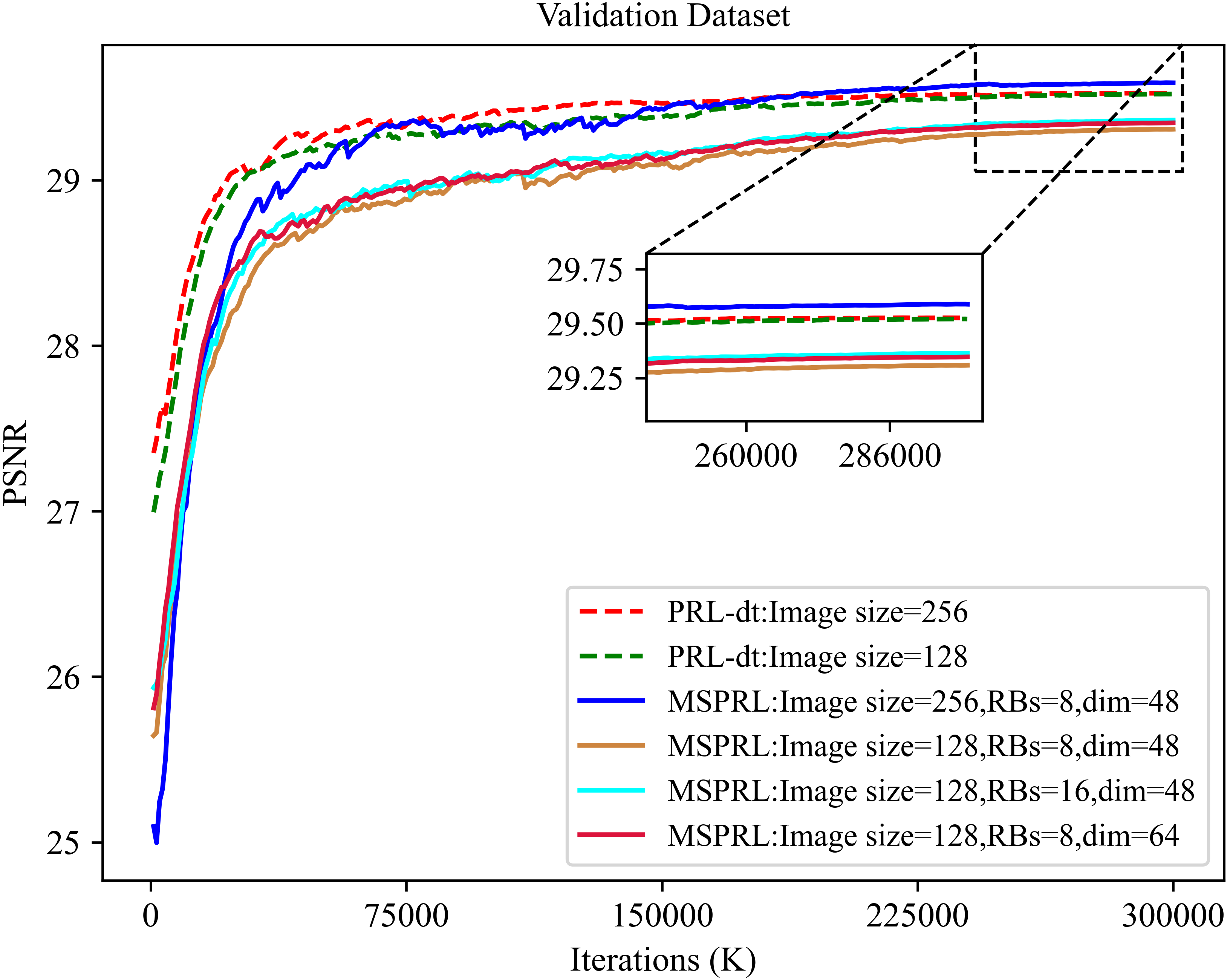}}
\end{minipage}
\\ \vspace{0.5em}
% }
\caption{Validation curve of PRL-dt and MSPRL in different training setting. We show PSNR on the 3000 validation images on the VOC2012 dataset. The PSNR performance is shown in \tabref{tab:4}.}
\label{fig:4}
\end{center}
\vspace{-1em}
\end{figure}

To explore the impact of training strategies, we conduct multiple experiments with different image sizes and loss functions using PRL and MSPRL models, respectively. We use the original PRL baseline and only use our different training strategies as shown in \tabref{tab:5}. The average improvement is approximately 1.5 dB in all test datasets, and is named \textbf{PRL-dt}. For image size, we found that the training time sharply decreased by approximately 70\% when using images smaller than 128 pixels and that the performance was comparable to larger images of size 256. We assert that this phenomenon is due to data augmentation, random sampling and more iterations, which make the model learn as much feature information as from large images on small image sizes. For different loss functions, we minimize the fast Fourier transform loss in the frequency domain, so that the model can be further optimized and improved in image details compared to only using a single L1 loss function. The experimental results are shown in \tabref{tab:4}. Meanwhile, we also test the performance of our MSPRL with different numbers of channels and residual blocks in \tabref{tab:7}. The validation PSNR curves of the model under these different settings are shown in \figref{fig:4}.

\begin{table}[ht]
\footnotesize
\begin{center}
    \caption{Performance comparison of different number of channels and residual blocks (RBs).}
    % \vspace{-0.5em}
    \label{tab:7}
    \begin{tabular*}{0.62\textwidth}{@{\extracolsep{\fill}}ccccccc@{\extracolsep{\fill}}}
     \toprule
     \multirow{2}{*}{Channels}& \multirow{2}{*}{RBs}& 
     \multicolumn{2}{c}{Place365} & \multicolumn{2}{c}{Kodak}\\
     \cmidrule{3-4}\cmidrule{5-6}
     & & PSNR & SSIM & PSNR & SSIM \\
     \midrule
     48 & 8 & 30.77 & 0.9020 & 31.89 & 0.8898 \\ 
     64 & 8 & 30.77 & 0.9022 & 31.89 & 0.8900 \\
     48 & 16 & 30.80 & 0.9025 & 31.93 & 0.8904 \\
     \botrule
    \end{tabular*}
\end{center}
\vspace{-2em}
\end{table}

\begin{table}[ht]
\footnotesize
\centering
\caption{Performance comparison at different channels and image sizes. \textcolor{green}{Green} and \textcolor{blue}{blue} show the best PSNR for PRL and MSPRL, respectively, under different settings. L1 means using a separate L1 loss function. PRL-dt uses the original PRL baseline, but in this paper uses different training strategies. Details are discussed in Sec. \ref{Impact}.}
% \vspace{-0.5em}
\label{tab:4}
% \resizebox{\linewidth}{!}{
    \begin{tabular}{ccccc}
     \toprule
     Model & Image size & Training time & Place365 & Kodak \\
     \midrule
     PRL & 256$\times$256 & - & 29.23 & 30.28 \\
     \cmidrule{1-5}
     \multirow{2}*{PRL-dt} & 256$\times$256 & 2 Days & \textcolor{green}{30.65} & \textcolor{green}{31.72} \\
     ~ & 128$\times$128 & 17 Hours & \textcolor{green}{30.65} & 31.71 \\
     \cmidrule{1-5}
     MSPRL(L1) & 128$\times$128 & 18 Hours & 30.75 & 31.82 \\
     \cmidrule{1-5}
     \multirow{2}*{MSPRL} & 256$\times$256 & 2.2 Days &30.76 & 31.87 \\
     ~ & 128$\times$128 & 18 Hours & \textcolor{blue}{30.77} & \textcolor{blue}{31.89} \\
     \botrule
    \end{tabular}
% }
% \vspace{-1em}
\end{table}

\begin{table*}[ht]
% \vspace{-1.0em}
% \small
\renewcommand\arraystretch{1.3}
\centering
\caption{Performance comparison of inverse halftoning methods on different datasets. We improve the performance of PRL only using a different training strategy, which model is named \textbf{PRL-dt}. Our MSPRL outperforms all other models and is closer to the true grayscale image in detail recovery. The top two results are marked in \textcolor{red}{red} and \textcolor{blue}{blue}.}
% \vspace{-0.5em}
\label{tab:1}
\resizebox{\linewidth}{!}{
\begin{tabular}{lcccccccccccccc}
 \toprule
 \multirow{2}{*}{Model} & \multicolumn{2}{c}{Place365}
 & \multicolumn{2}{c}{Kodak} & \multicolumn{2}{c}{Set5}
 & \multicolumn{2}{c}{Set14} & \multicolumn{2}{c}{BSD100}
 & \multicolumn{2}{c}{Urban100} & \multicolumn{2}{c}{Manga109}
 \\ \cmidrule{2-15}
  & PSNR & SSIM & PSNR & SSIM & PSNR & SSIM 
  & PSNR & SSIM & PSNR & SSIM & PSNR & SSIM & PSNR & SSIM
  \\ \midrule
 DnCNN \citep{zhang2017beyond} & 30.31 & 0.8913 & 31.24 & 0.8759 & 33.26 & 0.9192 & 30.76 & 0.8812 & 29.72 & 0.8600 & 29.81 & 0.9031 & 33.44 & 0.9427 \\
 VDSR \citep{kim2016accurate} & 30.15 & 0.8868 & 30.97 & 0.8718 & 32.92 & 0.9134 & 30.44 & 0.8758 & 29.53 & 0.8555 & 29.34 & 0.8964 & 32.87 & 0.9391 \\
 EDSR \citep{lim2017enhanced} & 30.48 & 0.8960 & 31.48 & 0.8830 & 33.42 & 0.9219 & 30.95 & 0.8857 & 29.86 & 0.8652 & 30.22 & 0.9106 & 33.90 & 0.9466 \\
 PRL \citep{xia2018deep} & 29.23 & 0.8840 & 30.28 & 0.8722 & 32.06 & 0.9103 & 29.97 & 0.8746 & 28.99 & 0.8525 & 29.39 & 0.9017 & 32.55 & 0.9365 \\
 GGRL \citep{yuan2019gradient} & 30.46 & 0.8960 & 31.44 &  0.8830 & - & - & - & - & 29.85 & 0.8654 & - & - & - & - \\ 
 MIMOUNet \citep{cho2021rethinking} & 30.56 & 0.8977 & 31.55 & 0.8855 & 33.54 & 0.9235 & 31.07 & 0.8883 & 29.91 & 0.8674 & 30.41 & 0.9140 & 34.21 & 0.9488 \\
 PRL-dt (ours) & \textcolor{blue}{30.65} & \textcolor{blue}{0.9000} & \textcolor{blue}{31.71} & \textcolor{blue}{0.8875} & \textcolor{blue}{33.70} & \textcolor{blue}{0.9254} & \textcolor{blue}{31.25} & \textcolor{blue}{0.8904} & \textcolor{blue}{30.01} & \textcolor{blue}{0.8691} & \textcolor{blue}{30.71} & \textcolor{blue}{0.9183} & \textcolor{blue}{34.50} & \textcolor{blue}{0.9502} \\
 MSPRL (ours) & \textcolor{red}{30.77} & \textcolor{red}{0.9020} & \textcolor{red}{31.89} & \textcolor{red}{0.8898} & \textcolor{red}{33.81} & \textcolor{red}{0.9264} & \textcolor{red}{31.40} & \textcolor{red}{0.8925} & \textcolor{red}{30.09} & \textcolor{red}{0.8708} & \textcolor{red}{31.10} & \textcolor{red}{0.9226} & \textcolor{red}{34.85} & \textcolor{red}{0.9518} \\
 \botrule
\end{tabular}
}
% \vspace{-1em}
\end{table*}

\subsection{Performance comparison}
We compare MSPRL with other inverse halftoning methods and CNN models of relevant vision tasks, such as DnCNN \citep{zhang2017beyond}, VDSR \citep{kim2016accurate} and EDSR \citep{lim2017enhanced}. The single baseline model in EDSR, which contains 16 residual blocks with 64 convolution kernel channels, is used in our task. We remove data pre/postprocessing and upscaling layers. For GGRL \citep{yuan2019gradient}, the public pretrained model is not available and their training dataset size is 8 times our dataset. Therefore, we only use the GGRL model in our training process, leading to some gaps in its performance compared to the original paper. In order to distinguish similar models, we also test MIMOUNet \citep{cho2021rethinking}. For a fair comparison, these methods employ our training strategies. Because DnCNN, VDSR and EDSR adopt our training strategy, their results are higher than the values of the corresponding models trained in \citep{xia2018deep}. The performance comparison is demonstrated in \tabref{tab:1}. The experimental results show that our MSPRL obtains the best performance on multiple datasets by 0.3 dB gain. Especially on the Urban100 dataset, MSPRL is 0.69 dB higher than MIMOUNet, meanwhile, other models outperform the original PRL due to our training strategies. We also changed the training strategy of PRL, named PRL-dt, and its model performance greatly improved compared with original PRL. The average PSNR on multiple datasets improved by approximately 1.5 dB only by changing the training strategy. Finally, MSPRL also outperforms PRL-dt on all datasets.

\begin{figure*}[ht]
\begin{center}
% \vspace{-0.1em}
    \captionsetup[subfloat]{labelsep=none,format=plain,labelformat=empty,font={scriptsize}}
        % Lena
        \begin{minipage}[b]{0.255\linewidth}
        	\subfloat[Lena]{\includegraphics[width=1\linewidth]{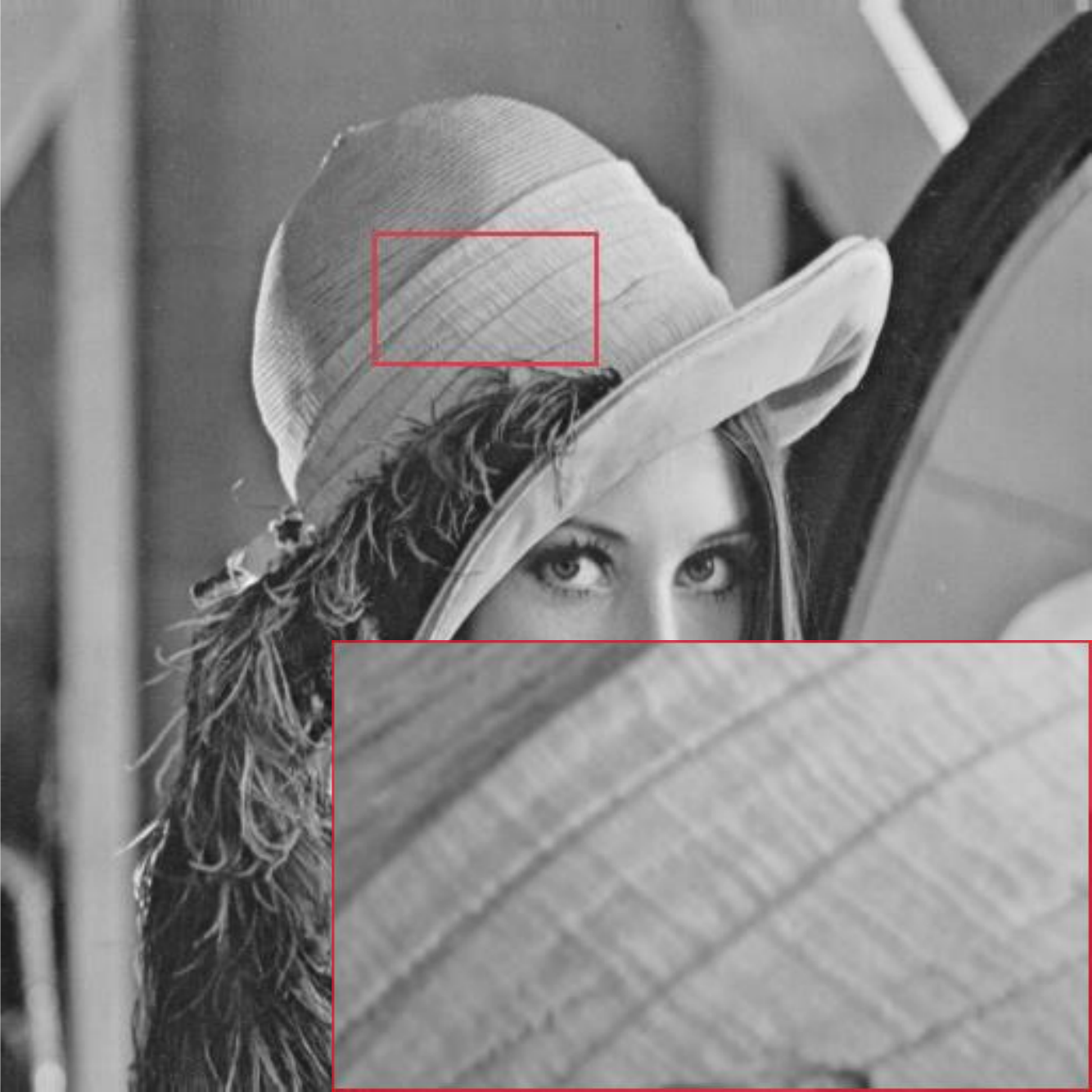}}
        \end{minipage}
        \begin{minipage}[b]{0.175\linewidth}
        	\subfloat[DnCNN]{\includegraphics[width=1\linewidth]{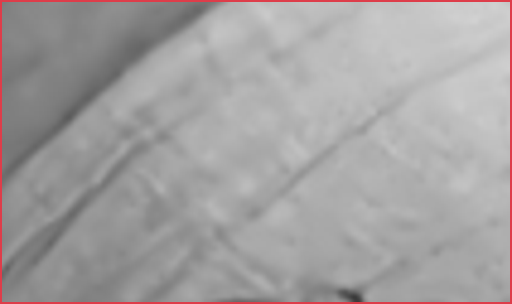}} \vspace{-0.8em}
        	\subfloat[GGRL]{\includegraphics[width=1\linewidth]{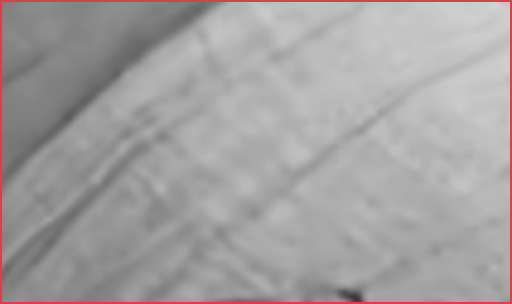}}
        \end{minipage}
        \begin{minipage}[b]{0.175\linewidth}
        	\subfloat[VDSR]{\includegraphics[width=1\linewidth]{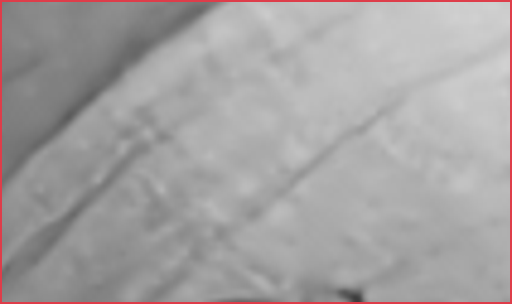}} \vspace{-0.8em}
        	\subfloat[MIMOUNet]{\includegraphics[width=1\linewidth]{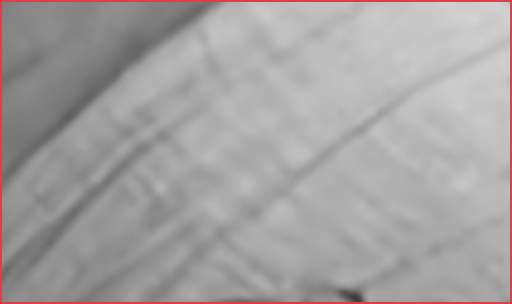}}
        \end{minipage}
        \begin{minipage}[b]{0.175\linewidth}
        	\subfloat[EDSR]{\includegraphics[width=1\linewidth]{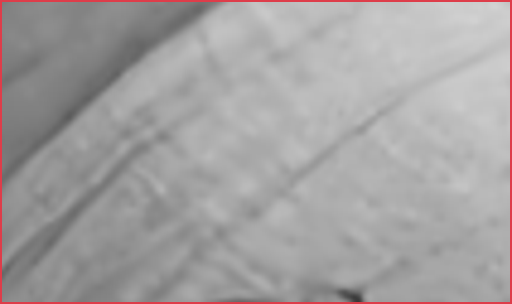}} \vspace{-0.8em}
        	\subfloat[PRL-dt (ours)]{\includegraphics[width=1\linewidth]{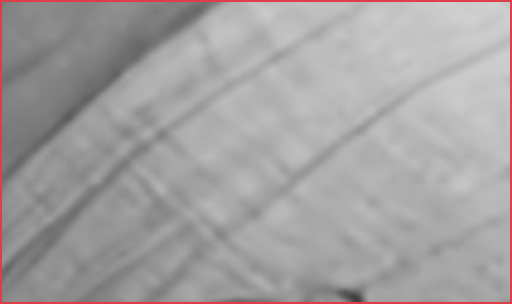}}
        \end{minipage}
        \begin{minipage}[b]{0.175\linewidth}
        	\subfloat[PRL]{\includegraphics[width=1\linewidth]{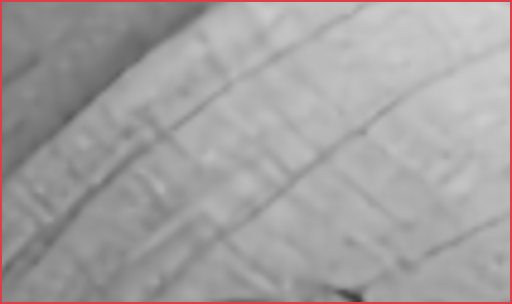}} \vspace{-0.8em}
        	\subfloat[MSPRL (ours)]{\includegraphics[width=1\linewidth]{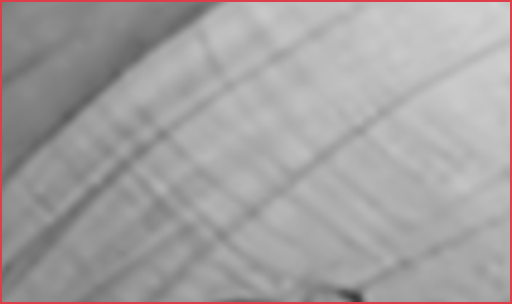}}
        \end{minipage}
        \\ \vspace{0.5em}
        % barbara
        \begin{minipage}[b]{0.255\linewidth}
        	\subfloat[Barbara]{\includegraphics[width=1\linewidth]{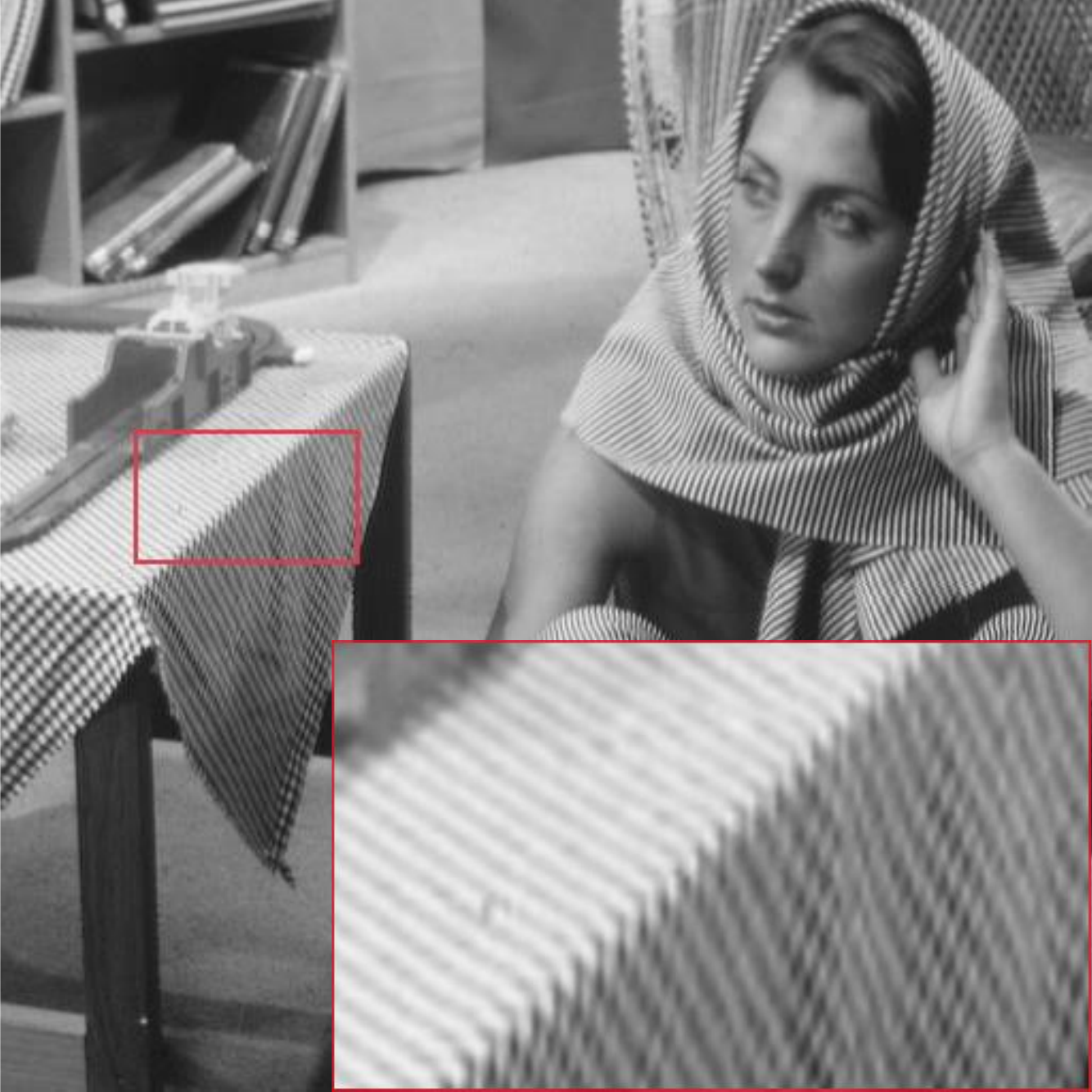}}
        \end{minipage}
        \begin{minipage}[b]{0.175\linewidth}
        	\subfloat[DnCNN]{\includegraphics[width=1\linewidth]{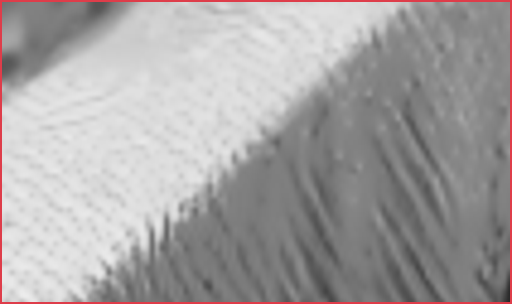}} \vspace{-0.8em}
        	\subfloat[GGRL]{\includegraphics[width=1\linewidth]{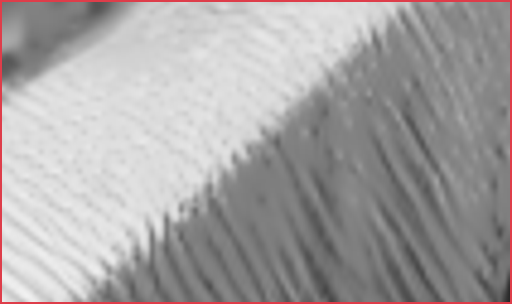}}
        \end{minipage}
        \begin{minipage}[b]{0.175\linewidth}
        	\subfloat[VDSR]{\includegraphics[width=1\linewidth]{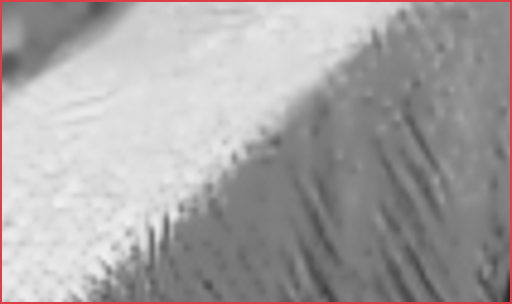}} \vspace{-0.8em}
        	\subfloat[MIMOUNet]{\includegraphics[width=1\linewidth]{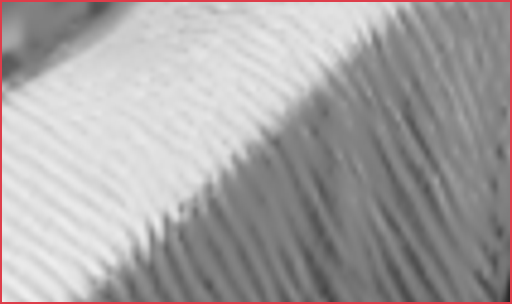}}
        \end{minipage}
        \begin{minipage}[b]{0.175\linewidth}
        	\subfloat[EDSR]{\includegraphics[width=1\linewidth]{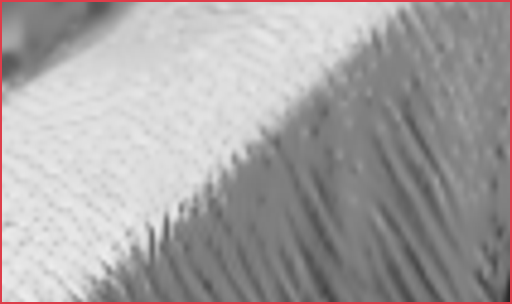}} \vspace{-0.8em}
        	\subfloat[PRL-dt (ours)]{\includegraphics[width=1\linewidth]{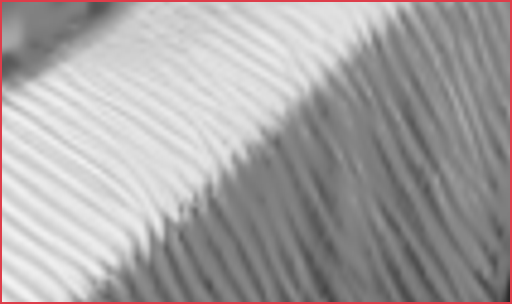}}
        \end{minipage}
        \begin{minipage}[b]{0.175\linewidth}
        	\subfloat[PRL]{\includegraphics[width=1\linewidth]{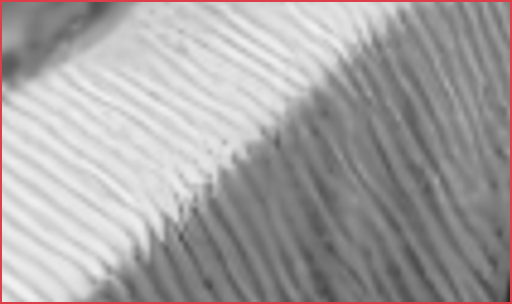}} \vspace{-0.8em}
        	\subfloat[MSPRL (ours)]{\includegraphics[width=1\linewidth]{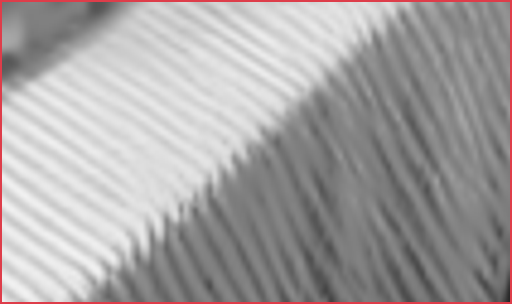}}
        \end{minipage}
        \\
    \end{center}
    \vspace{0.5em}
    \caption{Compared with the other methods, our MSPRL more effectively restores the image details.}
    \label{fig:c1}
% \vspace{-1em}
\end{figure*}

\begin{figure*}[ht]
% \vspace{-6.em}
\begin{center}
    \captionsetup[subfloat]{labelsep=none,format=plain,labelformat=empty,font={scriptsize}}
        % kodim19
        \begin{minipage}[b]{0.255\linewidth}
        	\subfloat[kodim19]{\includegraphics[width=1\linewidth]{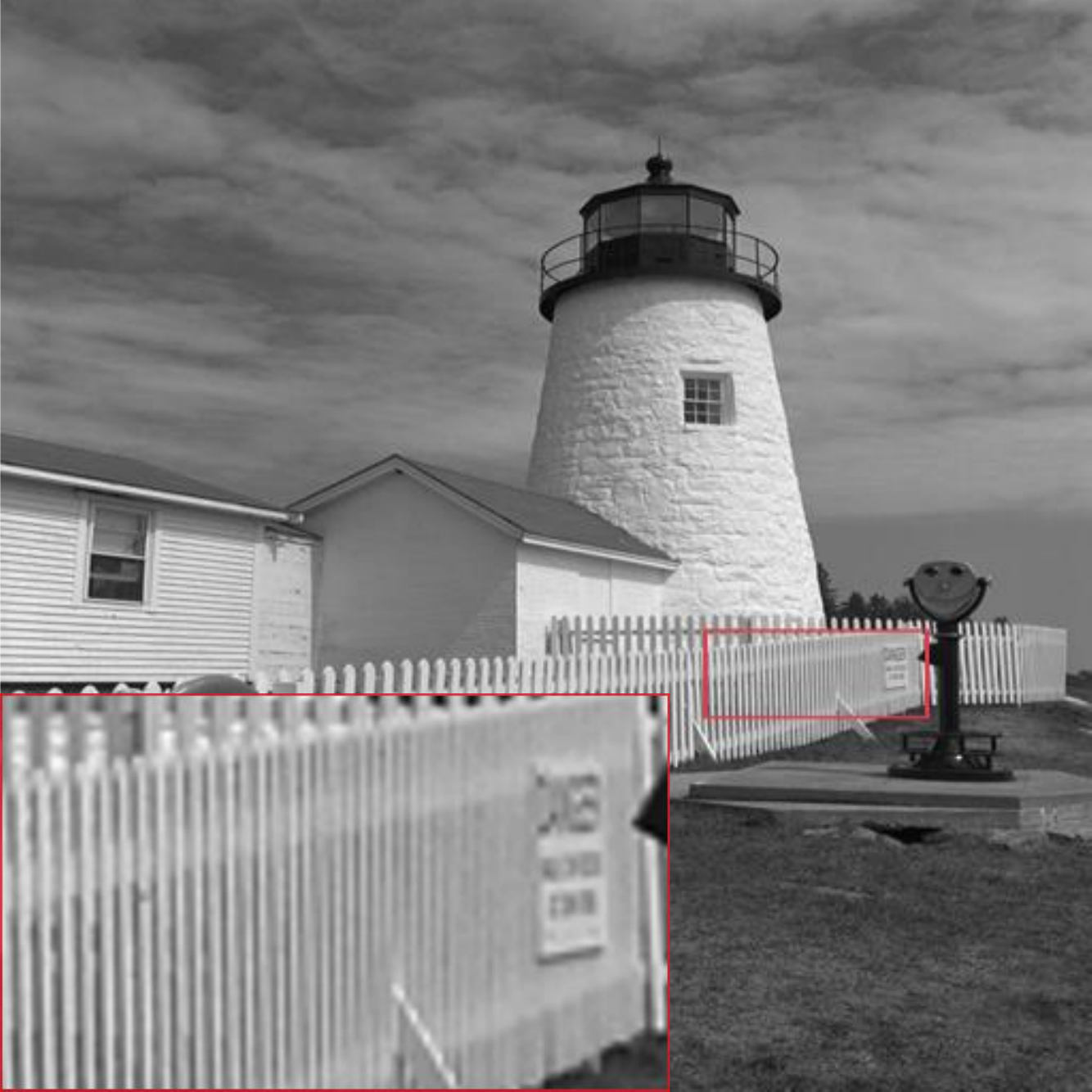}}
        \end{minipage}
        \begin{minipage}[b]{0.175\linewidth}
        	\subfloat[DnCNN]{\includegraphics[width=1\linewidth]{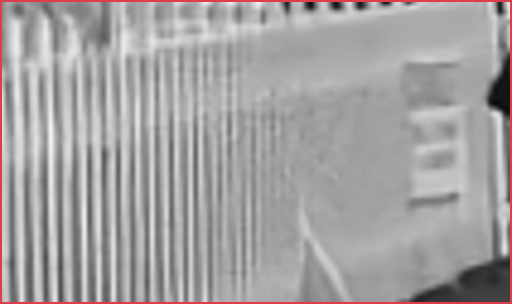}} \vspace{-0.8em}
        	\subfloat[GGRL]{\includegraphics[width=1\linewidth]{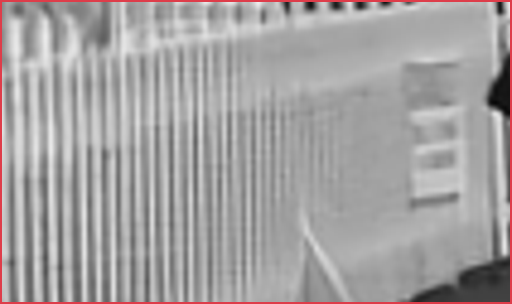}}
        \end{minipage}
        \begin{minipage}[b]{0.175\linewidth}
        	\subfloat[VDSR]{\includegraphics[width=1\linewidth]{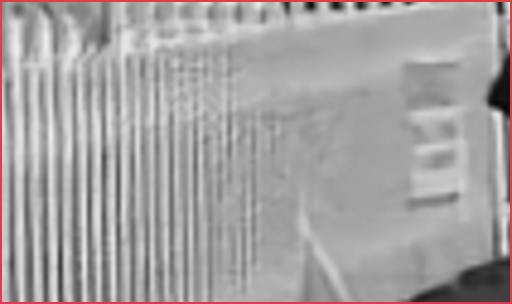}} \vspace{-0.8em}
        	\subfloat[MIMOUNet]{\includegraphics[width=1\linewidth]{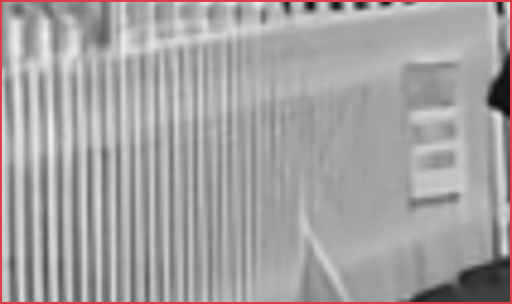}}
        \end{minipage}
        \begin{minipage}[b]{0.175\linewidth}
        	\subfloat[EDSR]{\includegraphics[width=1\linewidth]{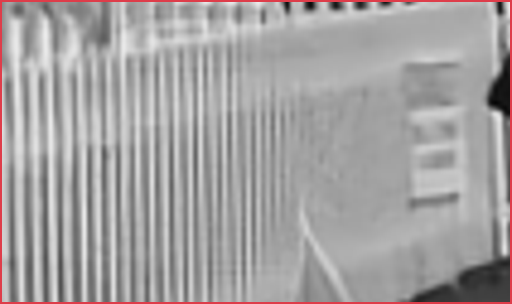}} \vspace{-0.8em}
        	\subfloat[PRL-dt (ours)]{\includegraphics[width=1\linewidth]{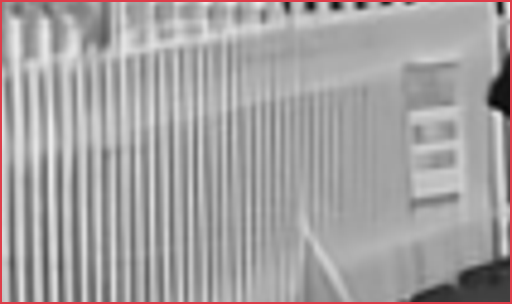}}
        \end{minipage}
        \begin{minipage}[b]{0.175\linewidth}
        	\subfloat[PRL]{\includegraphics[width=1\linewidth]{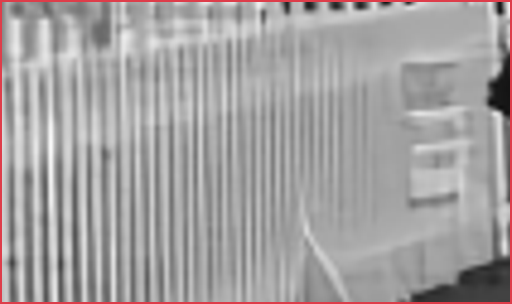}} \vspace{-0.8em}
        	\subfloat[MSPRL (ours)]{\includegraphics[width=1\linewidth]{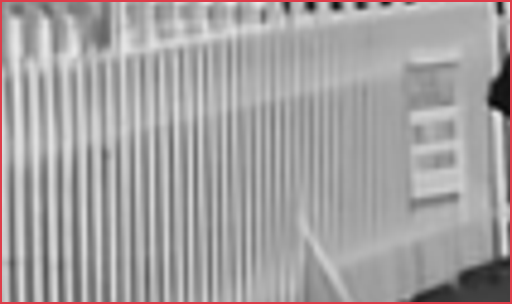}}
        \end{minipage}
        \\ \vspace{0.5em}
        % kodim24
        \begin{minipage}[b]{0.255\linewidth}
        	\subfloat[kodim24]{\includegraphics[width=1\linewidth]{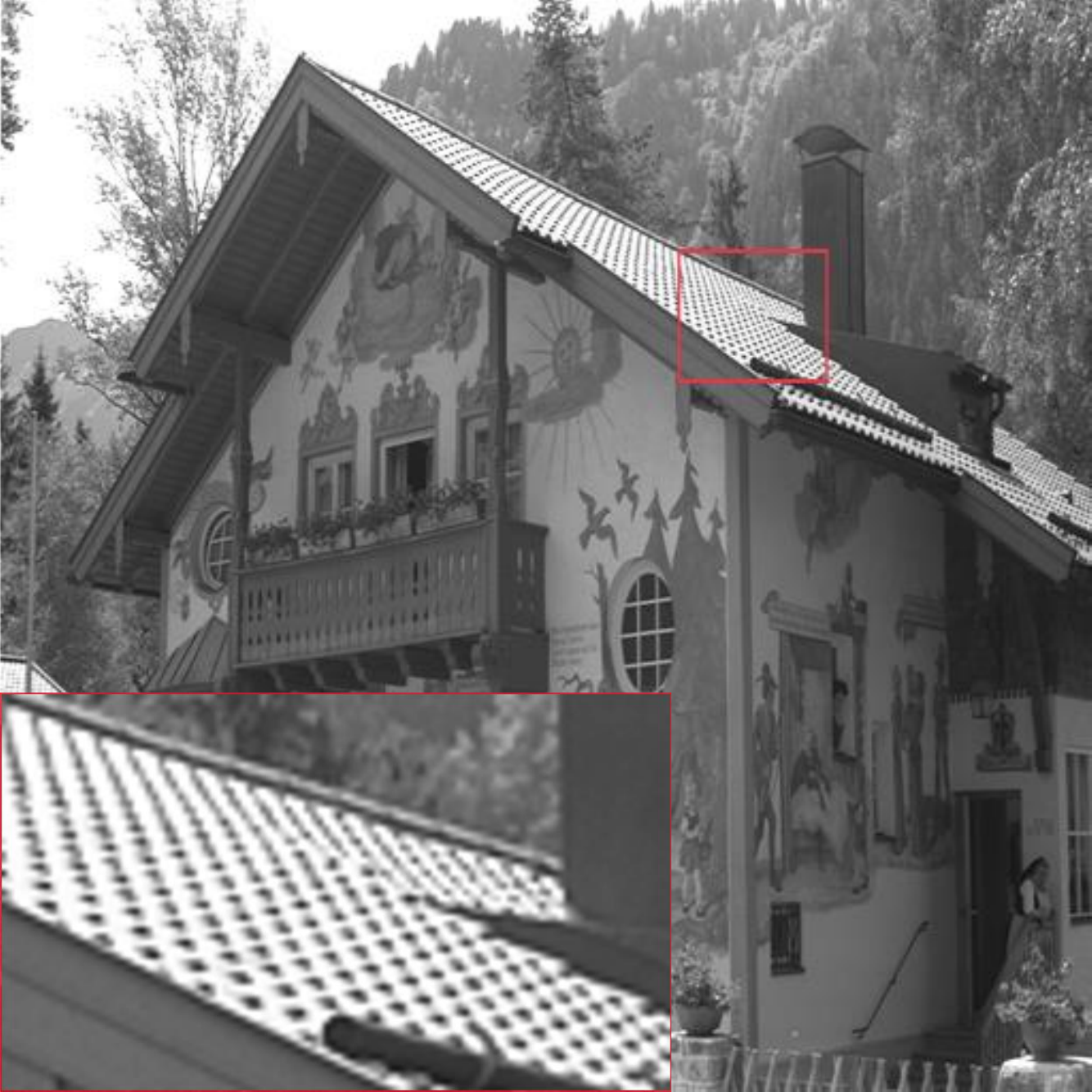}}
        \end{minipage}
        \begin{minipage}[b]{0.175\linewidth}
        	\subfloat[DnCNN]{\includegraphics[width=1\linewidth]{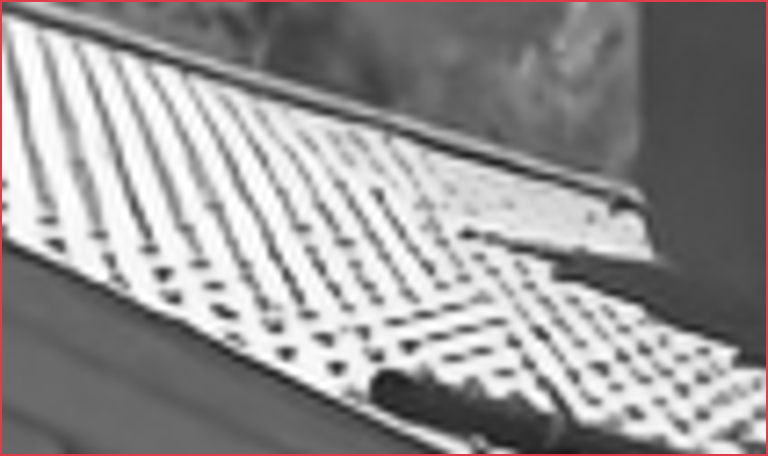}} \vspace{-0.8em}
        	\subfloat[GGRL]{\includegraphics[width=1\linewidth]{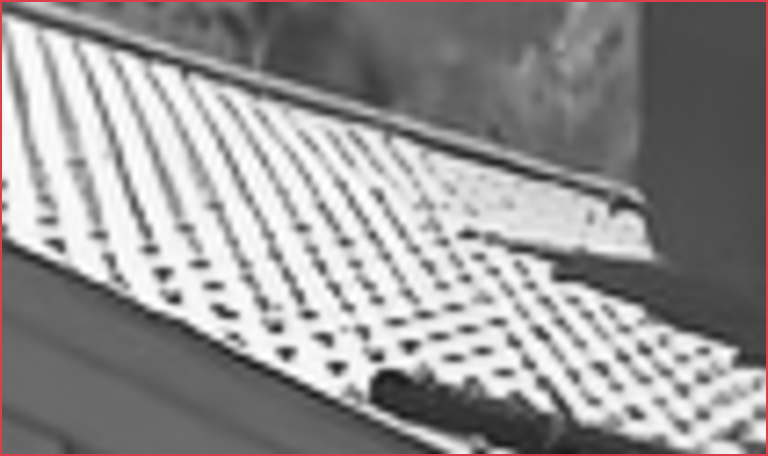}}
        \end{minipage}
        \begin{minipage}[b]{0.175\linewidth}
        	\subfloat[VDSR]{\includegraphics[width=1\linewidth]{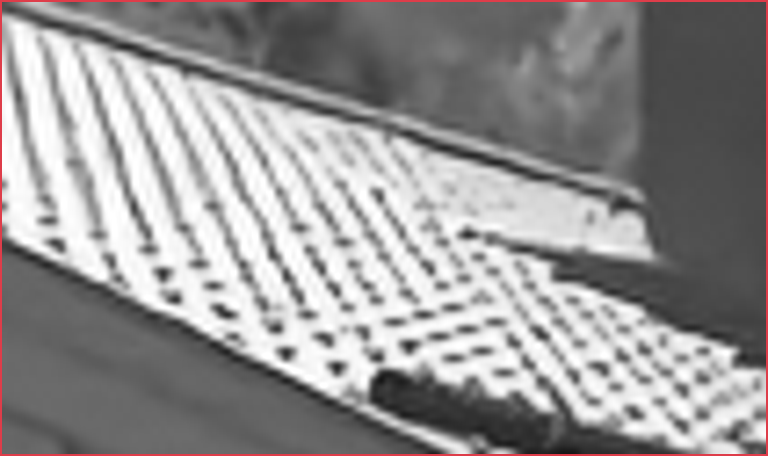}} \vspace{-0.8em}
        	\subfloat[MIMOUNet]{\includegraphics[width=1\linewidth]{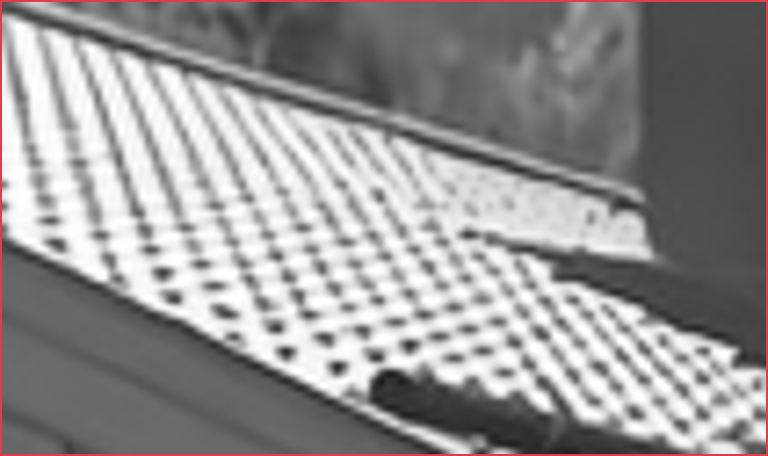}}
        \end{minipage}
        \begin{minipage}[b]{0.175\linewidth}
        	\subfloat[EDSR]{\includegraphics[width=1\linewidth]{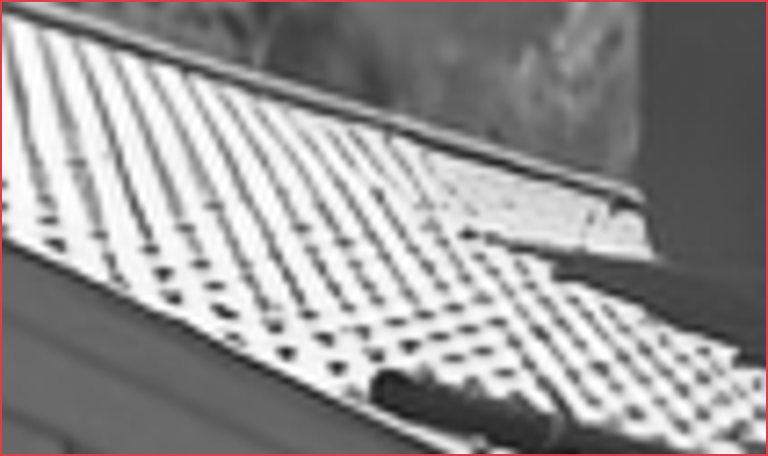}} \vspace{-0.8em}
        	\subfloat[PRL-dt (ours)]{\includegraphics[width=1\linewidth]{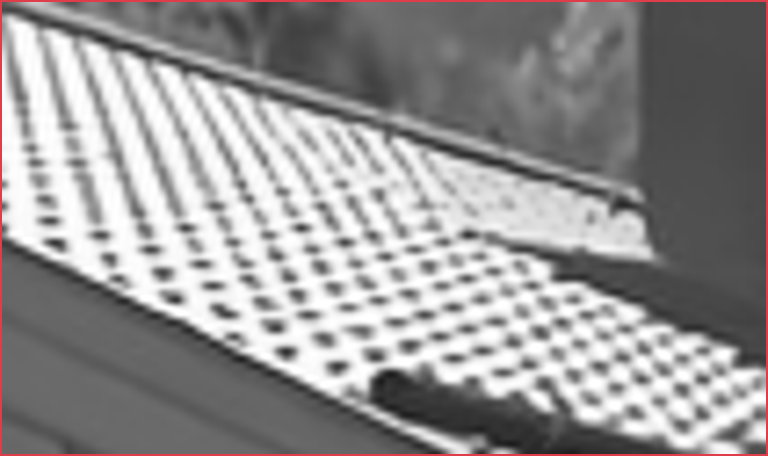}}
        \end{minipage}
        \begin{minipage}[b]{0.175\linewidth}
        	\subfloat[PRL]{\includegraphics[width=1\linewidth]{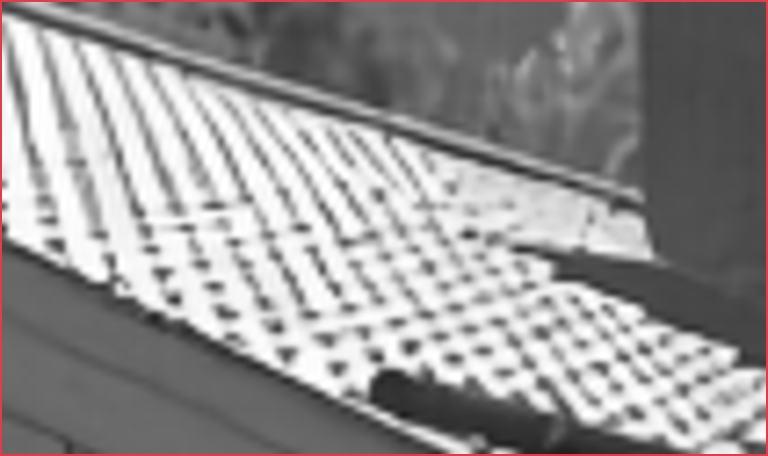}} \vspace{-0.8em}
        	\subfloat[MSPRL (ours)]{\includegraphics[width=1\linewidth]{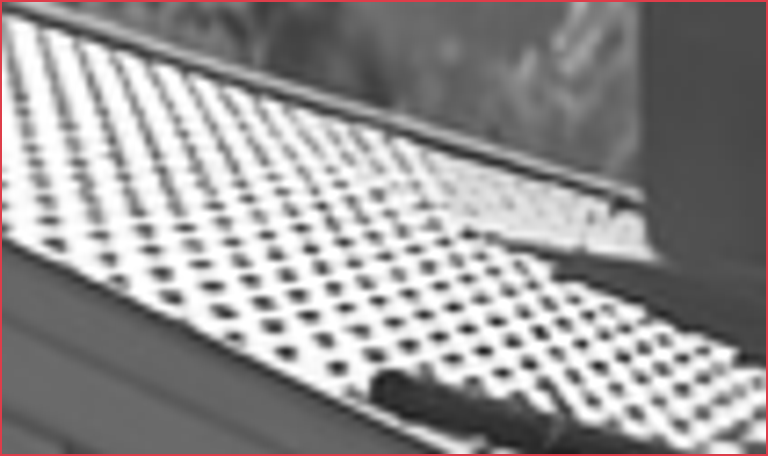}}
        \end{minipage}
        \\ \vspace{0.5em}
        % MukoukizuNoChonbo
        \begin{minipage}[b]{0.255\linewidth}
        	\subfloat[Manga109: MukoukizuNoChonbo]{\includegraphics[width=1\linewidth]{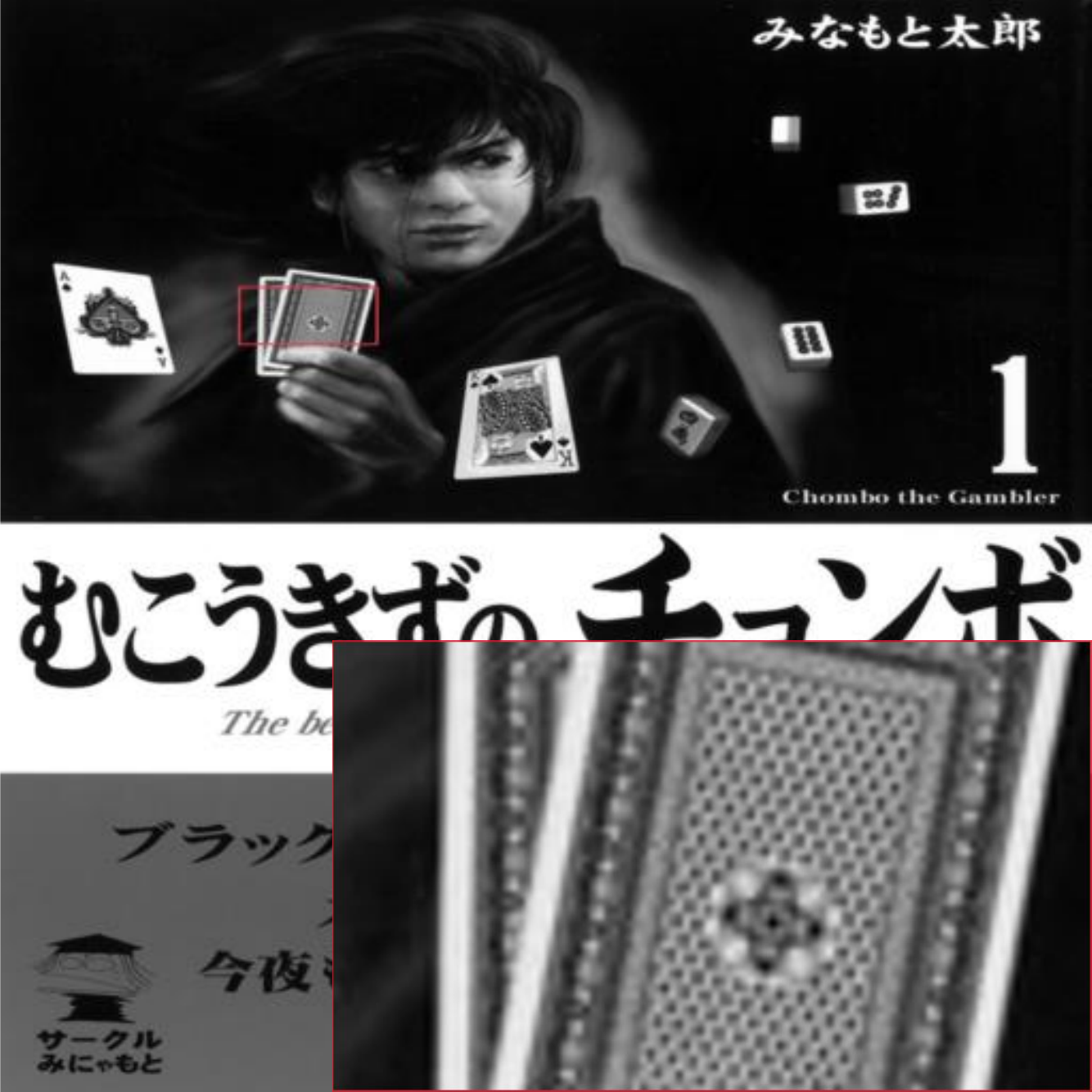}}
        \end{minipage}
        \begin{minipage}[b]{0.175\linewidth}
        	\subfloat[Halftone]{\includegraphics[width=1\linewidth]{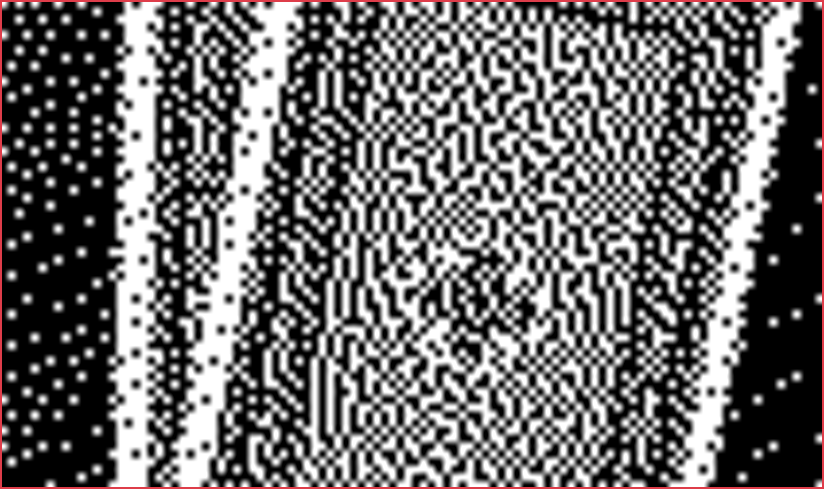}} \vspace{-0.8em}
        	\subfloat[PRL]{\includegraphics[width=1\linewidth]{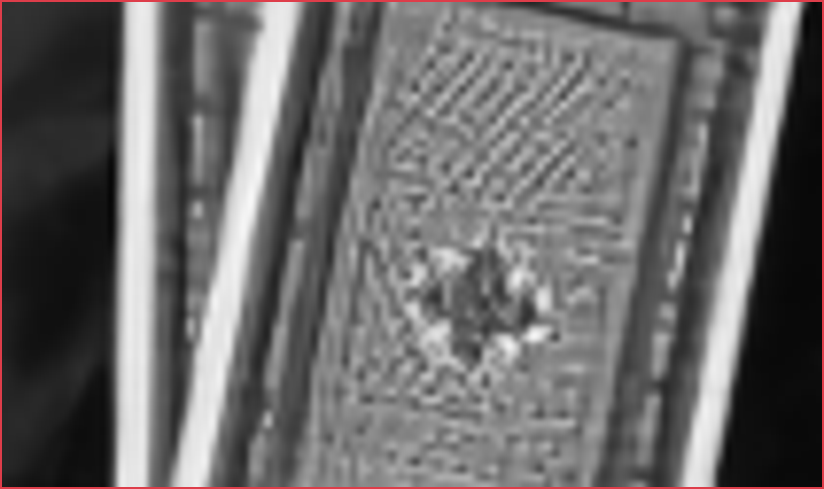}}
        \end{minipage}
        \begin{minipage}[b]{0.175\linewidth}
        	\subfloat[DnCNN]{\includegraphics[width=1\linewidth]{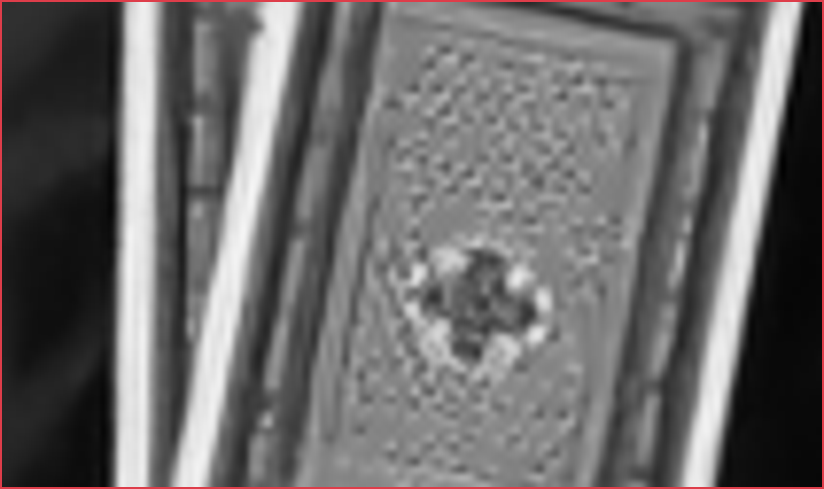}} \vspace{-0.8em}
        	\subfloat[MIMOUNet]{\includegraphics[width=1\linewidth]{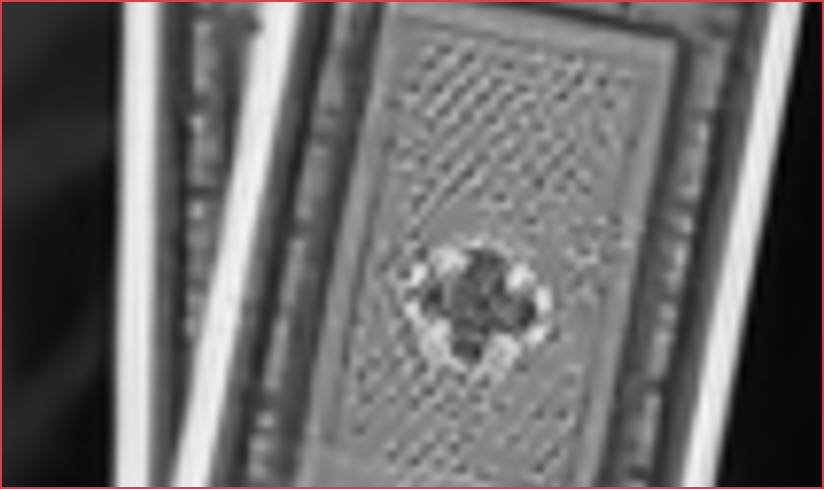}}
        \end{minipage}
        \begin{minipage}[b]{0.175\linewidth}
        	\subfloat[VDSR]{\includegraphics[width=1\linewidth]{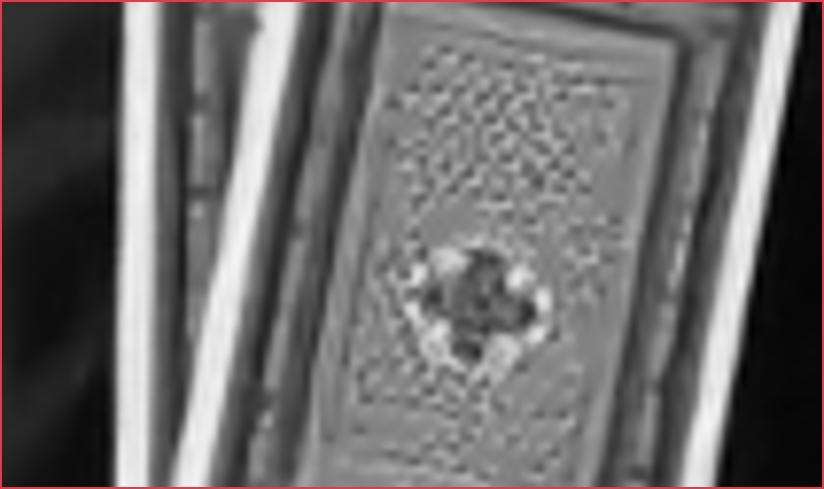}} \vspace{-0.8em}
        	\subfloat[PRL-dt (ours)]{\includegraphics[width=1\linewidth]{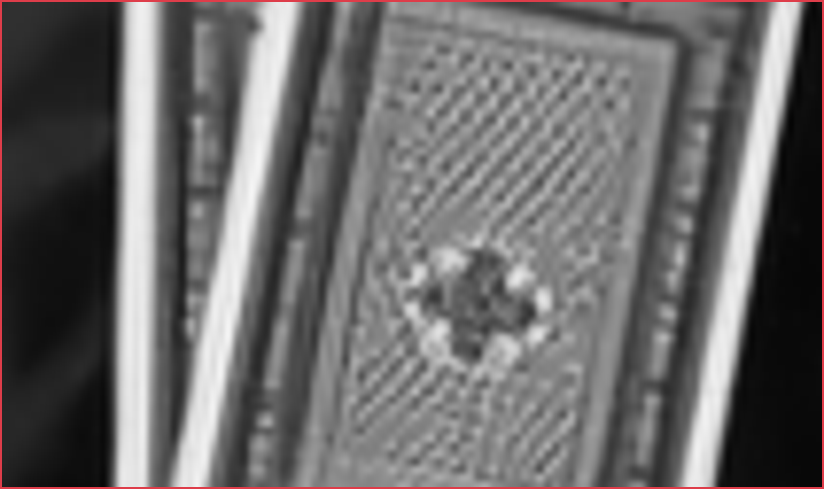}}
        \end{minipage}
        \begin{minipage}[b]{0.175\linewidth}
        	\subfloat[EDSR]{\includegraphics[width=1\linewidth]{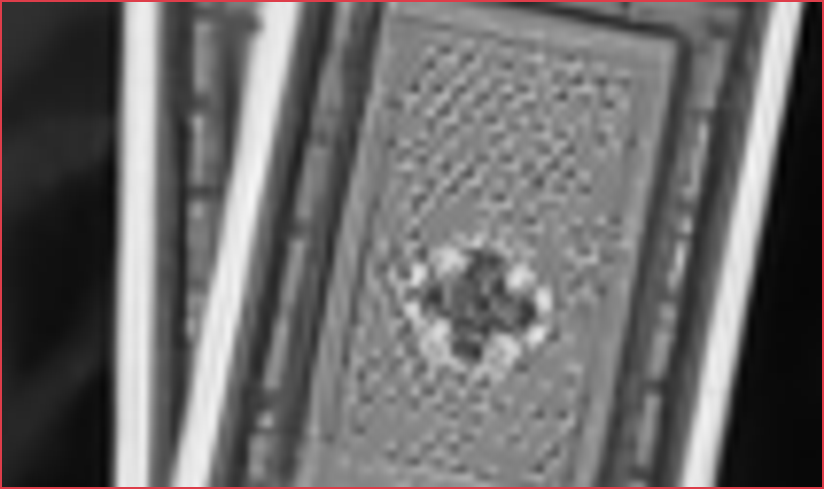}} \vspace{-0.8em}
        	\subfloat[MSPRL (ours)]{\includegraphics[width=1\linewidth]{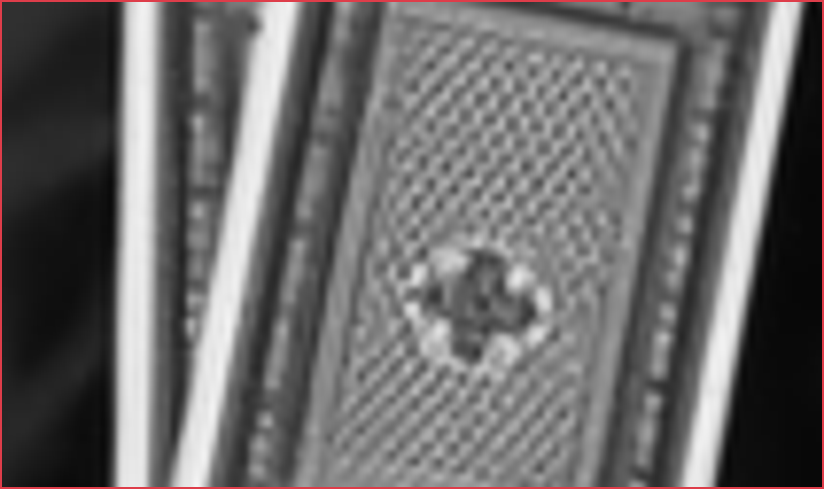}}
        \end{minipage}
        \\ \vspace{0.5em}
        % TetsuSan
        \begin{minipage}[b]{0.255\linewidth}
        	\subfloat[Manga109: TetsuSan]{\includegraphics[width=1\linewidth]{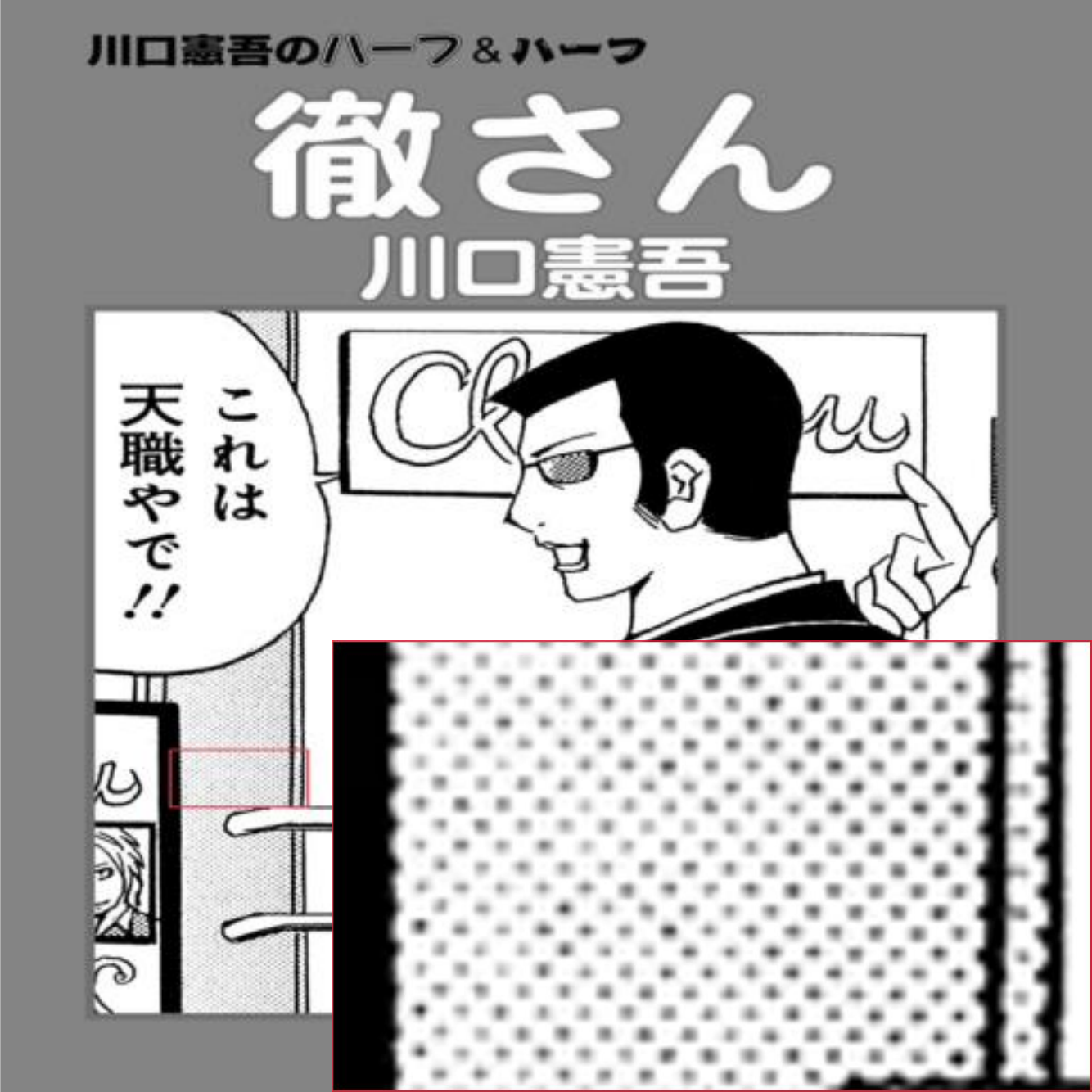}}
        \end{minipage}
        \begin{minipage}[b]{0.175\linewidth}
        	\subfloat[Halftone]{\includegraphics[width=1\linewidth]{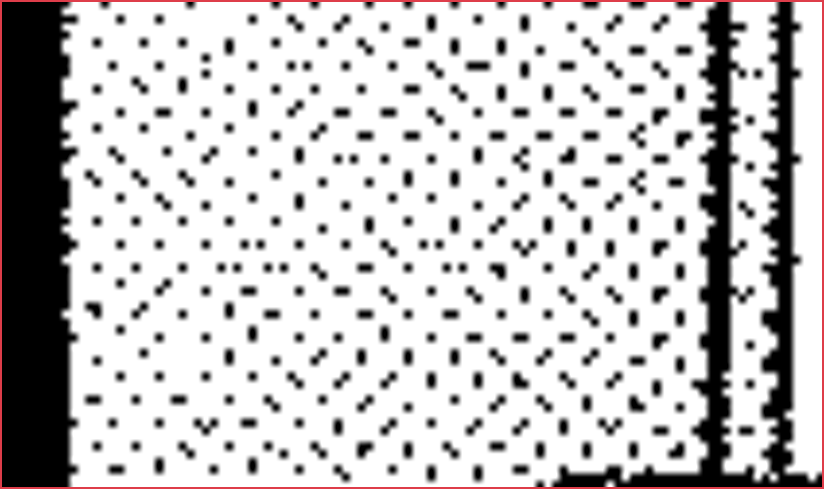}} \vspace{-0.8em}
        	\subfloat[PRL]{\includegraphics[width=1\linewidth]{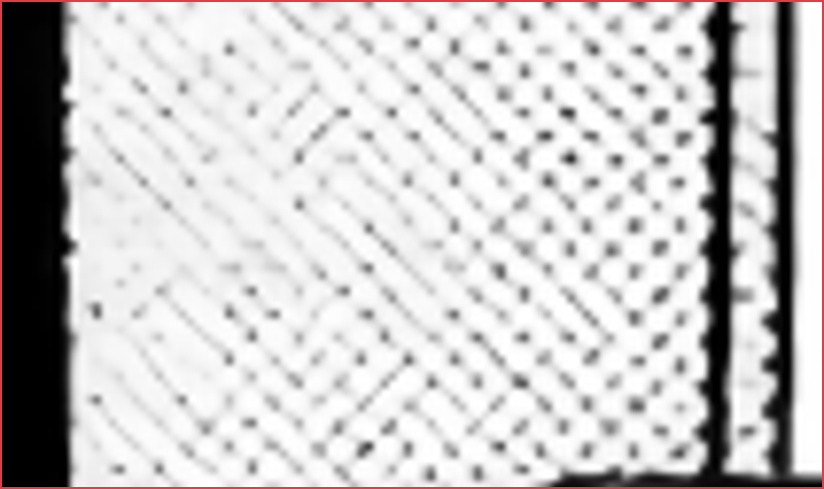}}
        \end{minipage}
        \begin{minipage}[b]{0.175\linewidth}
        	\subfloat[DnCNN]{\includegraphics[width=1\linewidth]{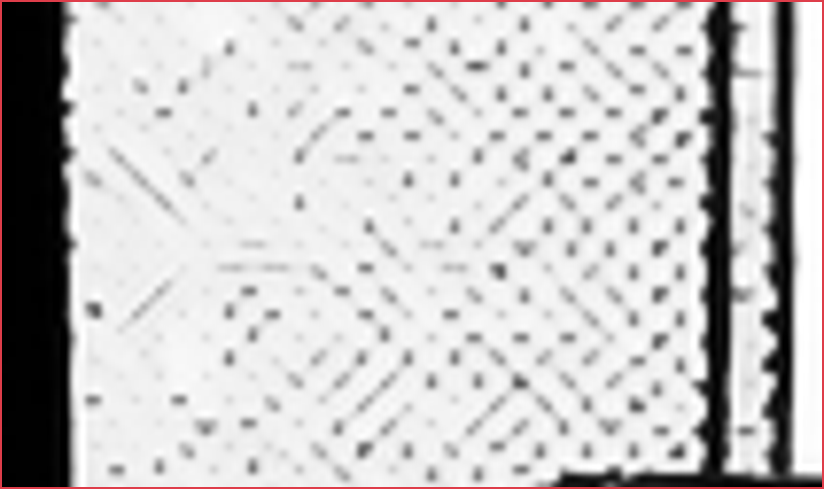}} \vspace{-0.8em}
        	\subfloat[MIMOUNet]{\includegraphics[width=1\linewidth]{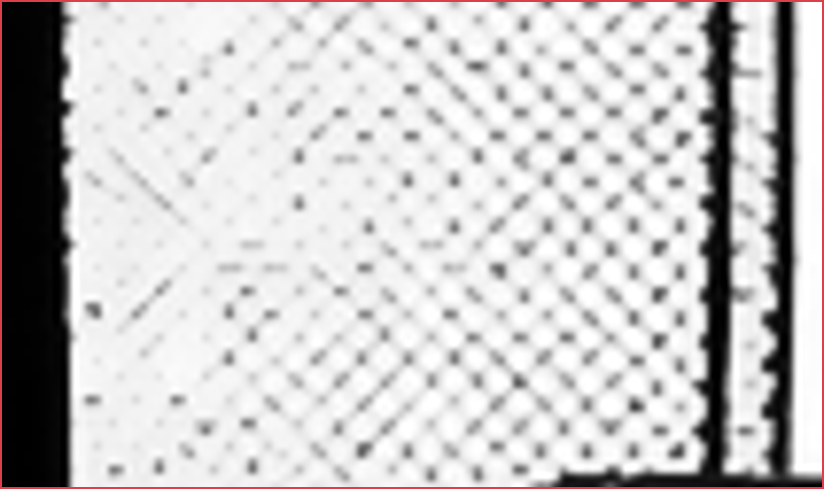}}
        \end{minipage}
        \begin{minipage}[b]{0.175\linewidth}
        	\subfloat[VDSR]{\includegraphics[width=1\linewidth]{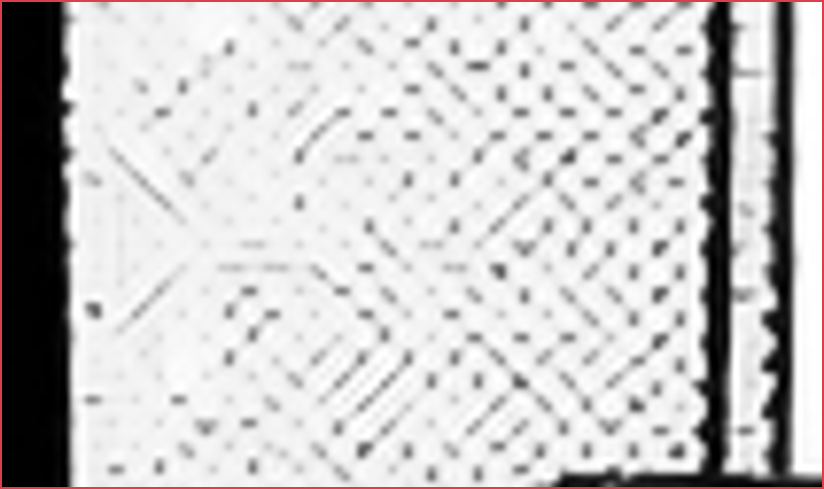}} \vspace{-0.8em}
        	\subfloat[PRL-dt (ours)]{\includegraphics[width=1\linewidth]{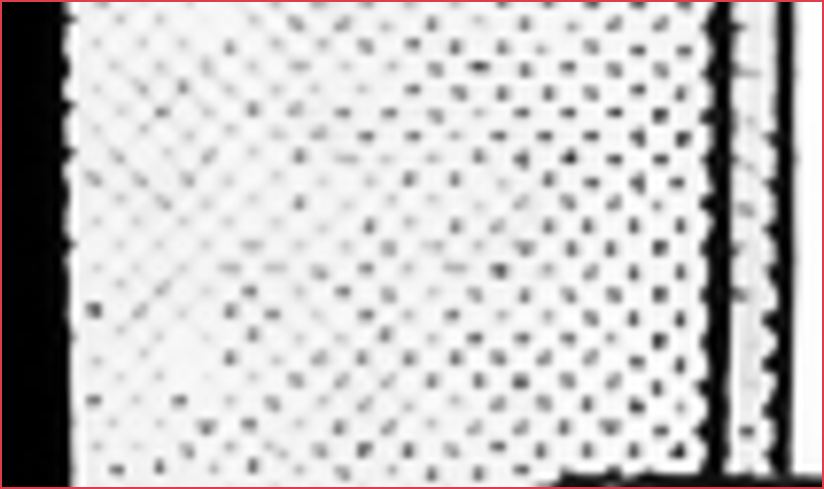}}
        \end{minipage}
        \begin{minipage}[b]{0.175\linewidth}
        	\subfloat[EDSR]{\includegraphics[width=1\linewidth]{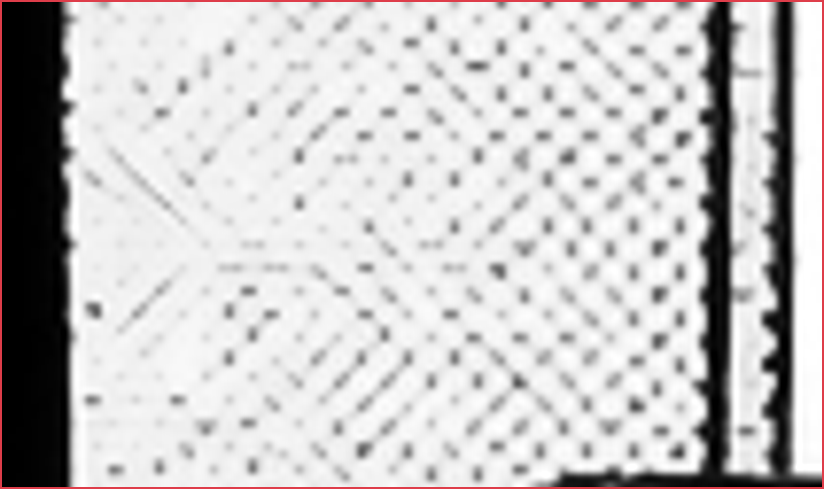}} \vspace{-0.8em}
        	\subfloat[MSPRL (ours)]{\includegraphics[width=1\linewidth]{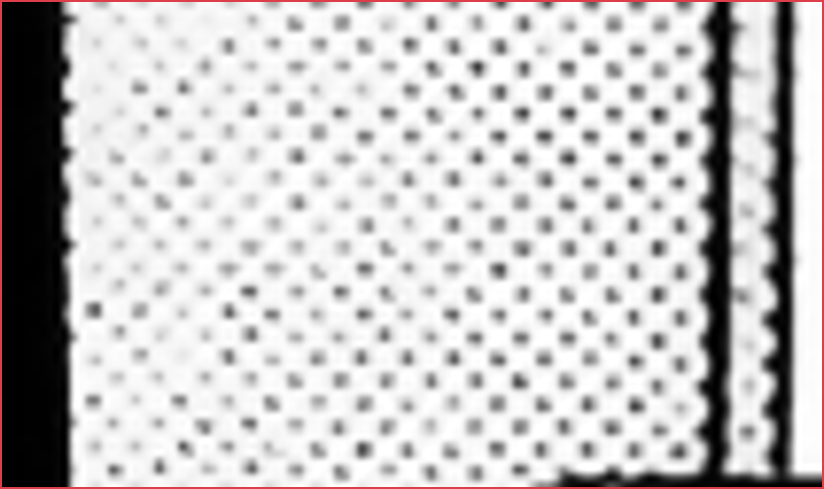}}
        \end{minipage}
        \end{center}
        \vspace{0.5em}
        \caption{Compared with the other approaches, our MSPRL more effectively restores the image details.}
    \label{fig:c2}
\end{figure*}

\begin{figure*}[ht]
\begin{center}
    \captionsetup[subfloat]{labelsep=none,format=plain,labelformat=empty,font={scriptsize}}
        % img_006
        \begin{minipage}[b]{0.255\linewidth}
        	\subfloat[Urban100: img\_006]{\includegraphics[width=1\linewidth]{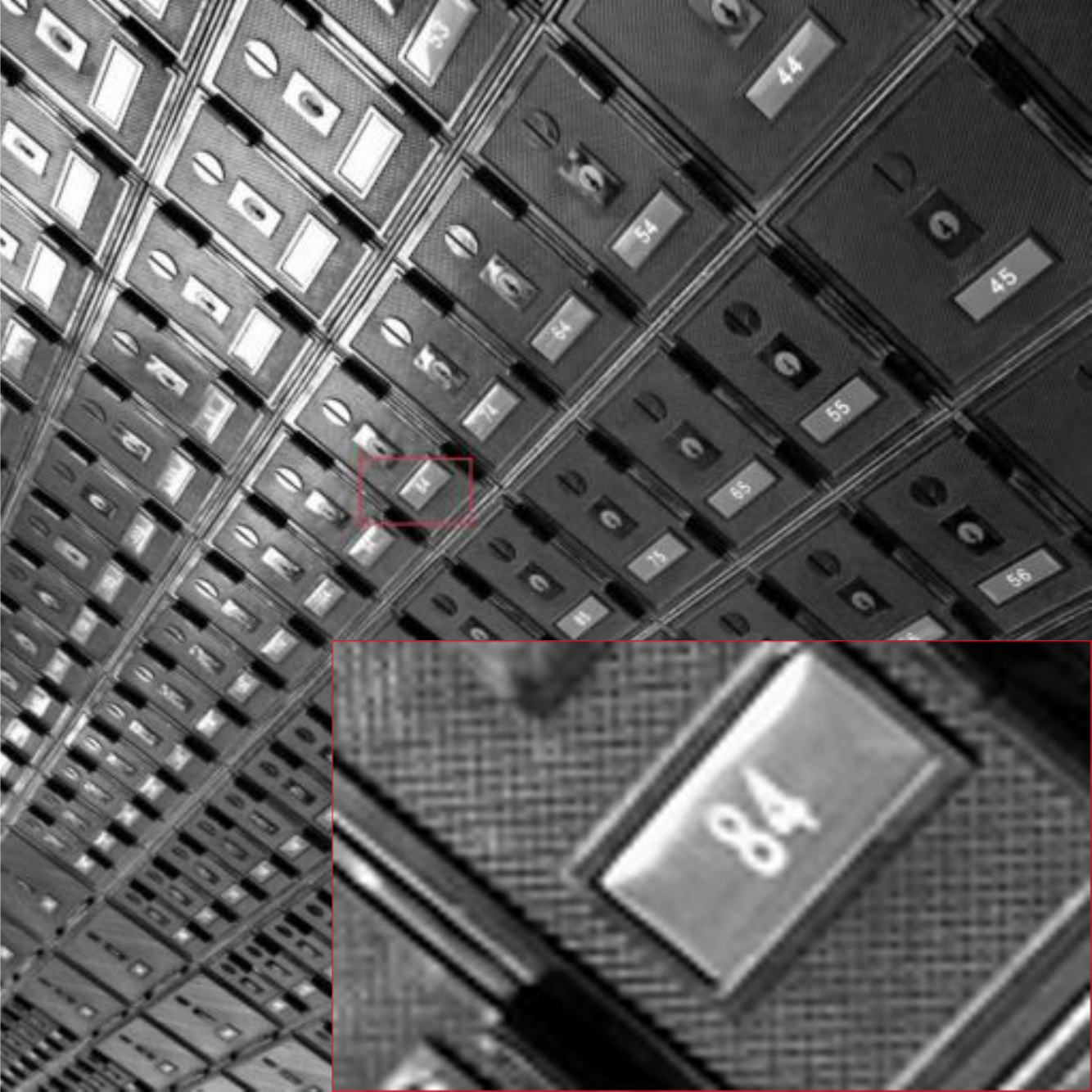}}
        \end{minipage}
        \begin{minipage}[b]{0.175\linewidth}
        	\subfloat[Halftone]{\includegraphics[width=1\linewidth]{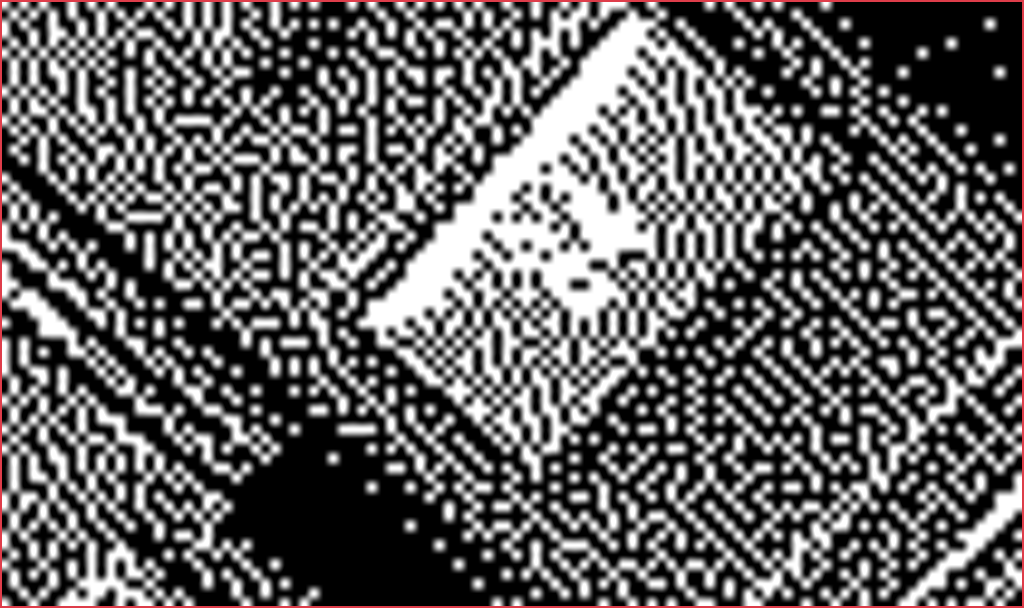}} \vspace{-0.8em}
        	\subfloat[PRL]{\includegraphics[width=1\linewidth]{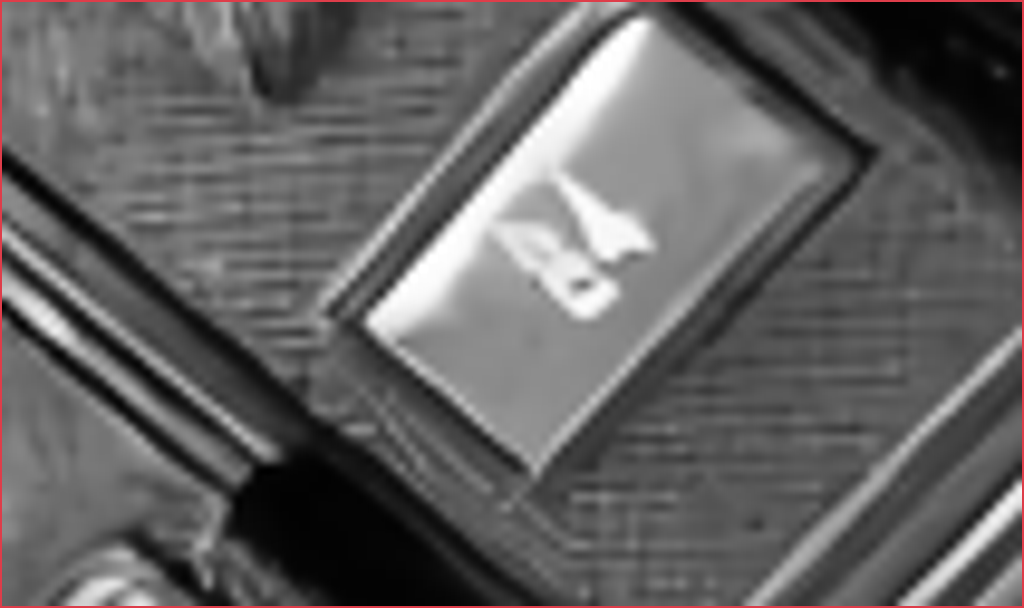}}
        \end{minipage}
        \begin{minipage}[b]{0.175\linewidth}
        	\subfloat[DnCNN]{\includegraphics[width=1\linewidth]{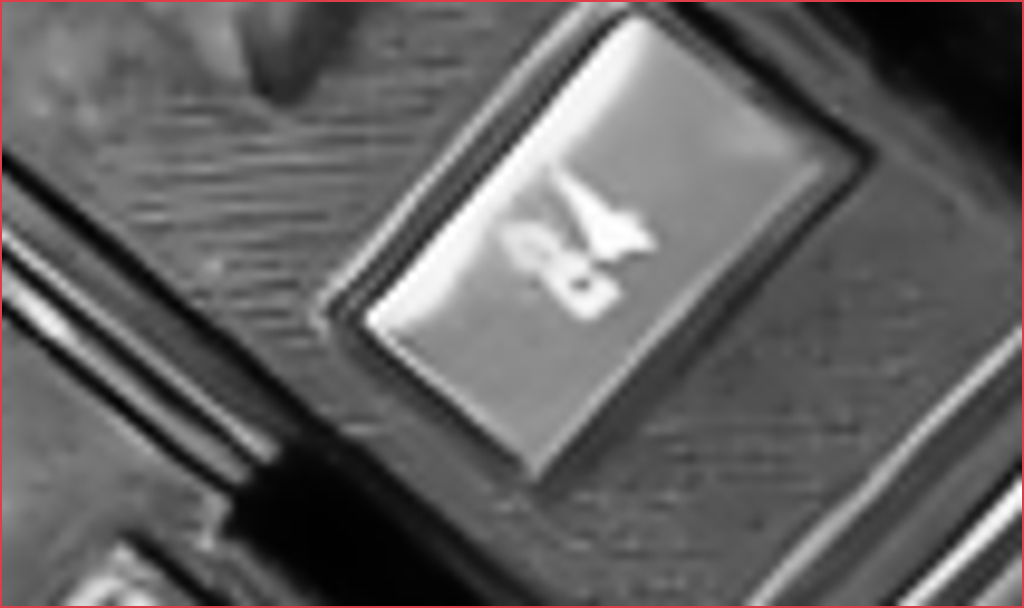}} \vspace{-0.8em}
        	\subfloat[MIMOUNet]{\includegraphics[width=1\linewidth]{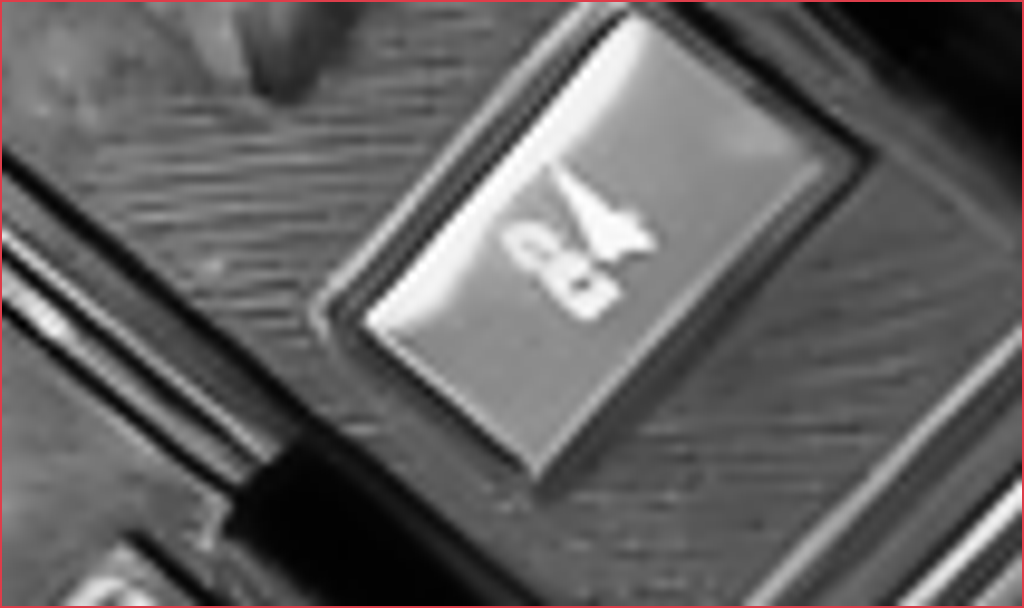}}
        \end{minipage}
        \begin{minipage}[b]{0.175\linewidth}
        	\subfloat[VDSR]{\includegraphics[width=1\linewidth]{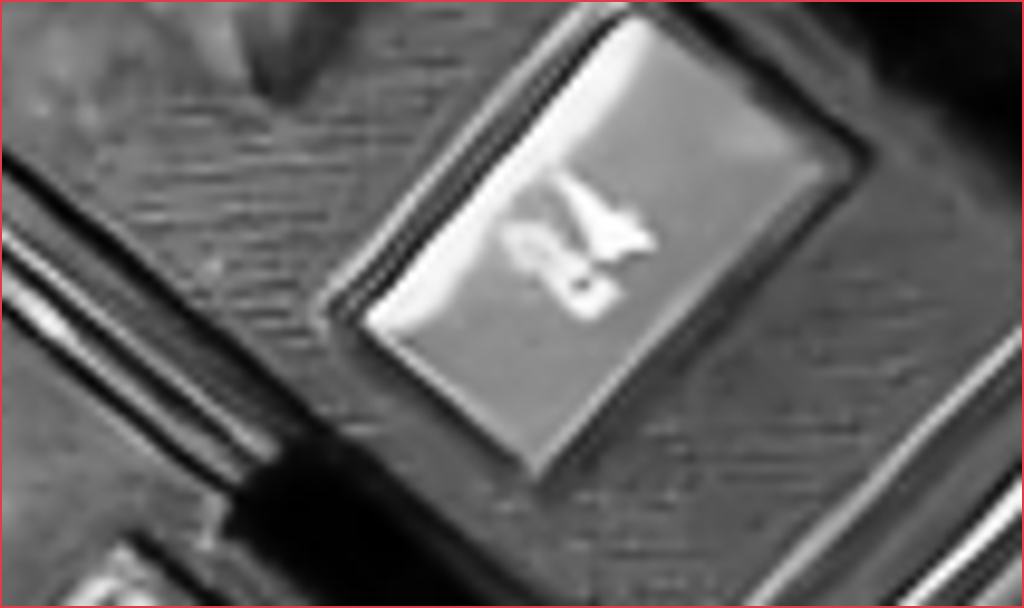}} \vspace{-0.8em}
        	\subfloat[PRL-dt (ours)]{\includegraphics[width=1\linewidth]{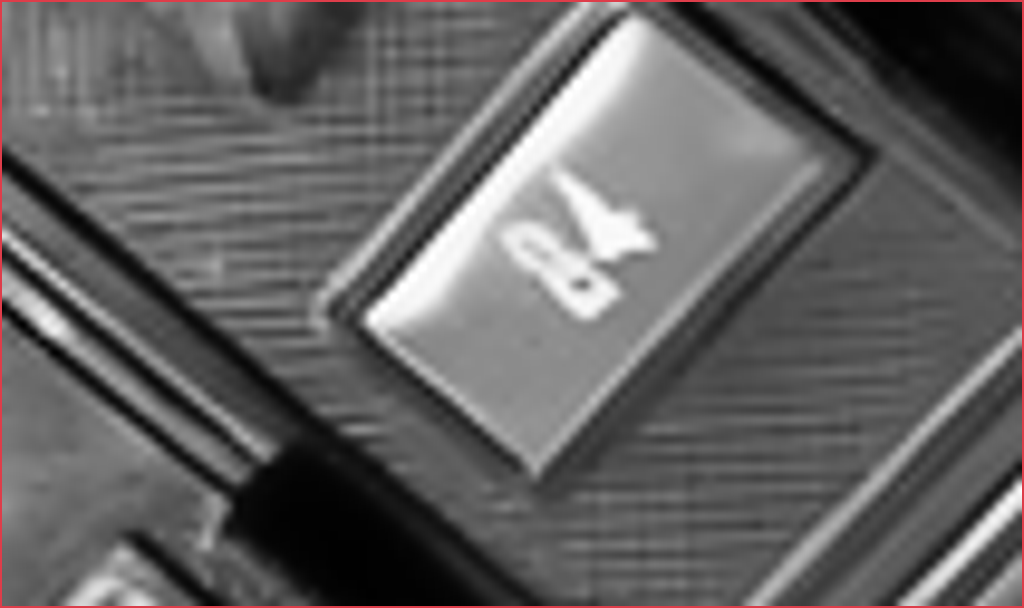}}
        \end{minipage}
        \begin{minipage}[b]{0.175\linewidth}
        	\subfloat[EDSR]{\includegraphics[width=1\linewidth]{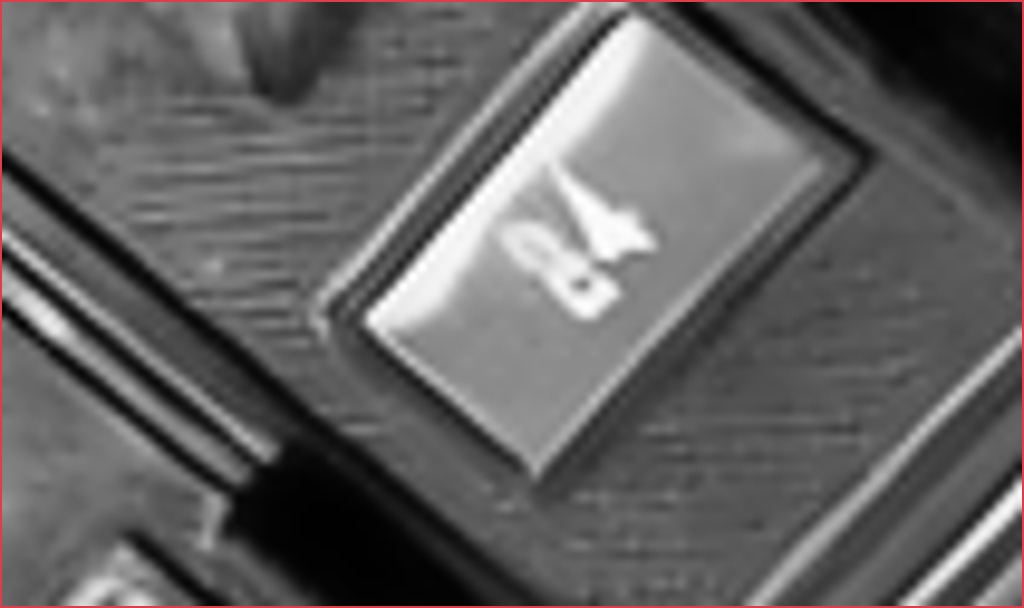}} \vspace{-0.8em}
        	\subfloat[MSPRL (ours)]{\includegraphics[width=1\linewidth]{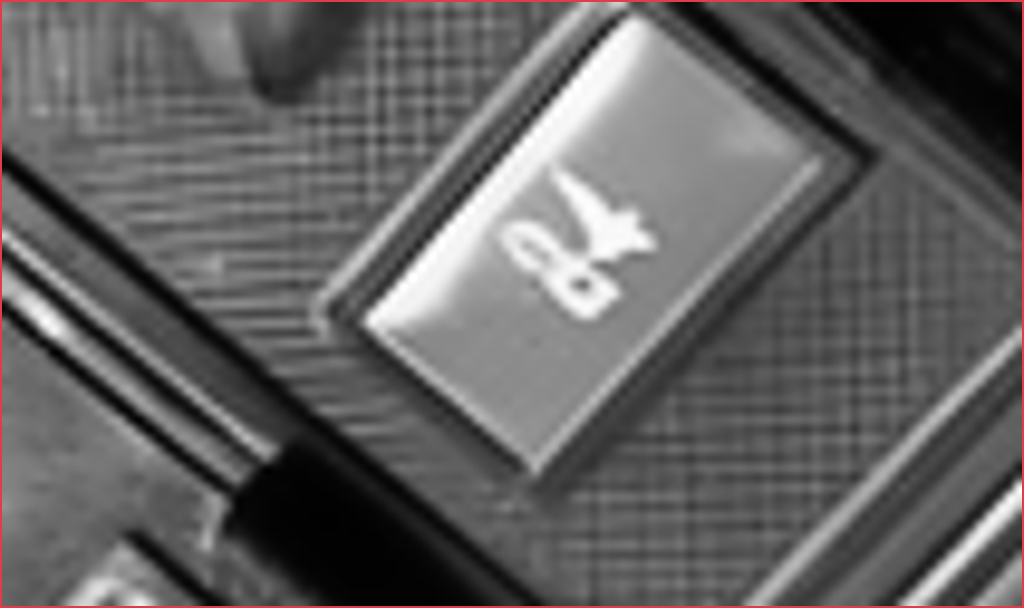}}
        \end{minipage}
        \\ \vspace{0.5em}
        % img_026
        \begin{minipage}[b]{0.255\linewidth}
        	\subfloat[Urban100: img\_026]{\includegraphics[width=1\linewidth]{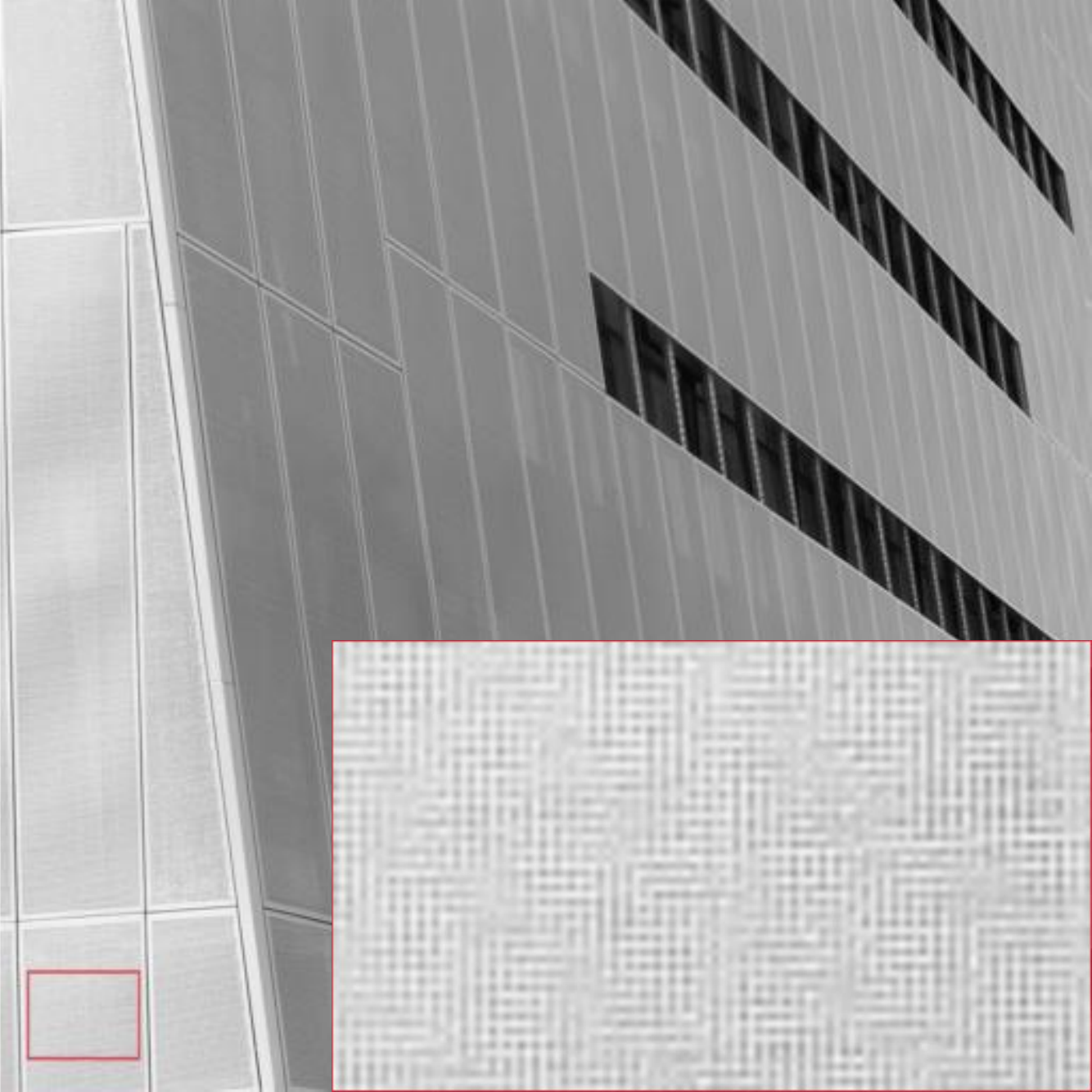}}
        \end{minipage}
        \begin{minipage}[b]{0.175\linewidth}
        	\subfloat[Halftone]{\includegraphics[width=1\linewidth]{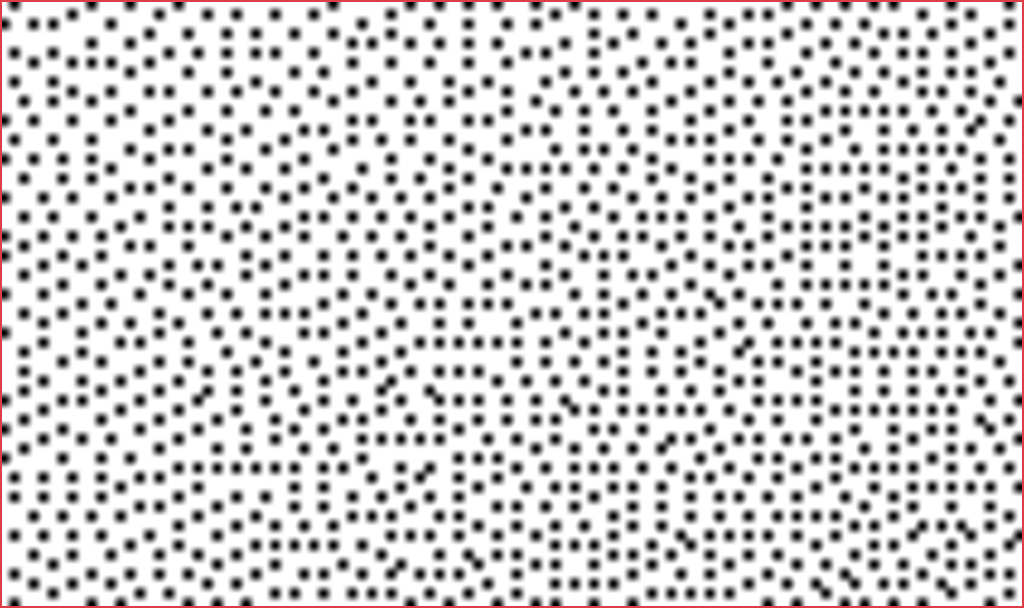}} \vspace{-0.8em}
        	\subfloat[PRL]{\includegraphics[width=1\linewidth]{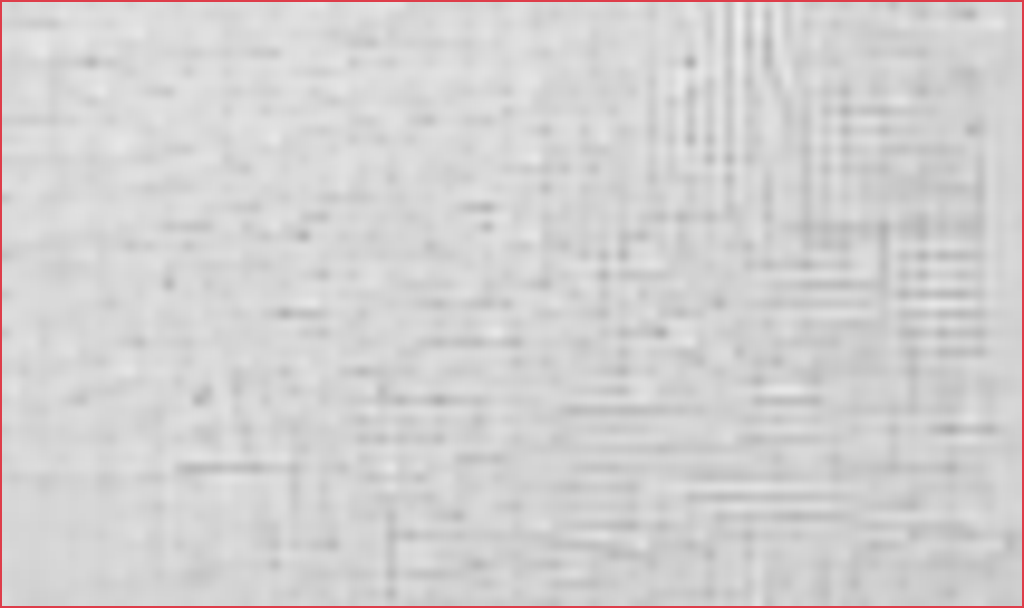}}
        \end{minipage}
        \begin{minipage}[b]{0.175\linewidth}
        	\subfloat[DnCNN]{\includegraphics[width=1\linewidth]{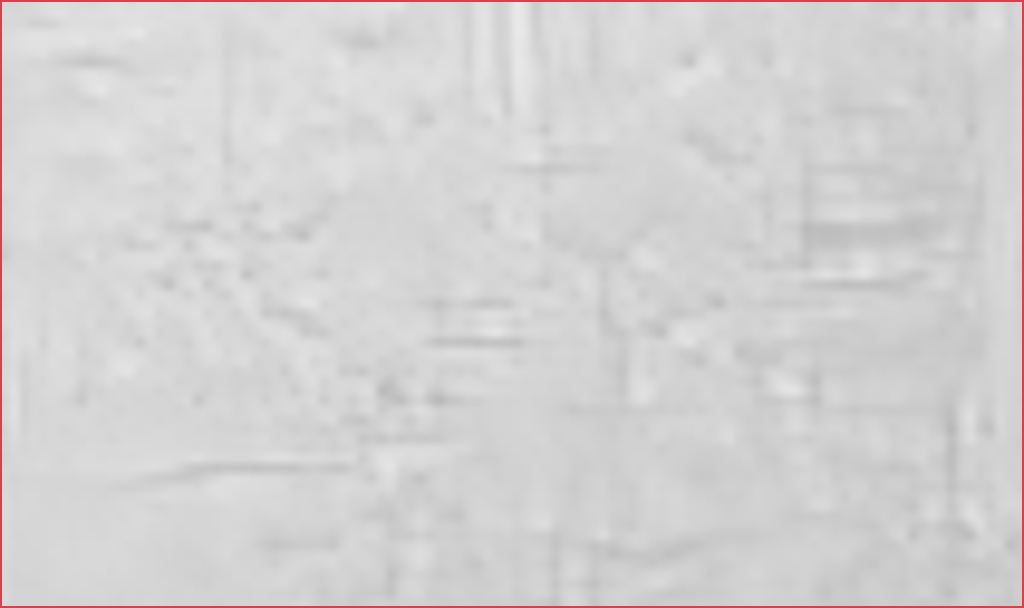}} \vspace{-0.8em}
        	\subfloat[MIMOUNet]{\includegraphics[width=1\linewidth]{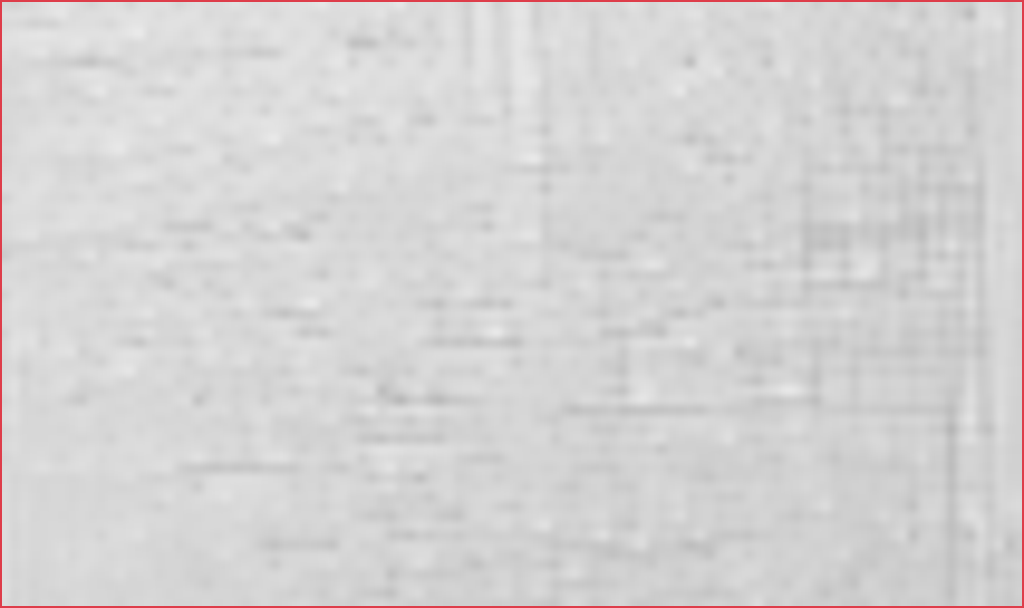}}
        \end{minipage}
        \begin{minipage}[b]{0.175\linewidth}
        	\subfloat[VDSR]{\includegraphics[width=1\linewidth]{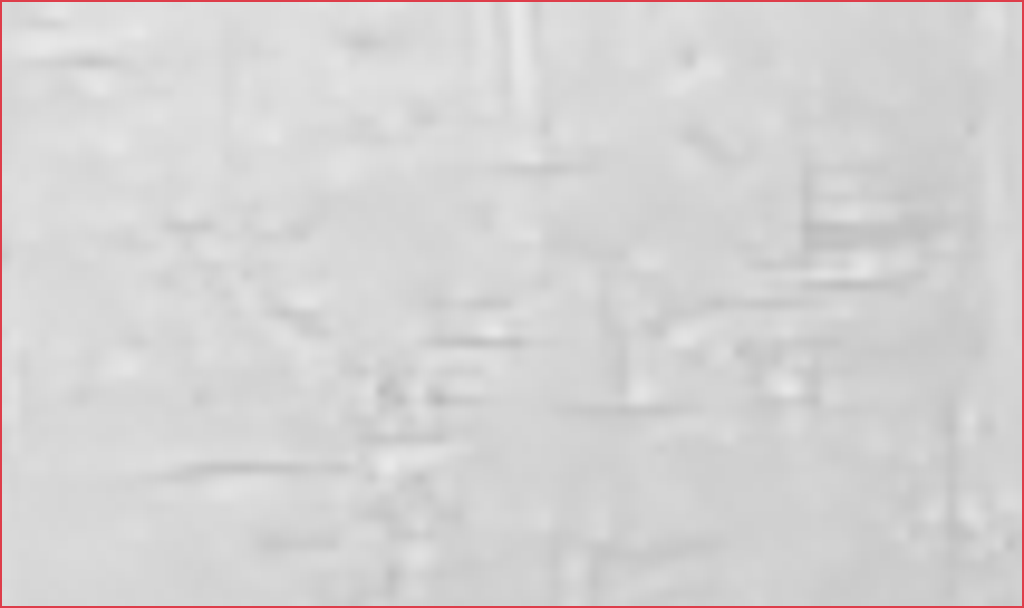}} \vspace{-0.8em}
        	\subfloat[PRL-dt (ours)]{\includegraphics[width=1\linewidth]{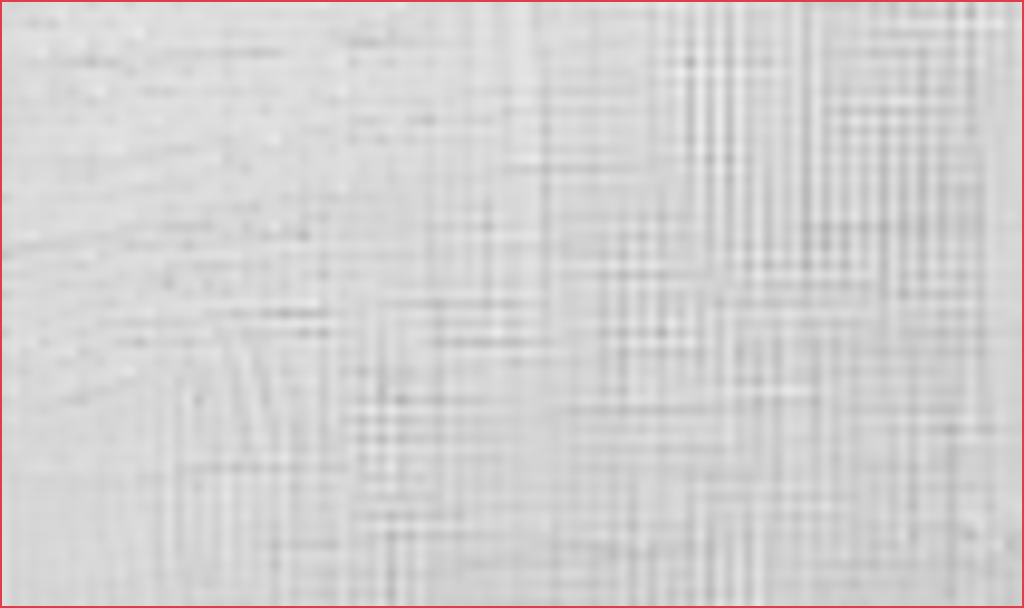}}
        \end{minipage}
        \begin{minipage}[b]{0.175\linewidth}
        	\subfloat[EDSR]{\includegraphics[width=1\linewidth]{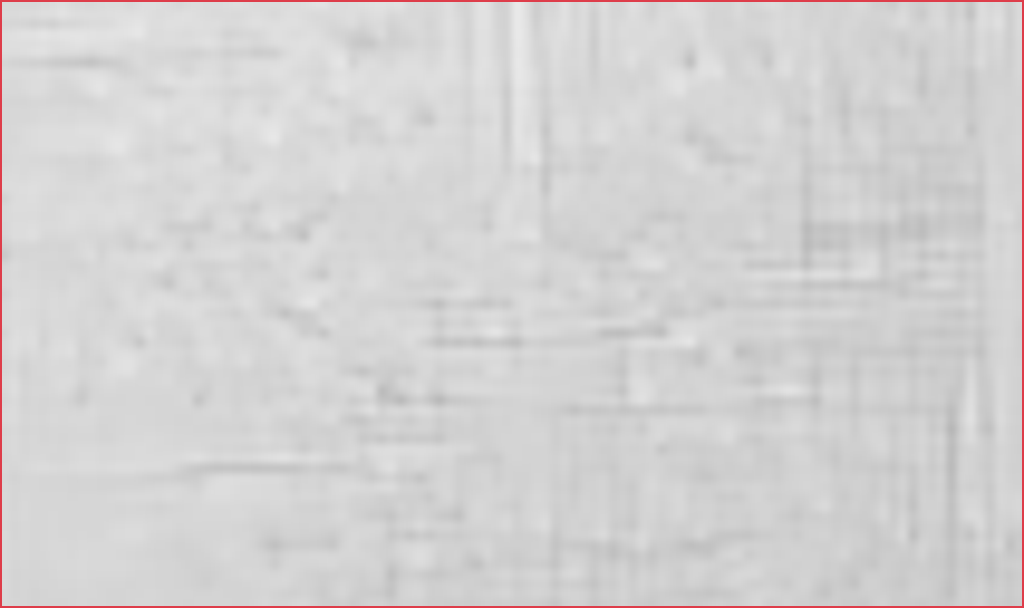}} \vspace{-0.8em}
        	\subfloat[MSPRL (ours)]{\includegraphics[width=1\linewidth]{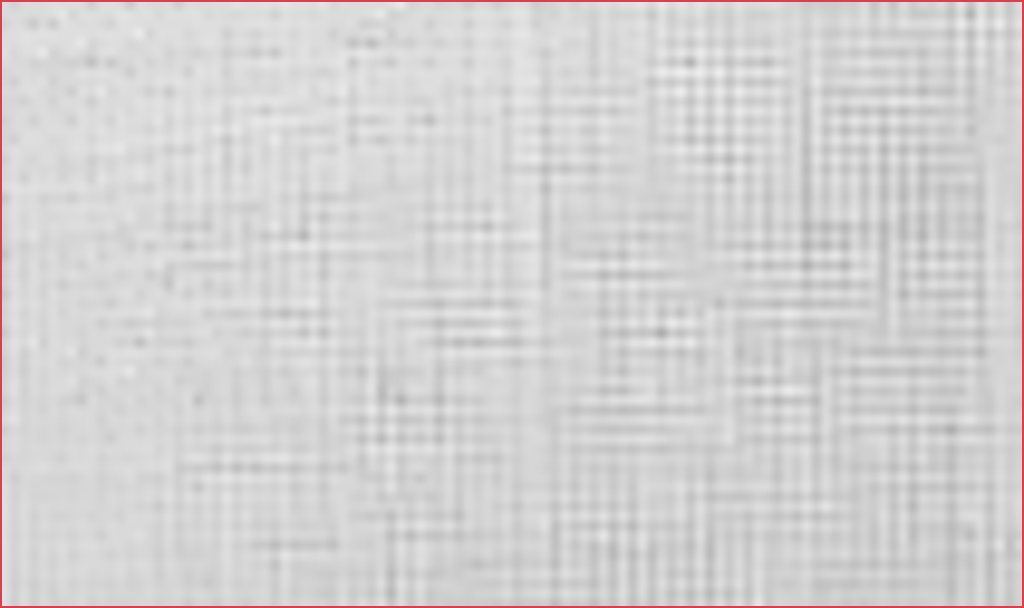}}
        \end{minipage}
        \\ \vspace{0.5em}
        % img_019
        \begin{minipage}[b]{0.255\linewidth}
        	\subfloat[Urban100: img\_019]{\includegraphics[width=1\linewidth]{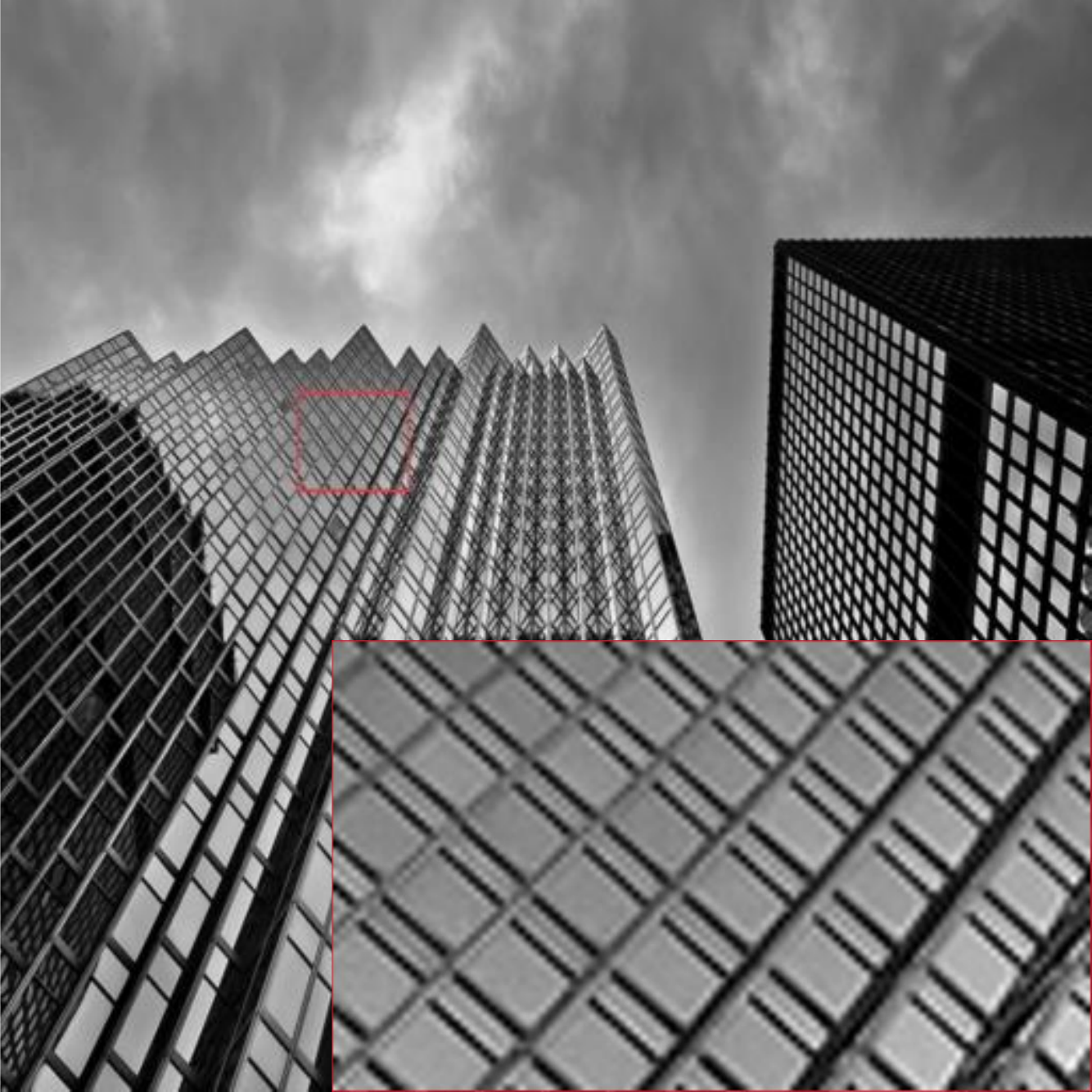}}
        \end{minipage}
        \begin{minipage}[b]{0.175\linewidth}
        	\subfloat[Halftone]{\includegraphics[width=1\linewidth]{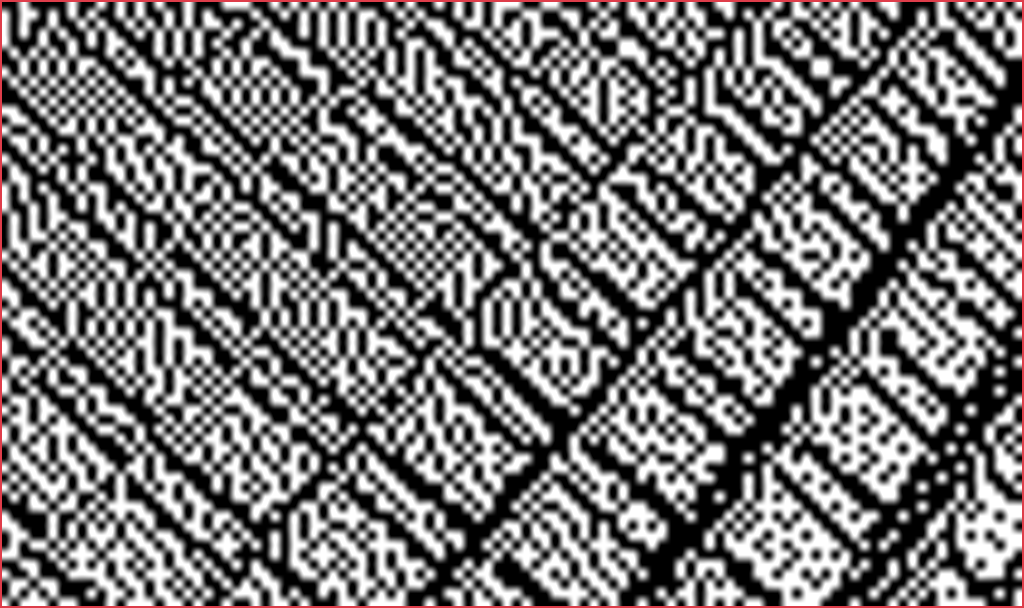}} \vspace{-0.8em}
        	\subfloat[PRL]{\includegraphics[width=1\linewidth]{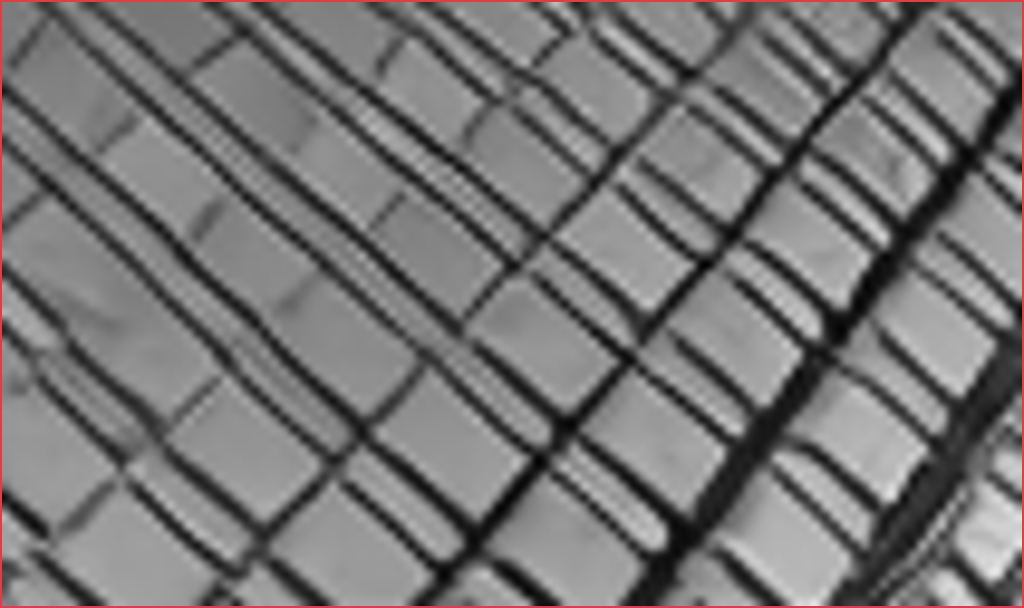}}
        \end{minipage}
        \begin{minipage}[b]{0.175\linewidth}
        	\subfloat[DnCNN]{\includegraphics[width=1\linewidth]{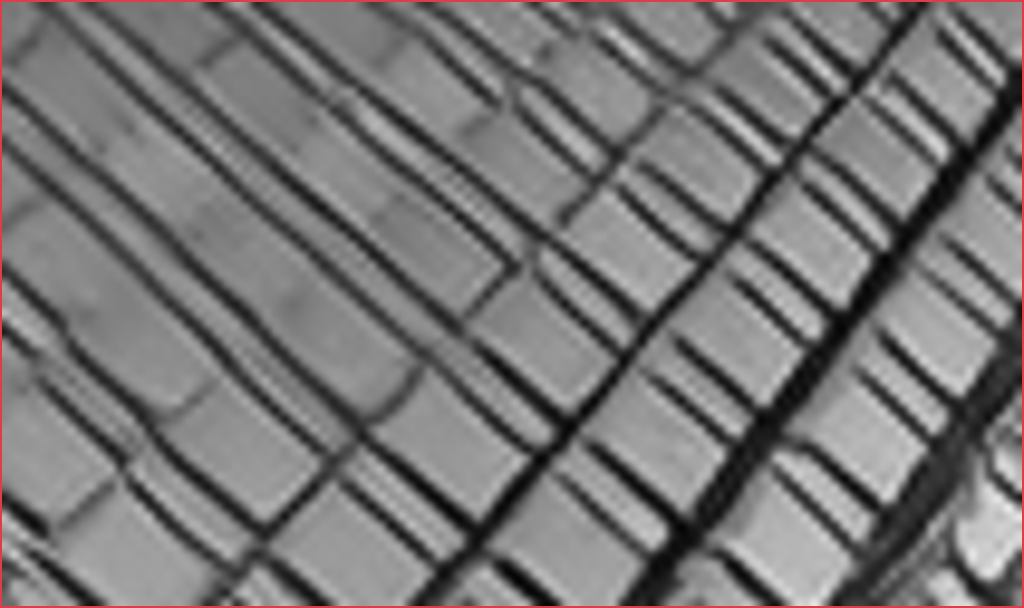}} \vspace{-0.8em}
        	\subfloat[MIMOUNet]{\includegraphics[width=1\linewidth]{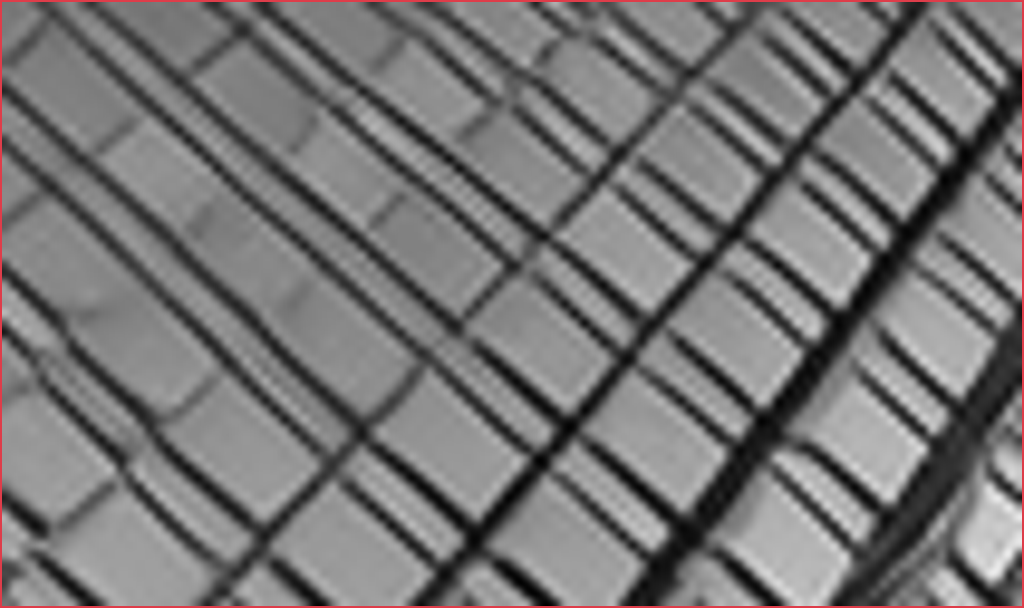}}
        \end{minipage}
        \begin{minipage}[b]{0.175\linewidth}
        	\subfloat[VDSR]{\includegraphics[width=1\linewidth]{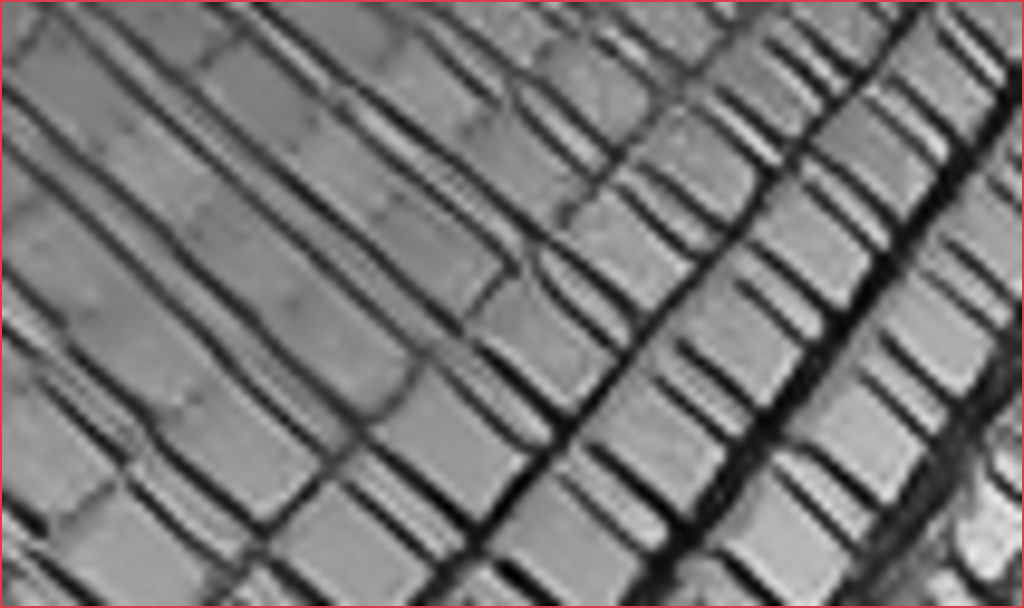}} \vspace{-0.8em}
        	\subfloat[PRL-dt (ours)]{\includegraphics[width=1\linewidth]{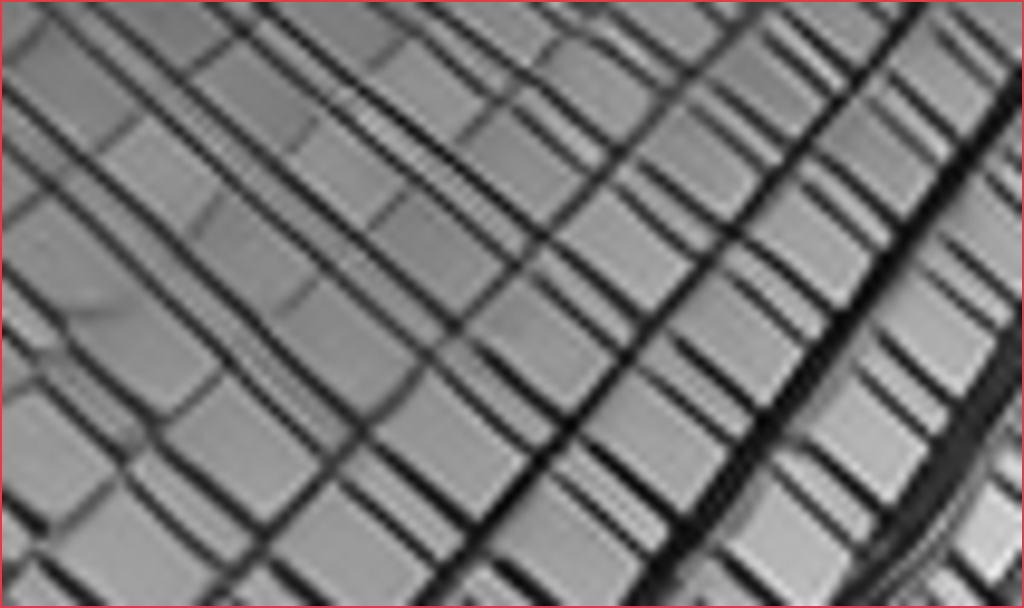}}
        \end{minipage}
        \begin{minipage}[b]{0.175\linewidth}
        	\subfloat[EDSR]{\includegraphics[width=1\linewidth]{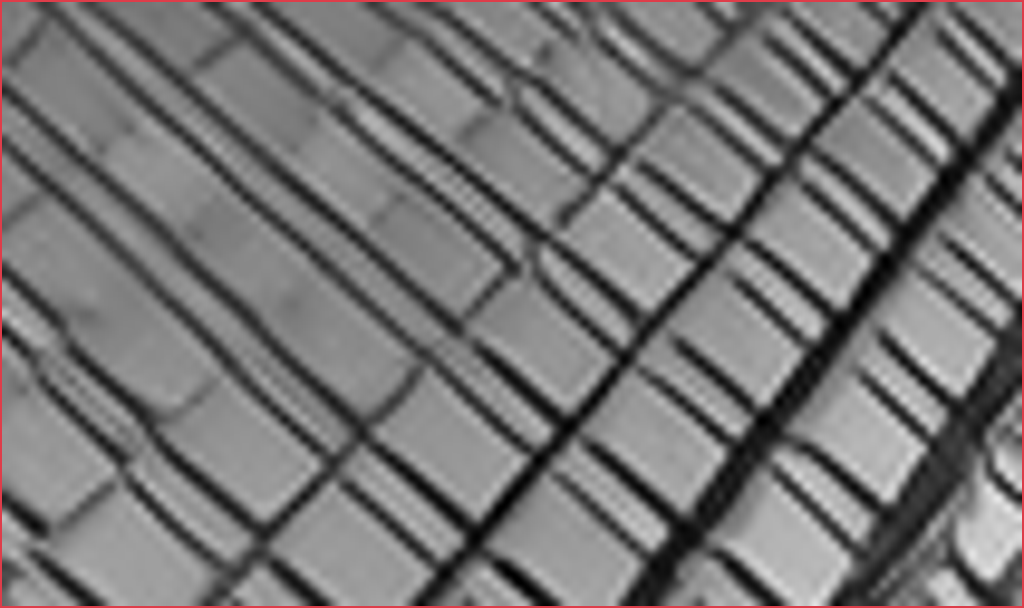}} \vspace{-0.8em}
        	\subfloat[MSPRL (ours)]{\includegraphics[width=1\linewidth]{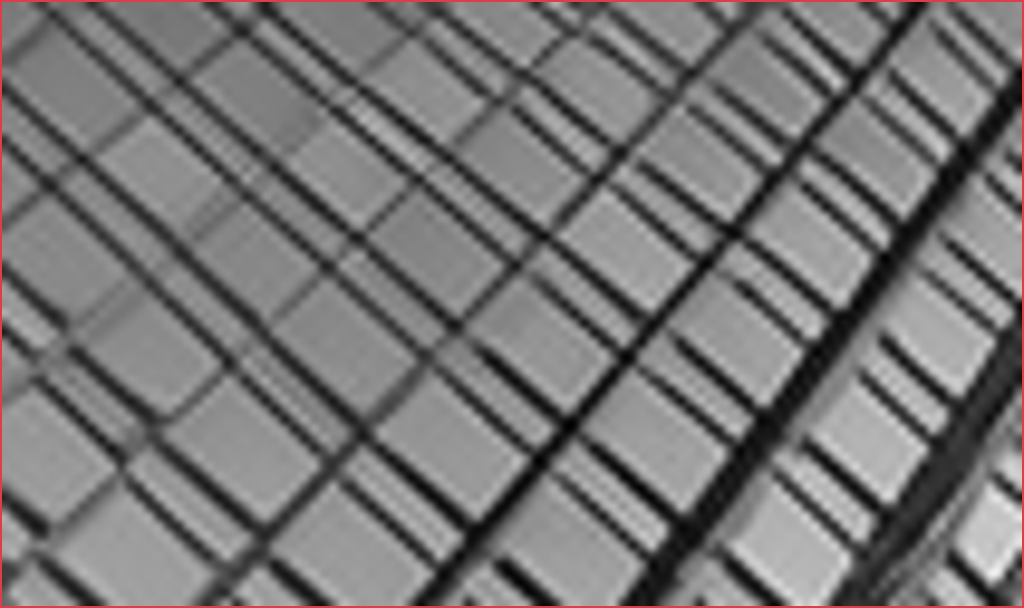}}
        \end{minipage}
        \\ \vspace{0.5em}
        % img_033
        \begin{minipage}[b]{0.255\linewidth}
        	\subfloat[Urban100: img\_033]{\includegraphics[width=1\linewidth]{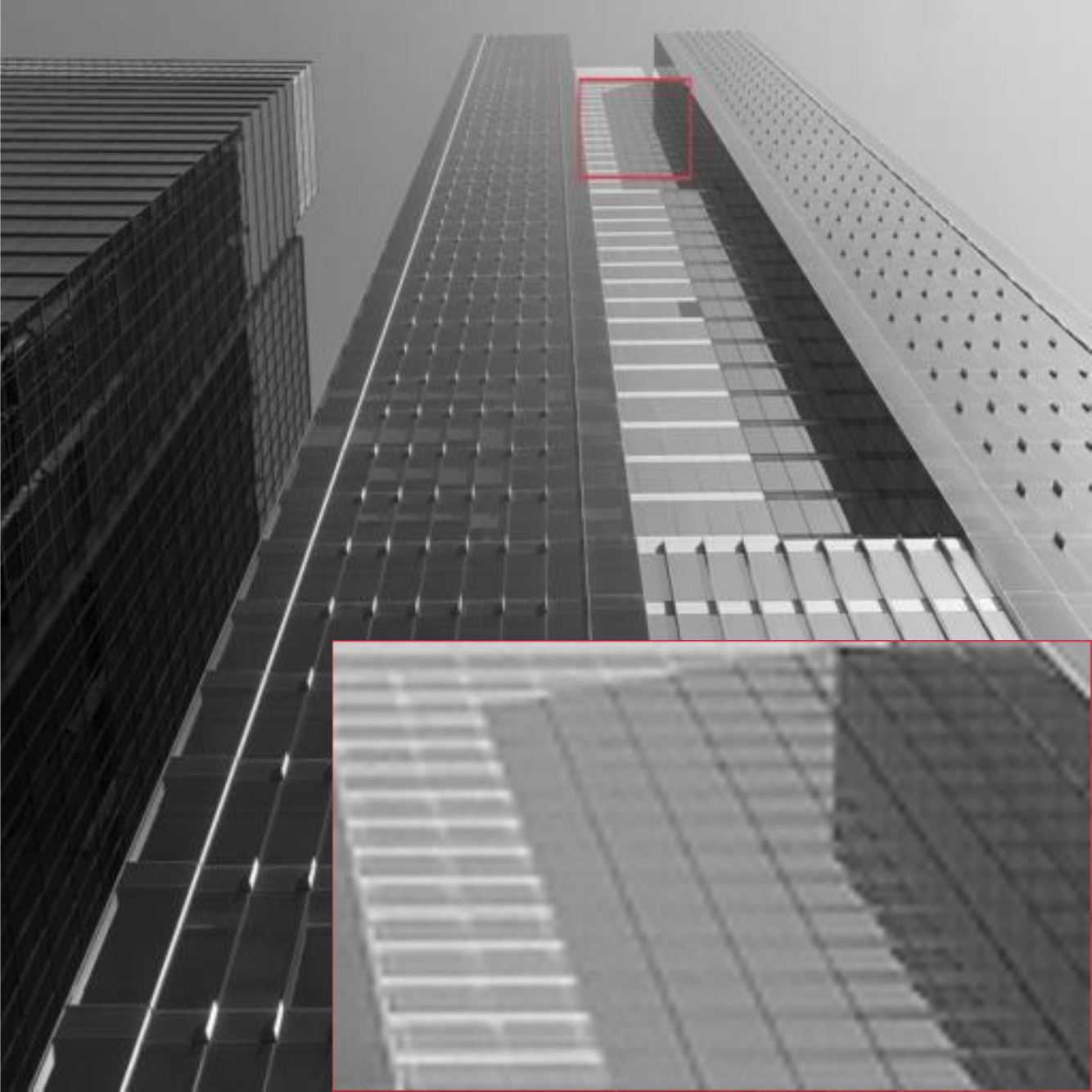}}
        \end{minipage}
        \begin{minipage}[b]{0.175\linewidth}
        	\subfloat[Halftone]{\includegraphics[width=1\linewidth]{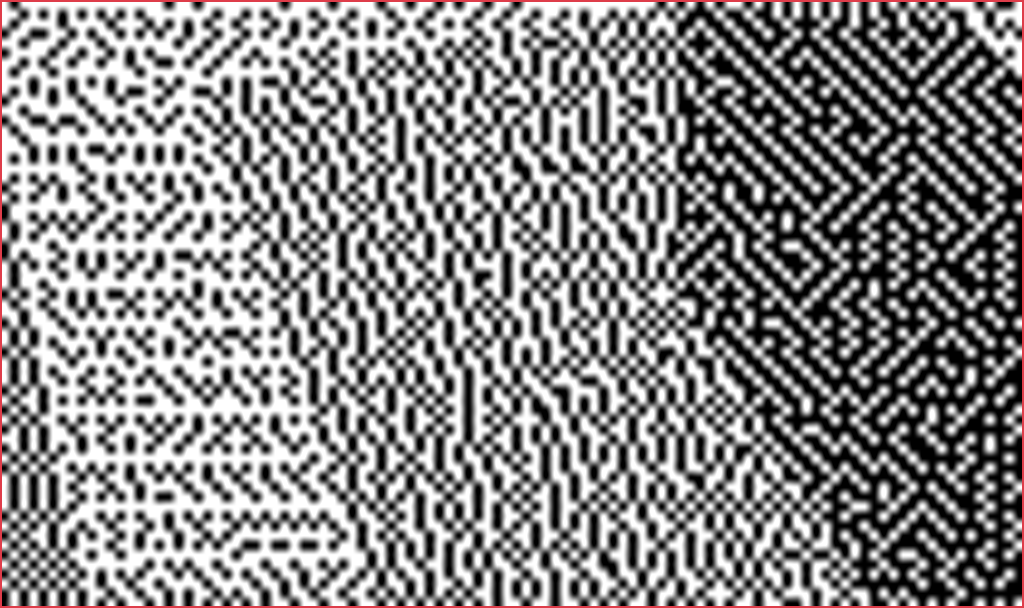}} \vspace{-0.8em}
        	\subfloat[PRL]{\includegraphics[width=1\linewidth]{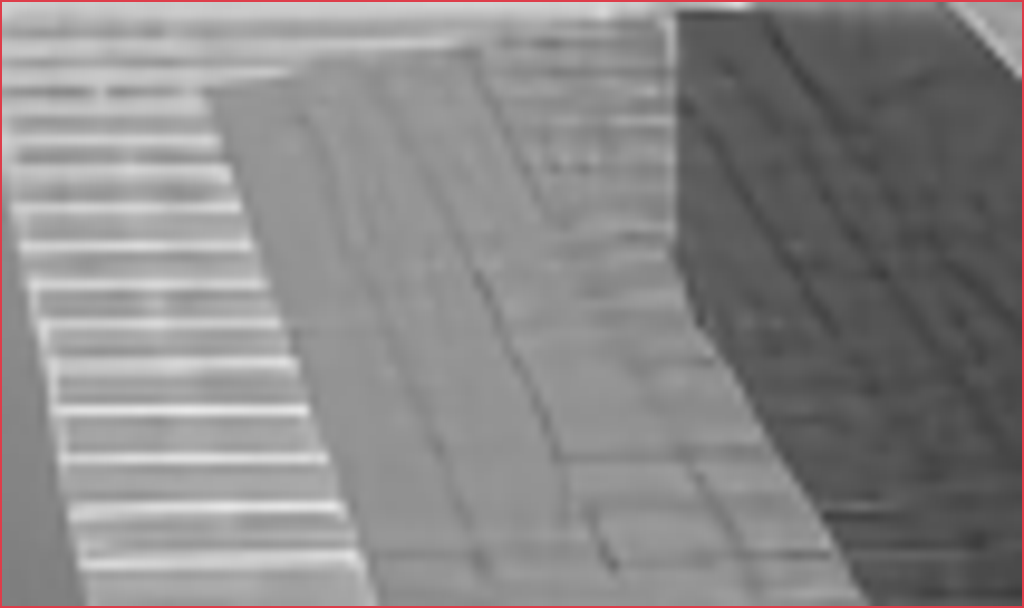}}
        \end{minipage}
        \begin{minipage}[b]{0.175\linewidth}
        	\subfloat[DnCNN]{\includegraphics[width=1\linewidth]{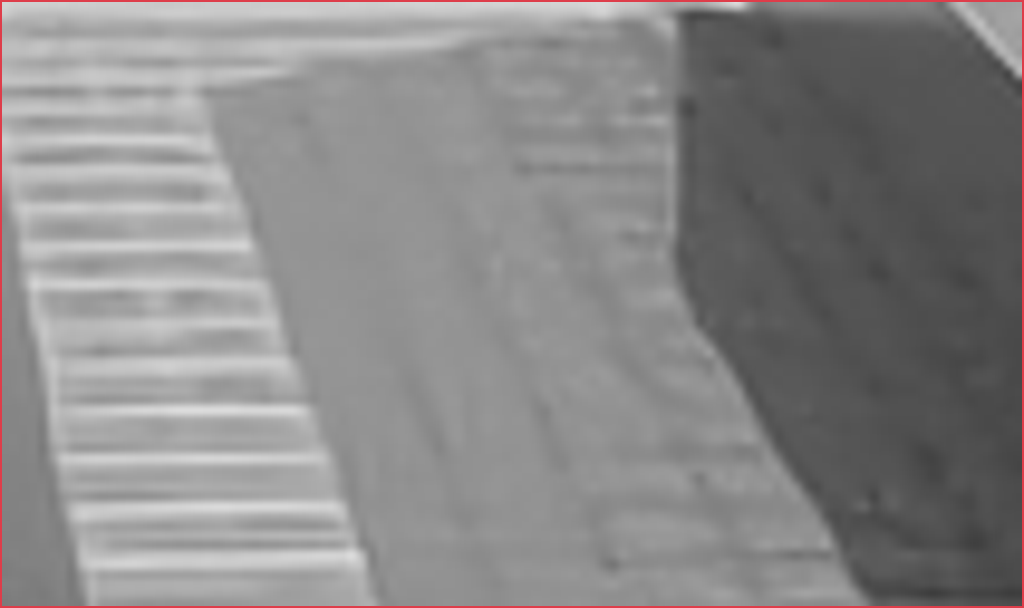}} \vspace{-0.8em}
        	\subfloat[MIMOUNet]{\includegraphics[width=1\linewidth]{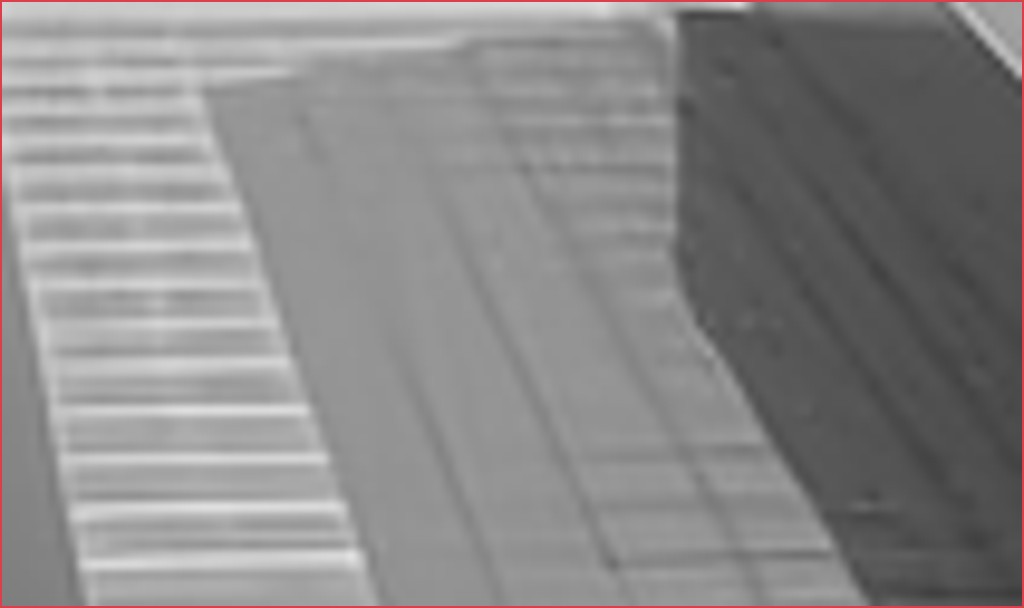}}
        \end{minipage}
        \begin{minipage}[b]{0.175\linewidth}
        	\subfloat[VDSR]{\includegraphics[width=1\linewidth]{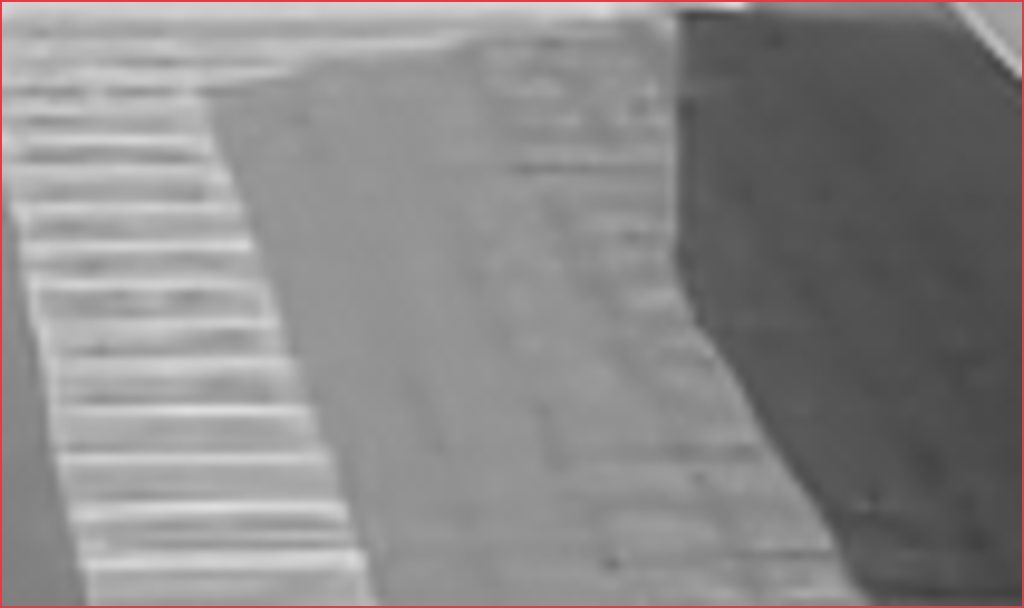}} \vspace{-0.8em}
        	\subfloat[PRL-dt (ours)]{\includegraphics[width=1\linewidth]{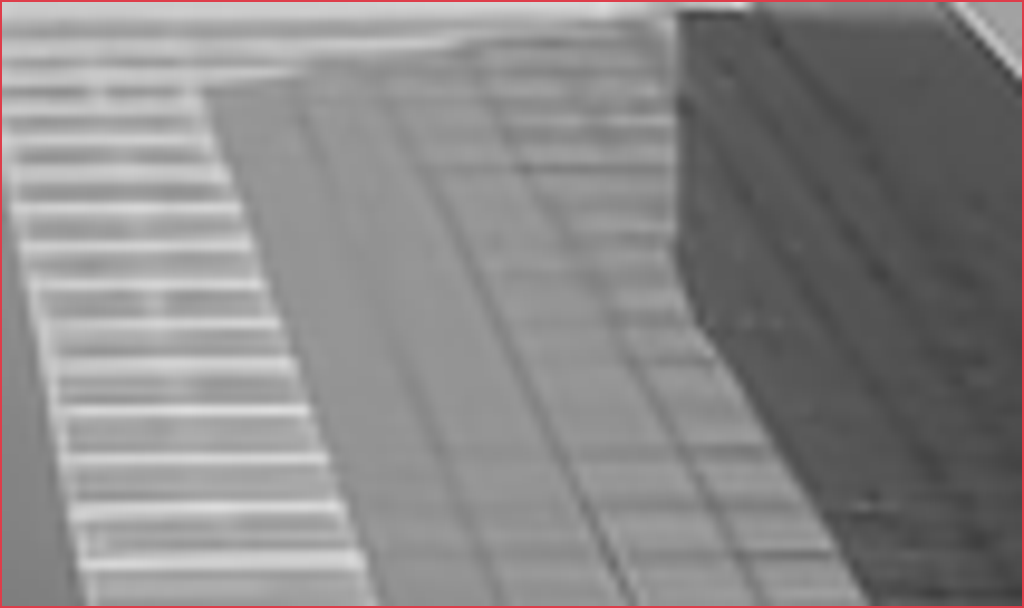}}
        \end{minipage}
        \begin{minipage}[b]{0.175\linewidth}
        	\subfloat[EDSR]{\includegraphics[width=1\linewidth]{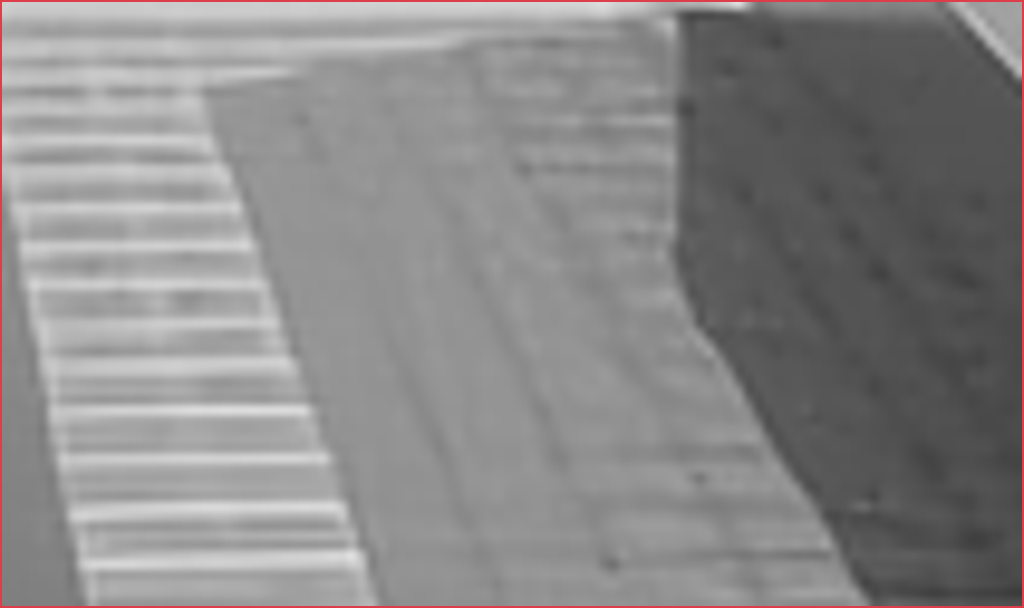}} \vspace{-0.8em}
        	\subfloat[MSPRL (ours)]{\includegraphics[width=1\linewidth]{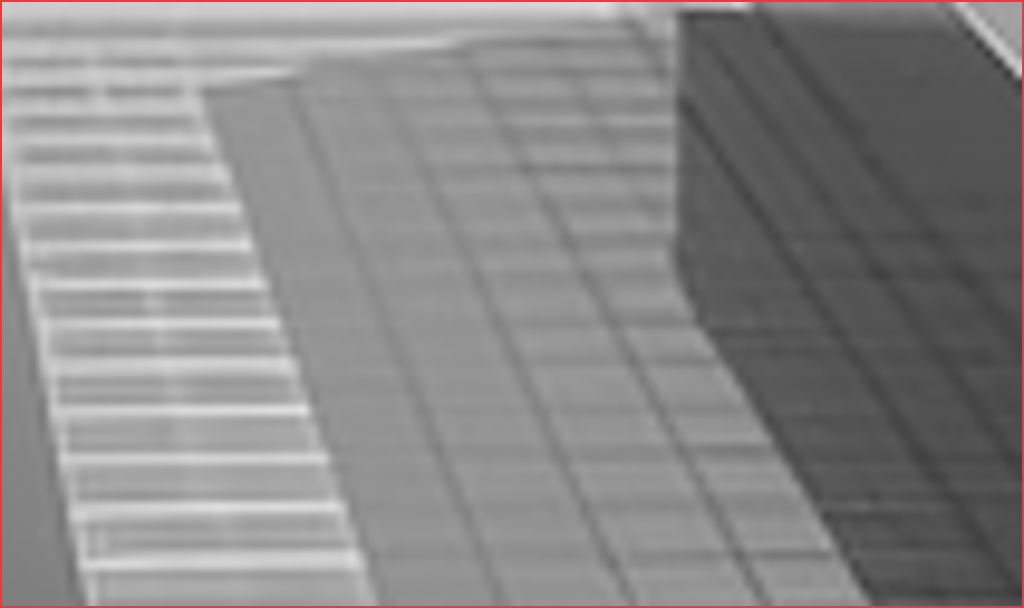}}
        \end{minipage}
        \end{center}
        \vspace{0.5em}
        \caption{Compared with the other approaches, our MSPRL more effectively restores the image details.}
    \label{fig:c3}
\end{figure*}

\begin{figure*}[ht]
\begin{center}
    \captionsetup[subfloat]{labelsep=none,format=plain,labelformat=empty,font={scriptsize}}
        % img_012
        \begin{minipage}[b]{0.255\linewidth}
        	\subfloat[Urban100: img\_012]{\includegraphics[width=1\linewidth]{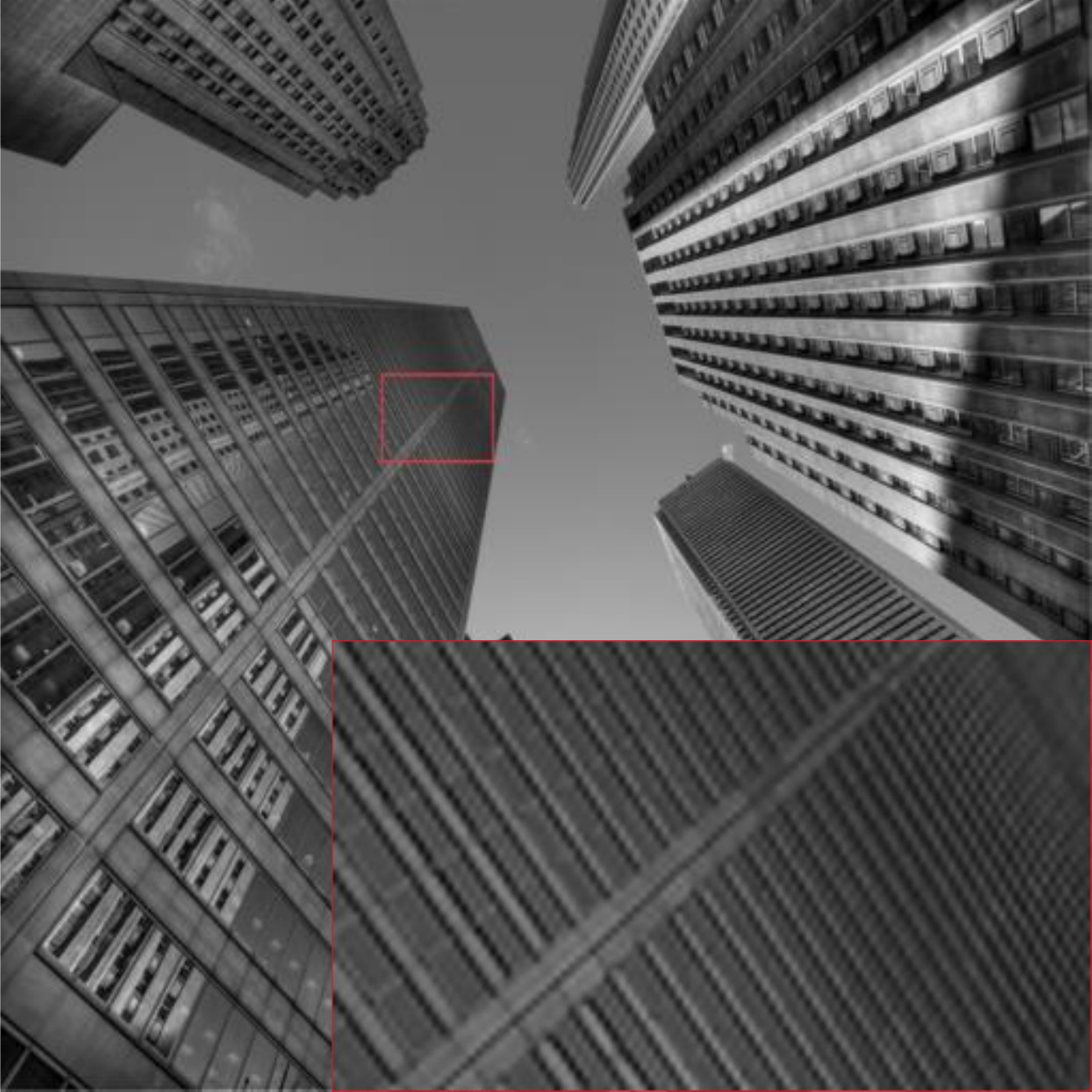}}
        \end{minipage}
        \begin{minipage}[b]{0.175\linewidth}
        	\subfloat[Halftone]{\includegraphics[width=1\linewidth]{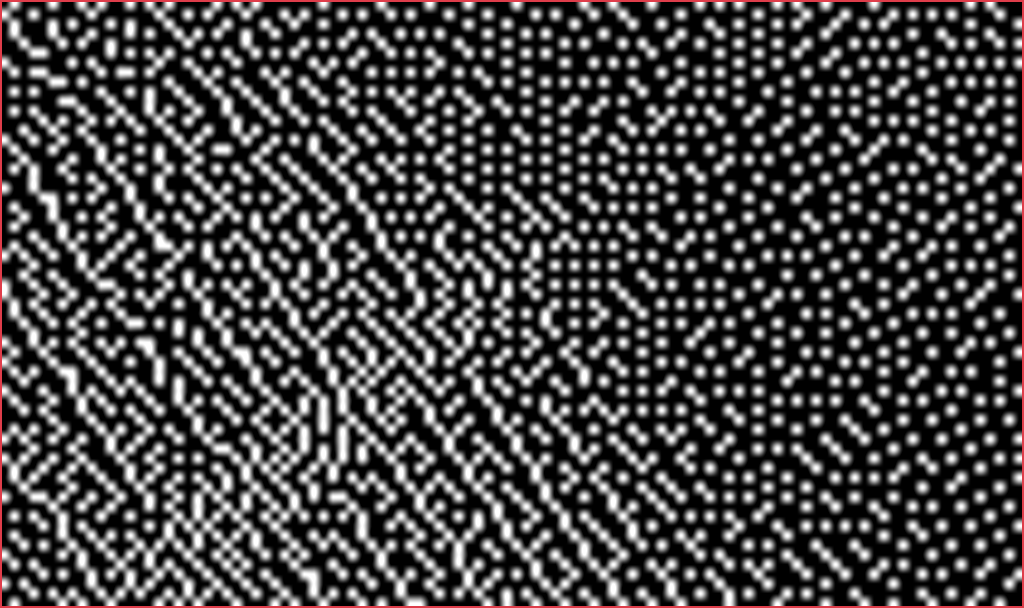}} \vspace{-0.8em}
        	\subfloat[PRL]{\includegraphics[width=1\linewidth]{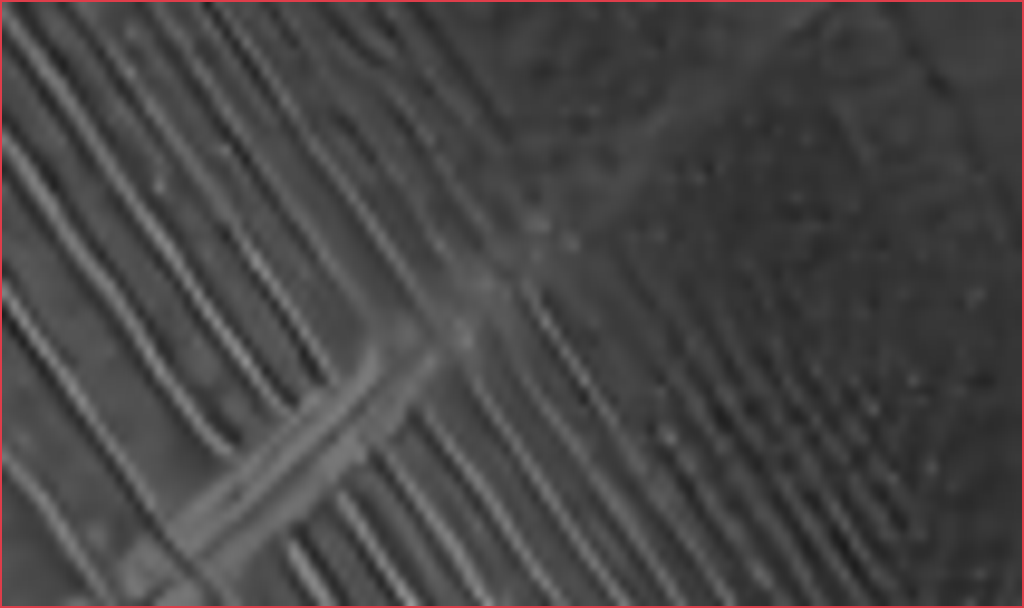}}
        \end{minipage}
        \begin{minipage}[b]{0.175\linewidth}
        	\subfloat[DnCNN]{\includegraphics[width=1\linewidth]{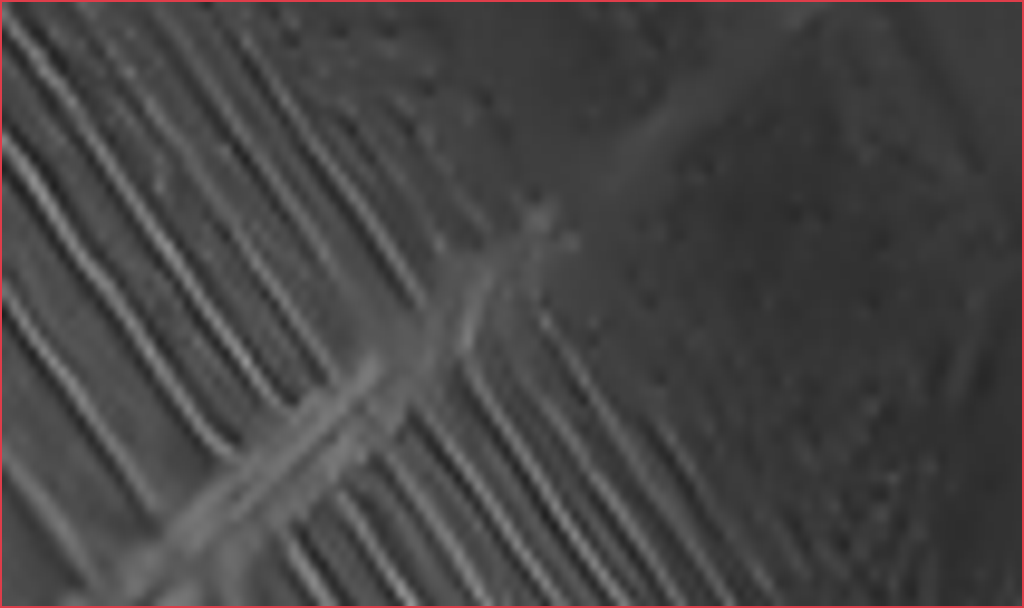}} \vspace{-0.8em}
        	\subfloat[MIMOUNet]{\includegraphics[width=1\linewidth]{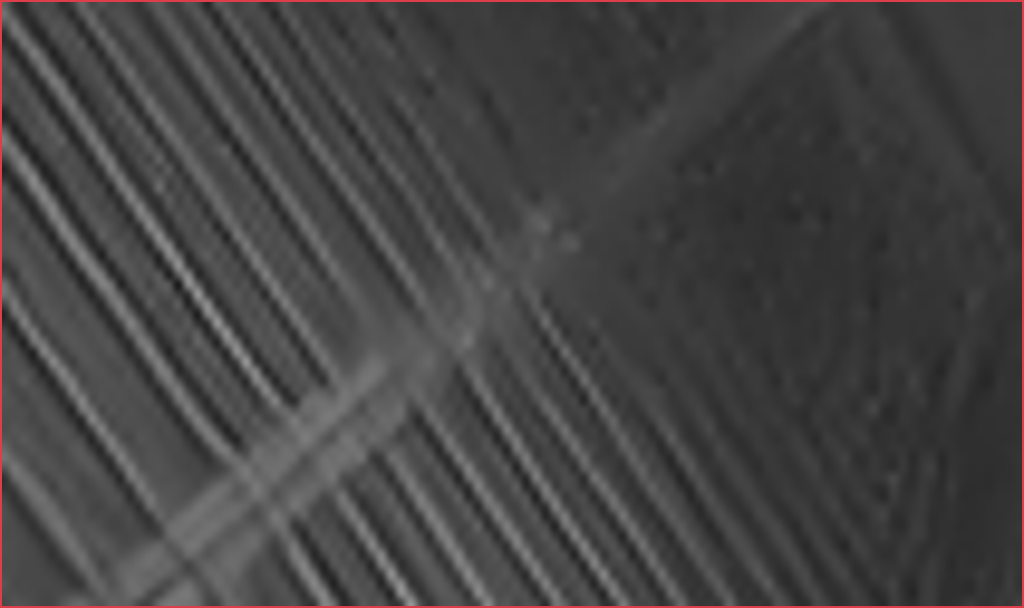}}
        \end{minipage}
        \begin{minipage}[b]{0.175\linewidth}
        	\subfloat[VDSR]{\includegraphics[width=1\linewidth]{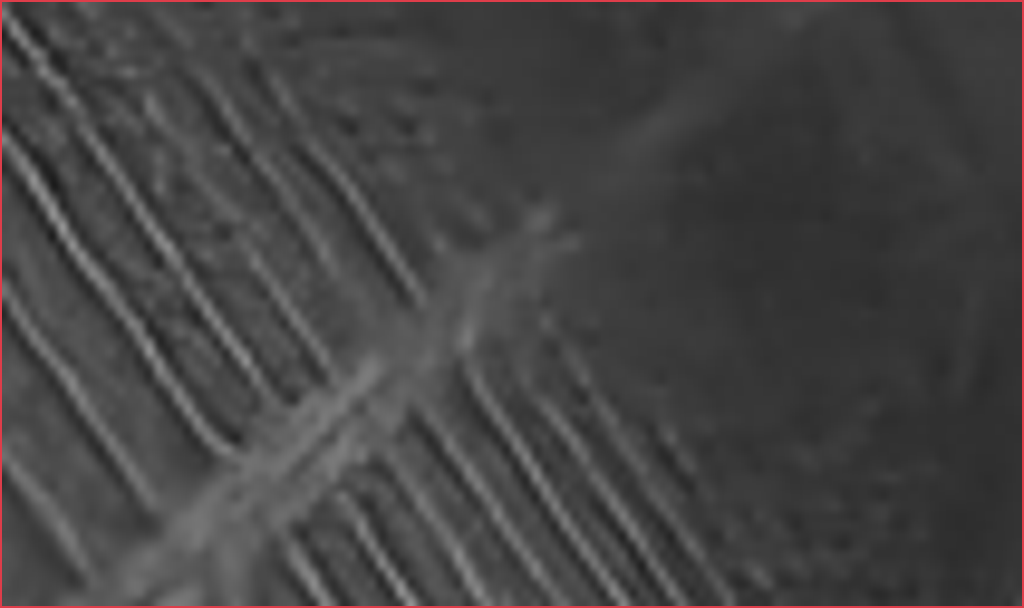}} \vspace{-0.8em}
        	\subfloat[PRL-dt (ours)]{\includegraphics[width=1\linewidth]{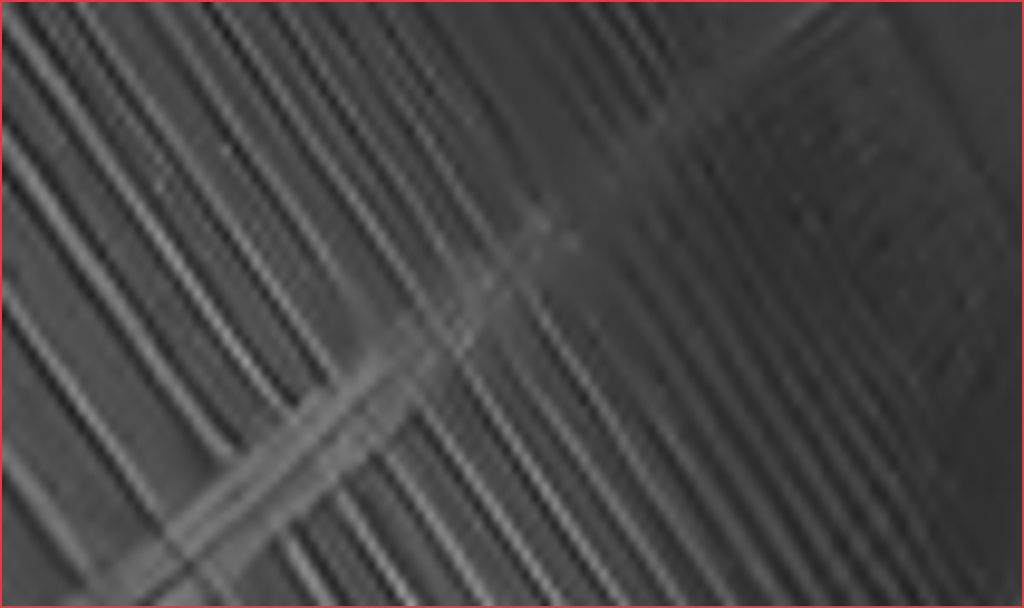}}
        \end{minipage}
        \begin{minipage}[b]{0.175\linewidth}
        	\subfloat[EDSR]{\includegraphics[width=1\linewidth]{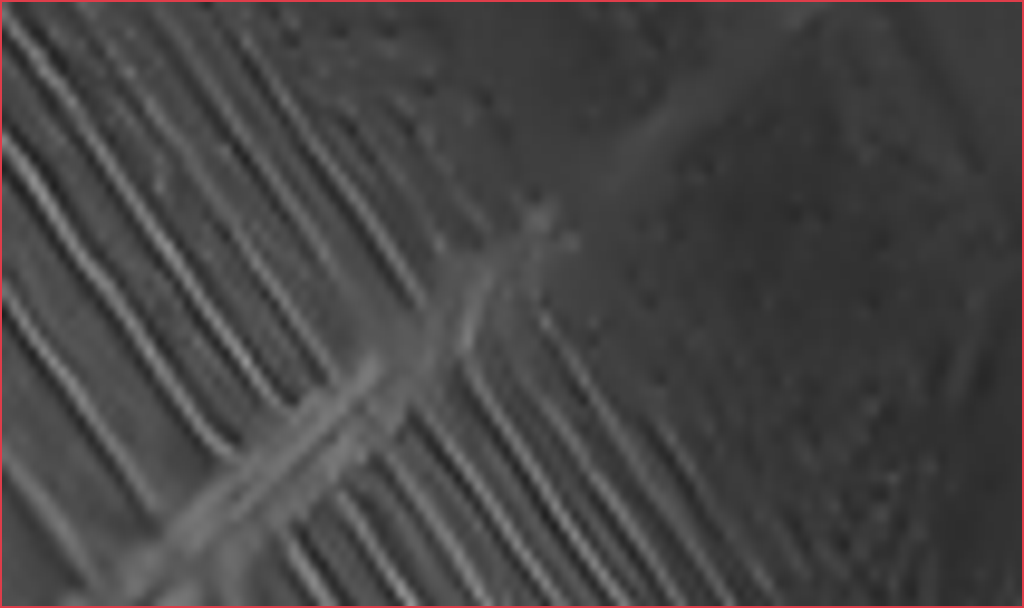}} \vspace{-0.8em}
        	\subfloat[MSPRL (ours)]{\includegraphics[width=1\linewidth]{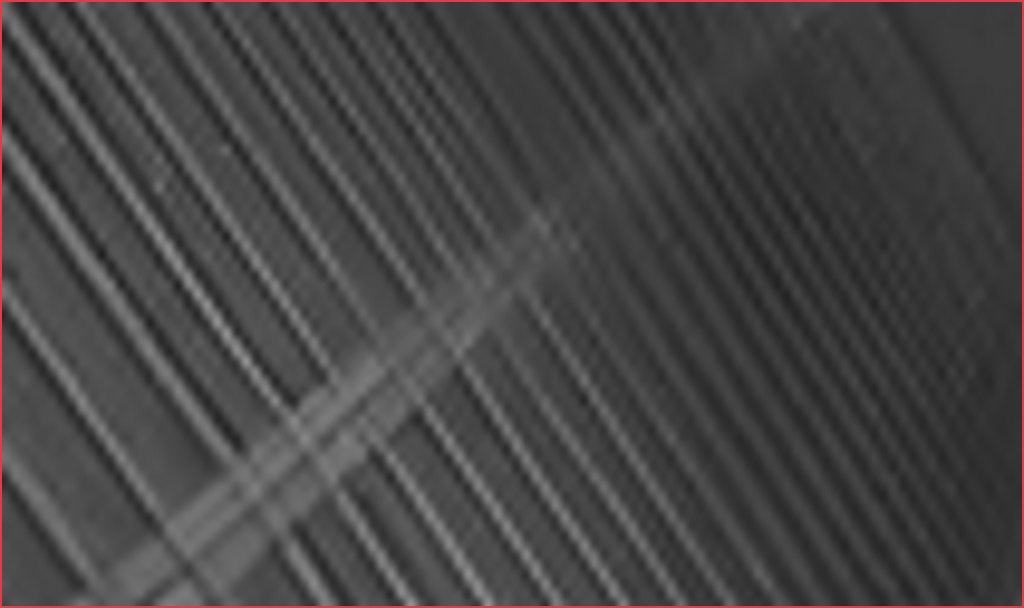}}
        \end{minipage}
        \\ \vspace{0.5em}
        % img_046
        \begin{minipage}[b]{0.255\linewidth}
        	\subfloat[Urban100: img\_046]{\includegraphics[width=1\linewidth]{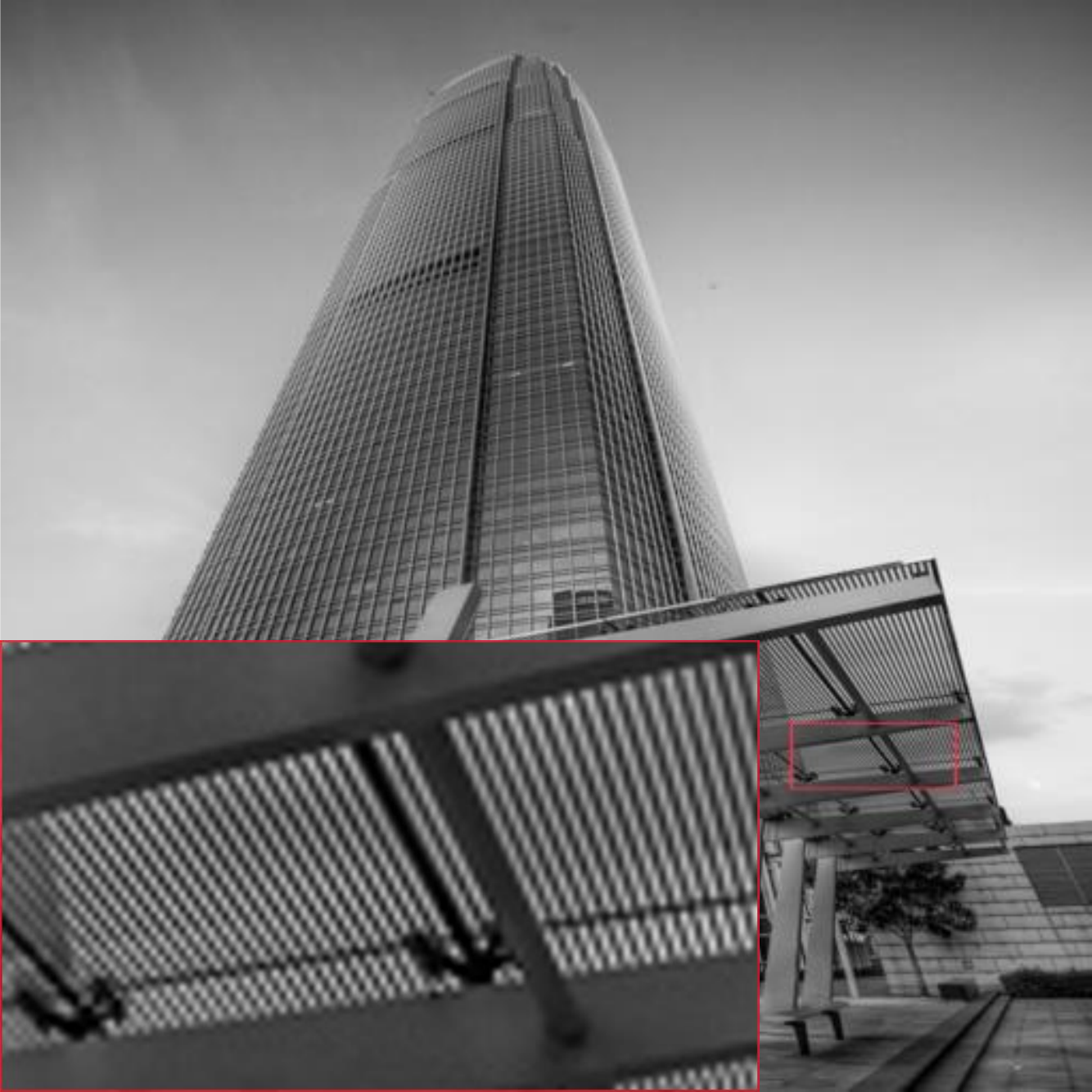}}
        \end{minipage}
        \begin{minipage}[b]{0.175\linewidth}
        	\subfloat[Halftone]{\includegraphics[width=1\linewidth]{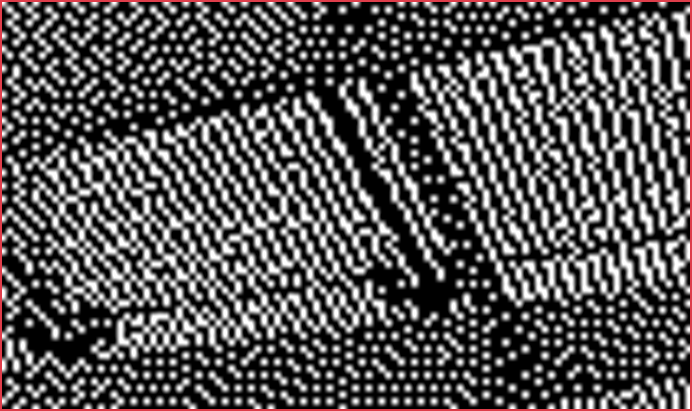}} \vspace{-0.8em}
        	\subfloat[PRL]{\includegraphics[width=1\linewidth]{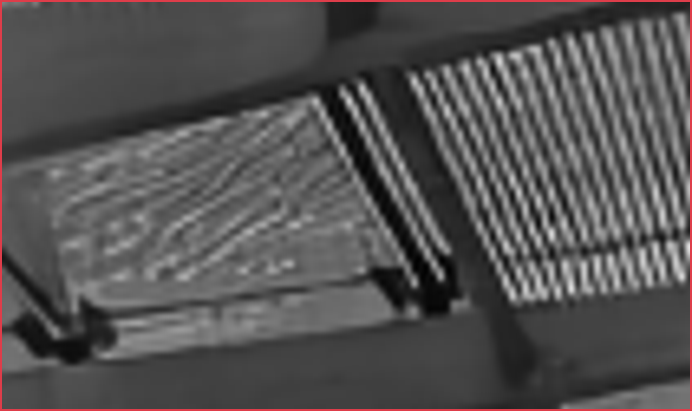}}
        \end{minipage}
        \begin{minipage}[b]{0.175\linewidth}
        	\subfloat[DnCNN]{\includegraphics[width=1\linewidth]{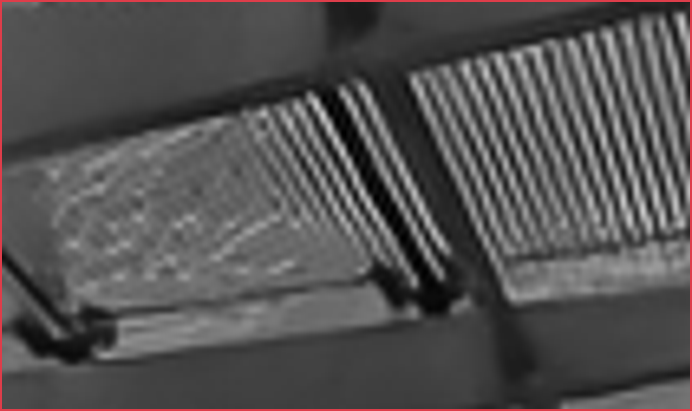}} \vspace{-0.8em}
        	\subfloat[MIMOUNet]{\includegraphics[width=1\linewidth]{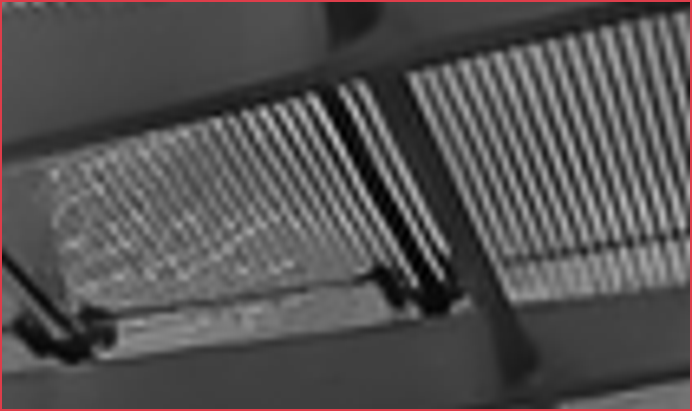}}
        \end{minipage}
        \begin{minipage}[b]{0.175\linewidth}
        	\subfloat[VDSR]{\includegraphics[width=1\linewidth]{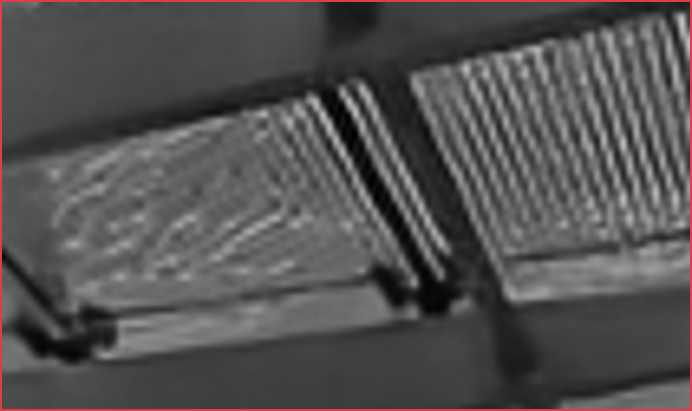}} \vspace{-0.8em}
        	\subfloat[PRL-dt (ours)]{\includegraphics[width=1\linewidth]{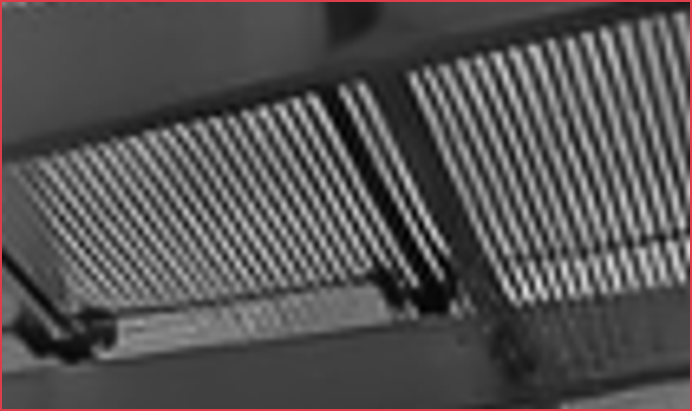}}
        \end{minipage}
        \begin{minipage}[b]{0.175\linewidth}
        	\subfloat[EDSR]{\includegraphics[width=1\linewidth]{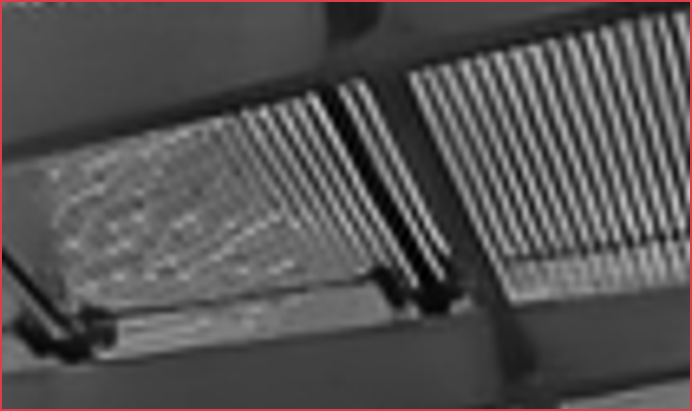}} \vspace{-0.8em}
        	\subfloat[MSPRL (ours)]{\includegraphics[width=1\linewidth]{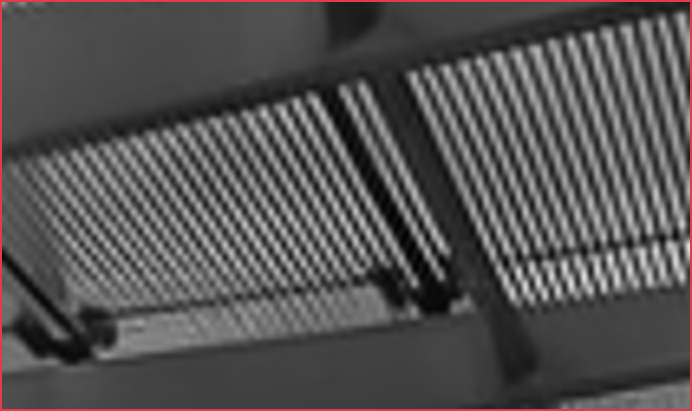}}
        \end{minipage}
        \\ \vspace{0.5em}
        % img_078
        \begin{minipage}[b]{0.255\linewidth}
        	\subfloat[Urban100: img\_078]{\includegraphics[width=1\linewidth]{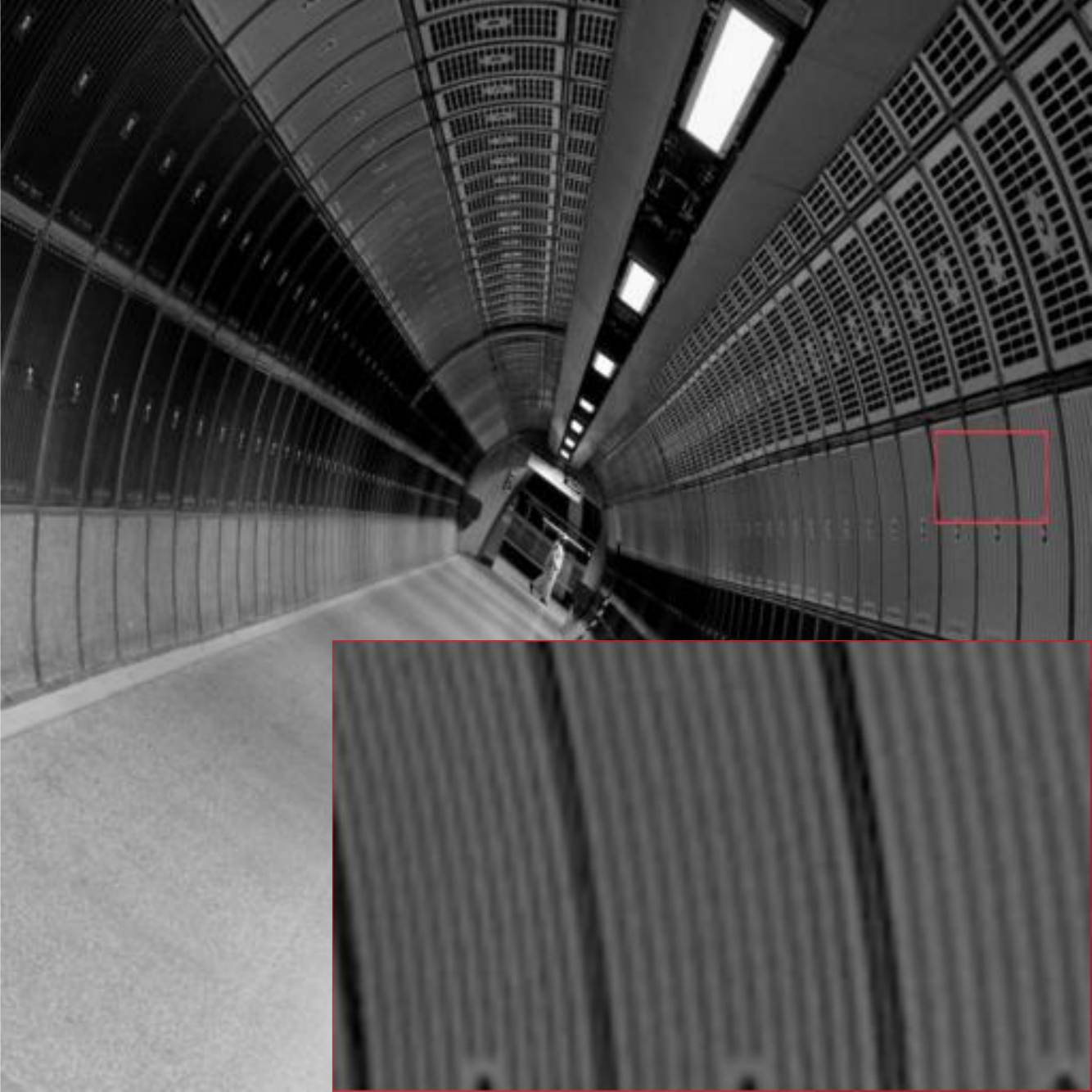}}
        \end{minipage}
        \begin{minipage}[b]{0.175\linewidth}
        	\subfloat[Halftone]{\includegraphics[width=1\linewidth]{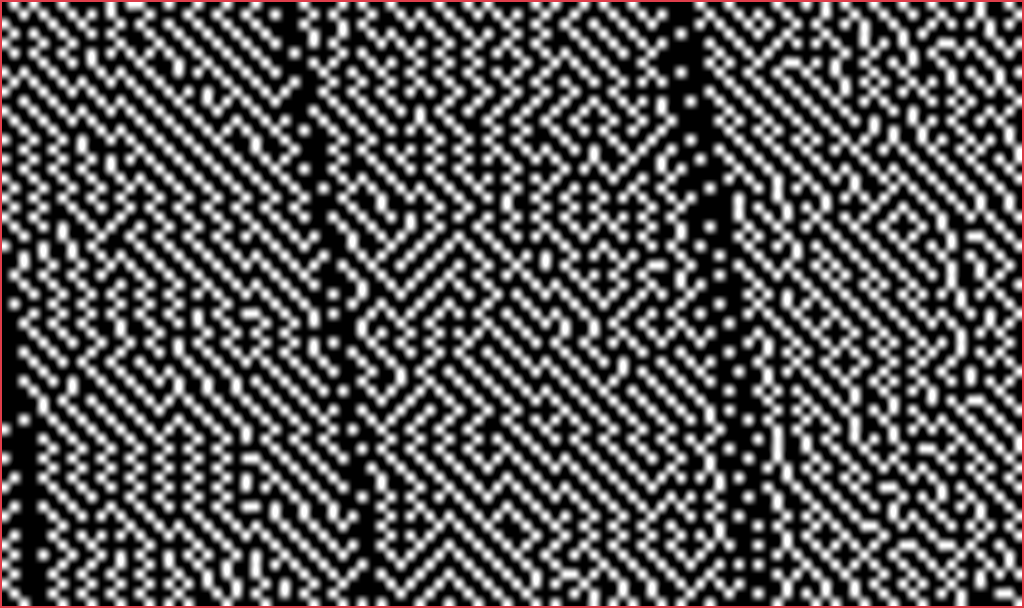}} \vspace{-0.8em}
        	\subfloat[PRL]{\includegraphics[width=1\linewidth]{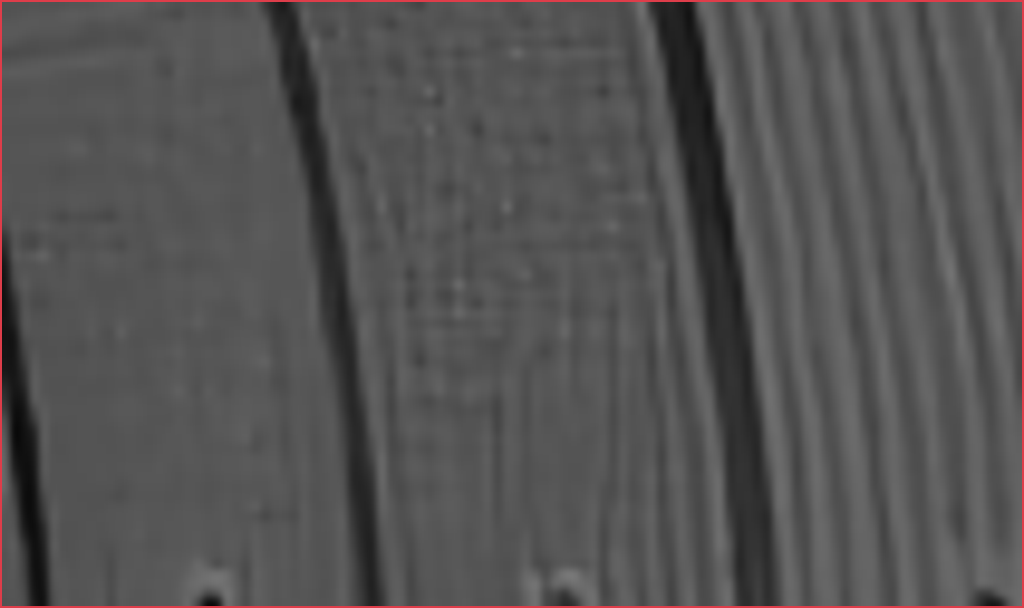}}
        \end{minipage}
        \begin{minipage}[b]{0.175\linewidth}
        	\subfloat[DnCNN]{\includegraphics[width=1\linewidth]{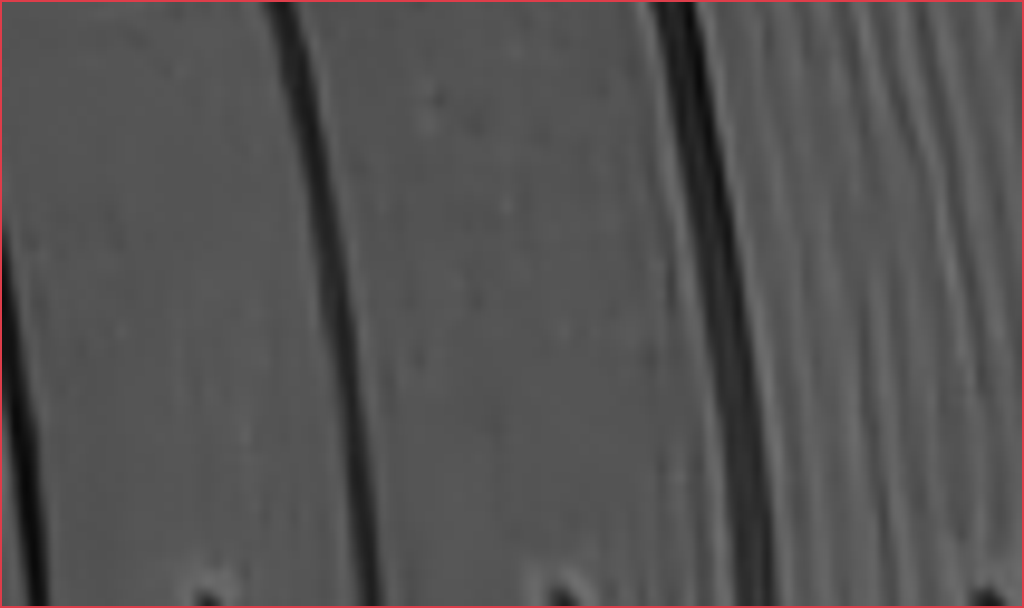}} \vspace{-0.8em}
        	\subfloat[MIMOUNet]{\includegraphics[width=1\linewidth]{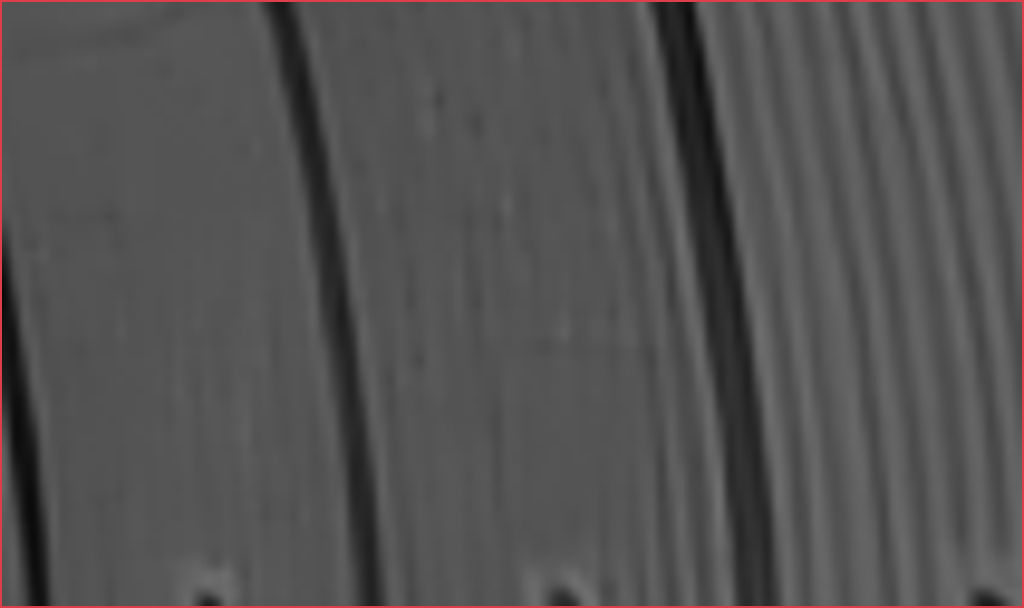}}
        \end{minipage}
        \begin{minipage}[b]{0.175\linewidth}
        	\subfloat[VDSR]{\includegraphics[width=1\linewidth]{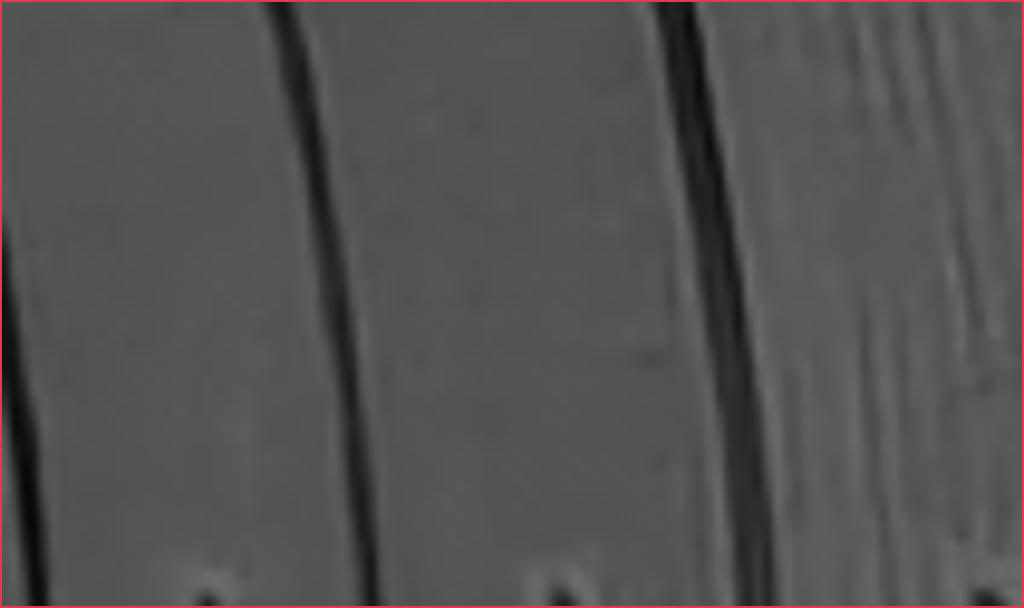}} \vspace{-0.8em}
        	\subfloat[PRL-dt (ours)]{\includegraphics[width=1\linewidth]{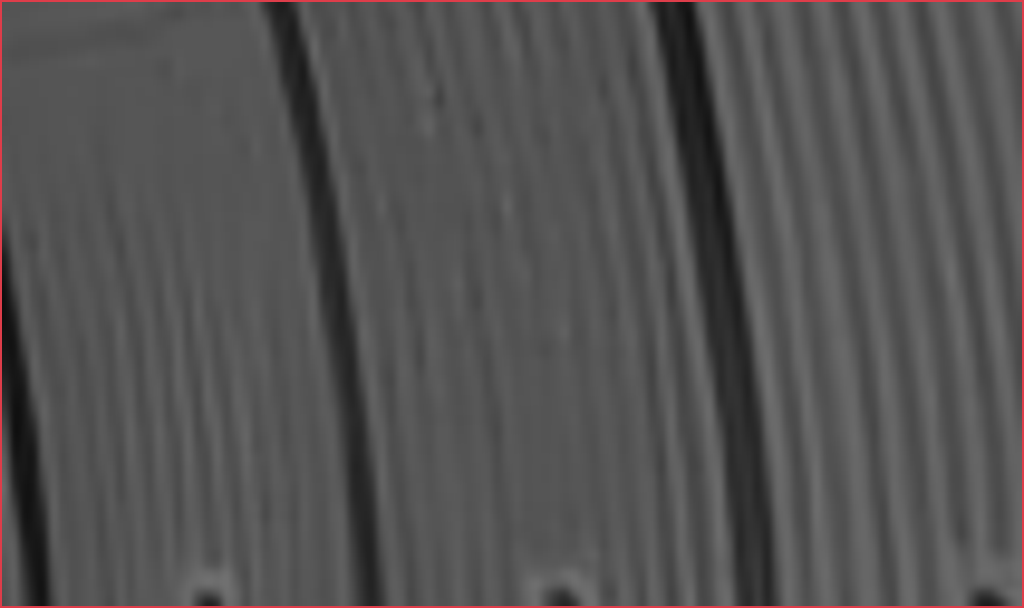}}
        \end{minipage}
        \begin{minipage}[b]{0.175\linewidth}
        	\subfloat[EDSR]{\includegraphics[width=1\linewidth]{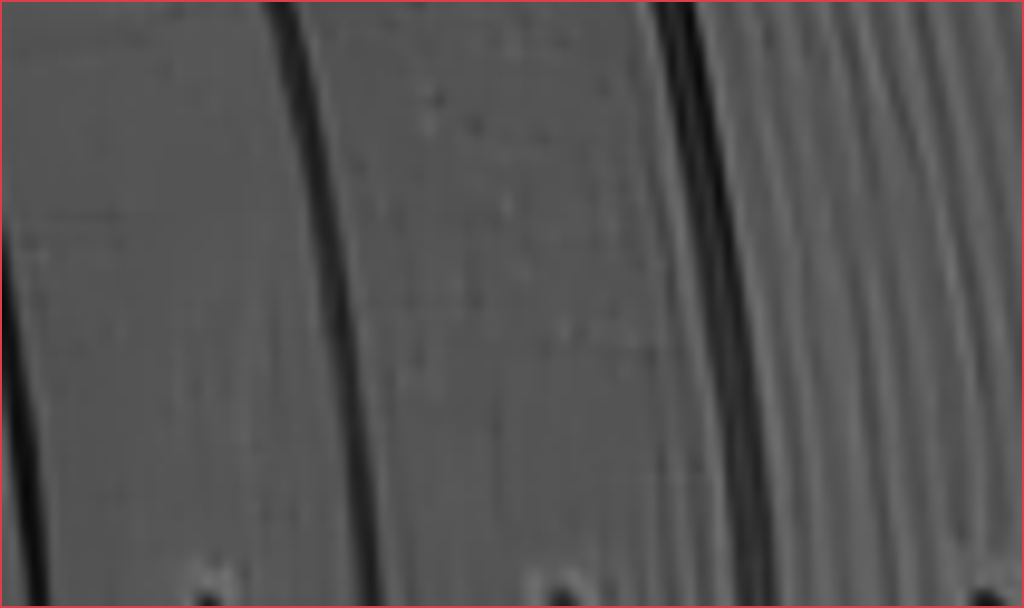}} \vspace{-0.8em}
        	\subfloat[MSPRL (ours)]{\includegraphics[width=1\linewidth]{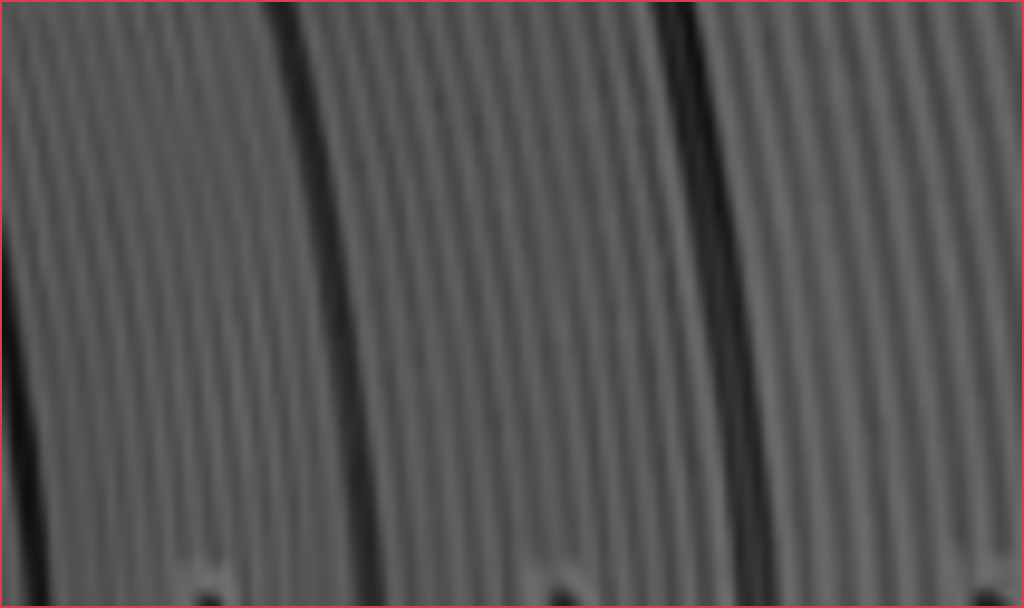}}
        \end{minipage}
        \\ \vspace{0.5em}
        % img_092
        \begin{minipage}[b]{0.255\linewidth}
        	\subfloat[Urban100: img\_092]{\includegraphics[width=1\linewidth]{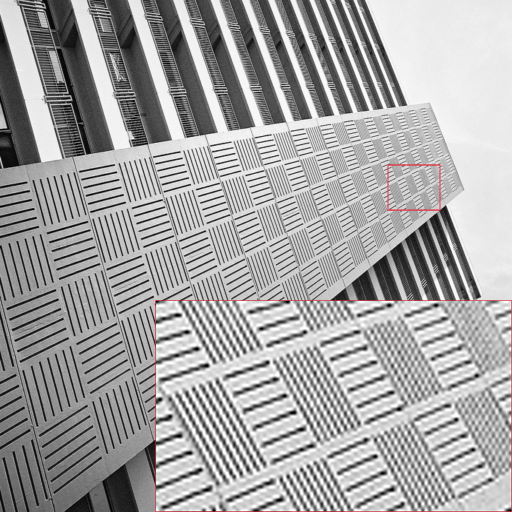}}
        \end{minipage}
        \begin{minipage}[b]{0.175\linewidth}
        	\subfloat[Halftone]{\includegraphics[width=1\linewidth]{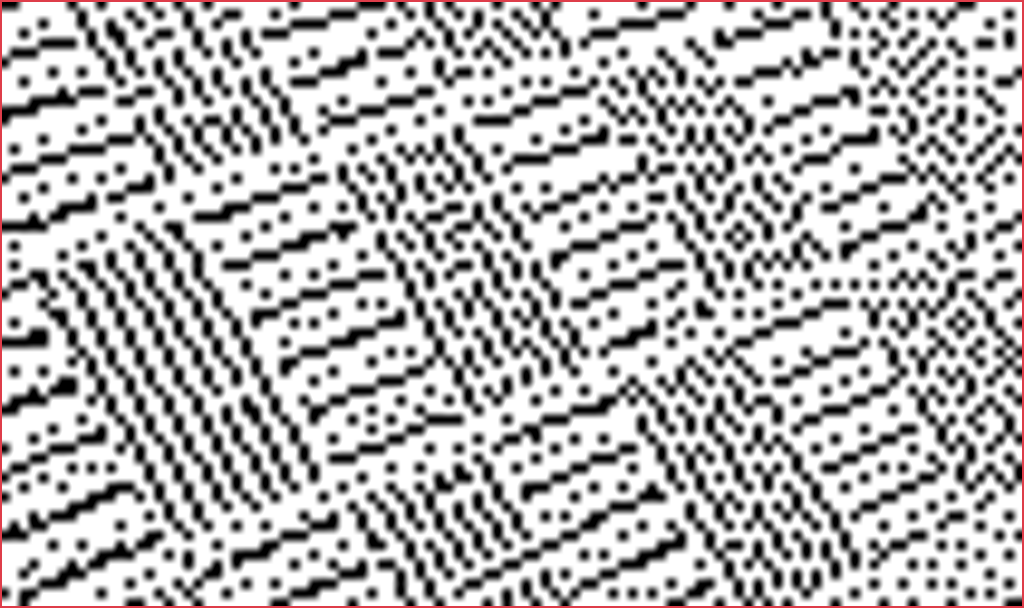}} \vspace{-0.8em}
        	\subfloat[PRL]{\includegraphics[width=1\linewidth]{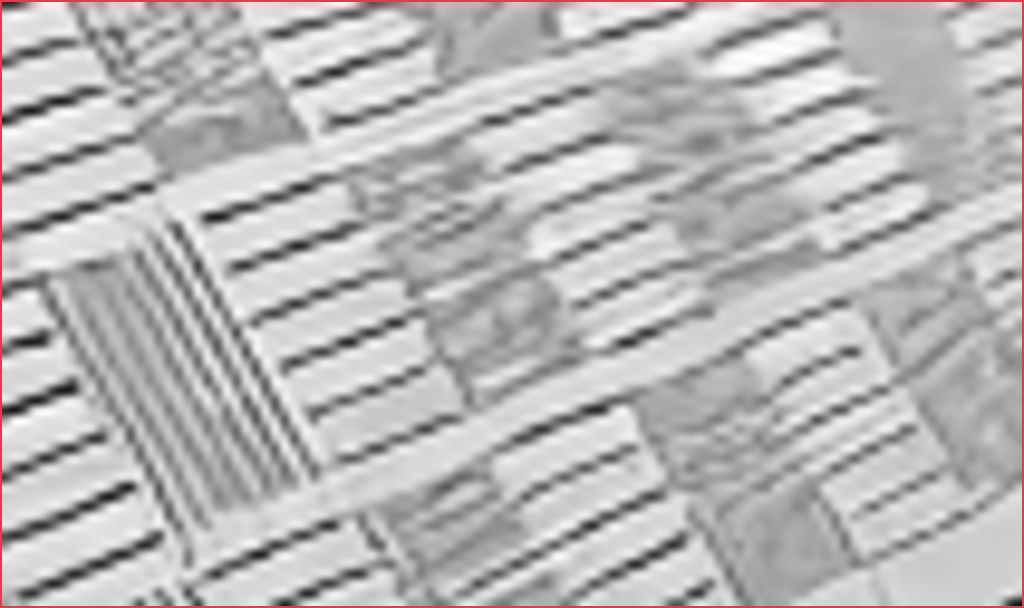}}
        \end{minipage}
        \begin{minipage}[b]{0.175\linewidth}
        	\subfloat[DnCNN]{\includegraphics[width=1\linewidth]{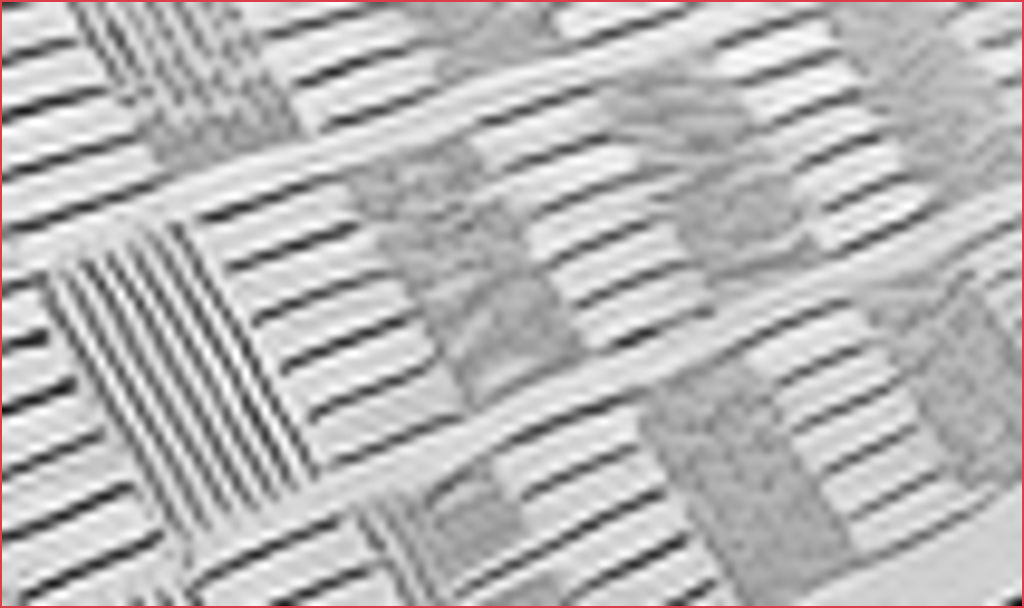}} \vspace{-0.8em}
        	\subfloat[MIMOUNet]{\includegraphics[width=1\linewidth]{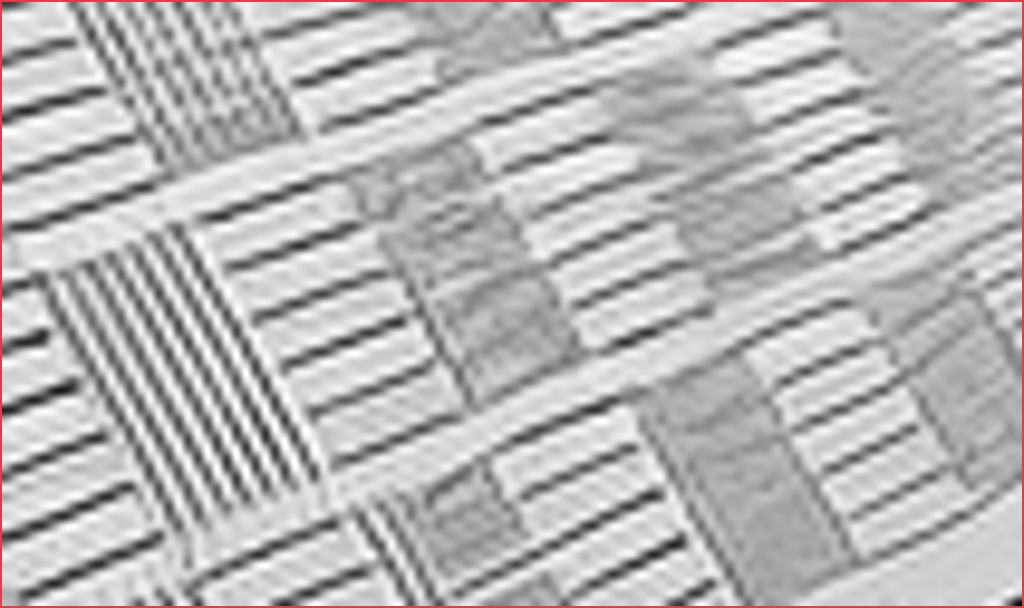}}
        \end{minipage}
        \begin{minipage}[b]{0.175\linewidth}
        	\subfloat[VDSR]{\includegraphics[width=1\linewidth]{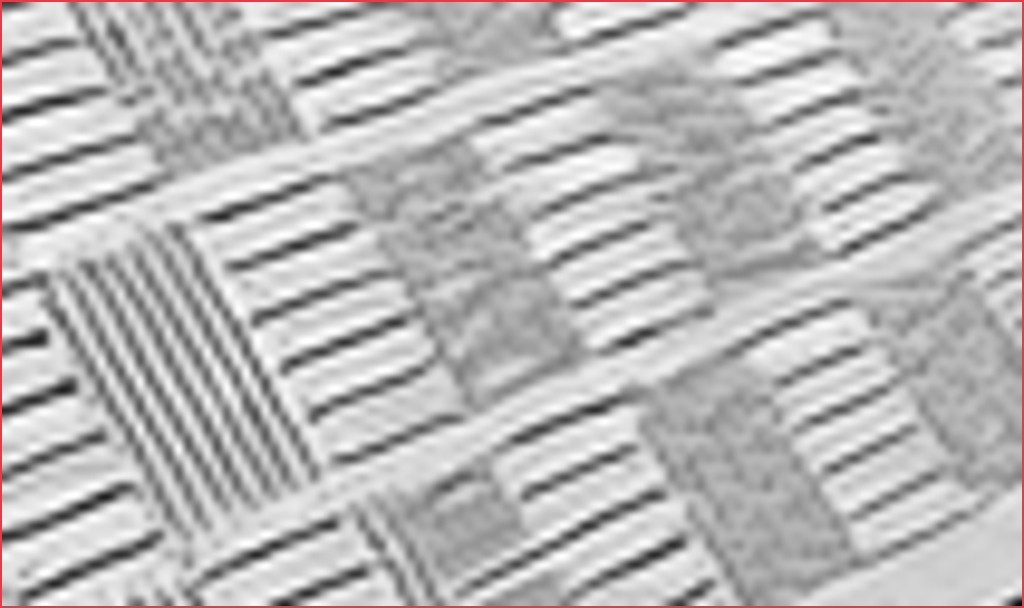}} \vspace{-0.8em}
        	\subfloat[PRL-dt (ours)]{\includegraphics[width=1\linewidth]{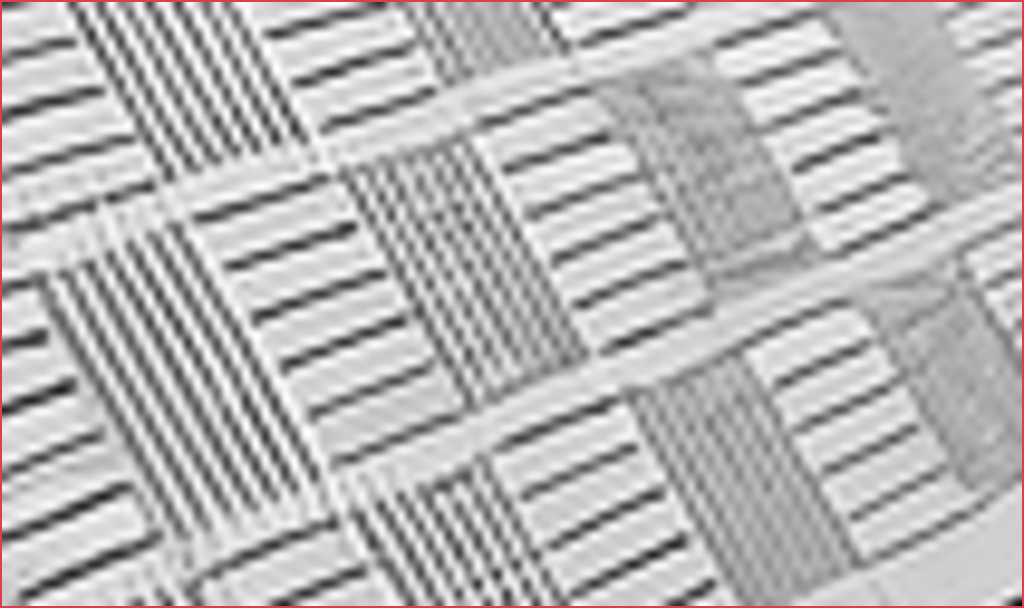}}
        \end{minipage}
        \begin{minipage}[b]{0.175\linewidth}
        	\subfloat[EDSR]{\includegraphics[width=1\linewidth]{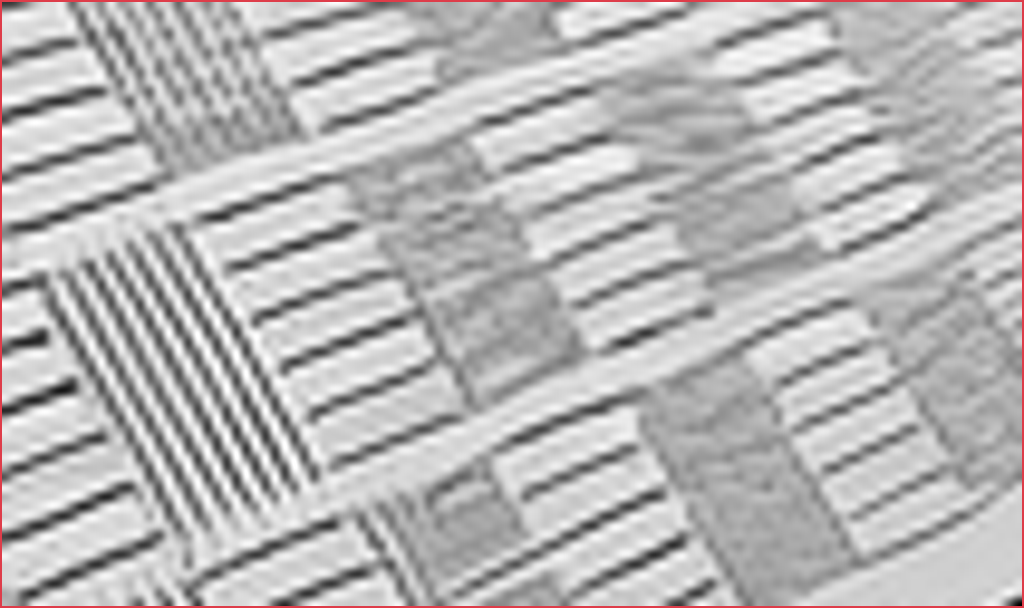}} \vspace{-0.8em}
        	\subfloat[MSPRL (ours)]{\includegraphics[width=1\linewidth]{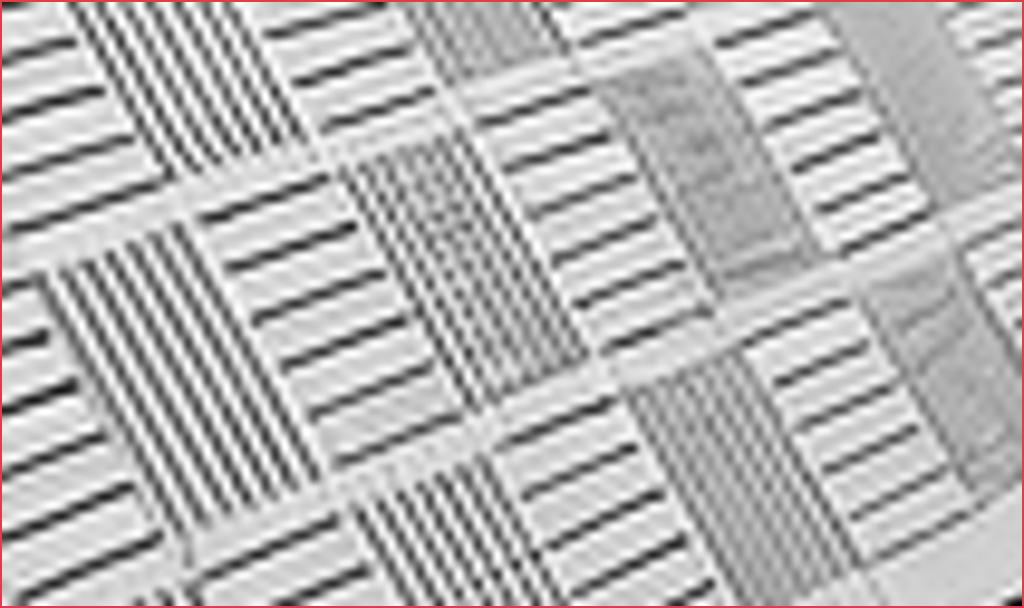}}
        \end{minipage}
        \end{center}
        \vspace{0.5em}
        \caption{Compared with the other approaches, our MSPRL more effectively restores the image details.}
    \label{fig:c4}
\end{figure*}

\begin{figure*}[ht]
\begin{center}
    \captionsetup[subfloat]{labelsep=none,format=plain,labelformat=empty,font={scriptsize}}
        % kodim14
        \begin{minipage}[b]{0.255\linewidth}
        	\subfloat[kodim14]{\includegraphics[width=1\linewidth]{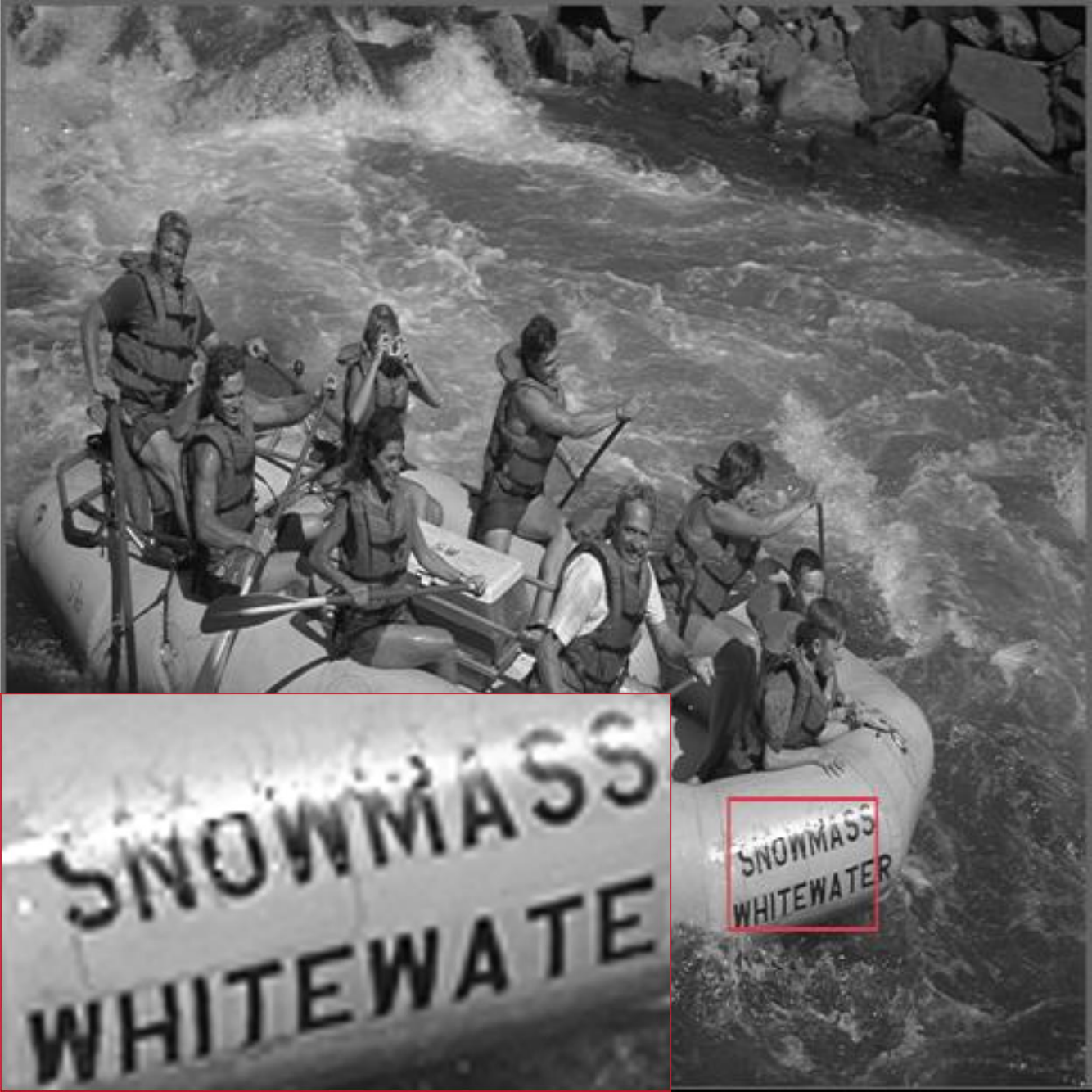}}
        \end{minipage}
        \begin{minipage}[b]{0.175\linewidth}
        	\subfloat[DnCNN]{\includegraphics[width=1\linewidth]{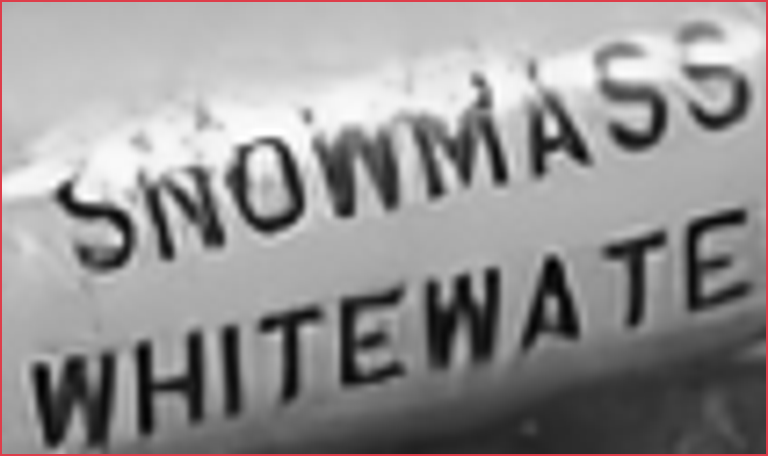}} \vspace{-0.8em}
        	\subfloat[GGRL]{\includegraphics[width=1\linewidth]{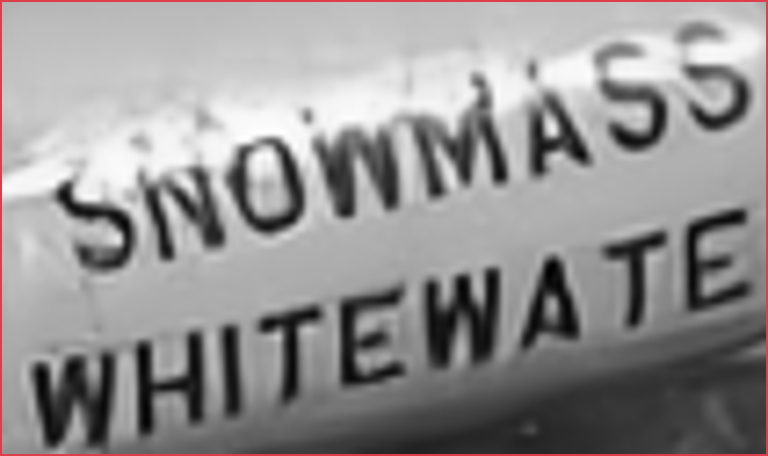}}
        \end{minipage}
        \begin{minipage}[b]{0.175\linewidth}
        	\subfloat[VDSR]{\includegraphics[width=1\linewidth]{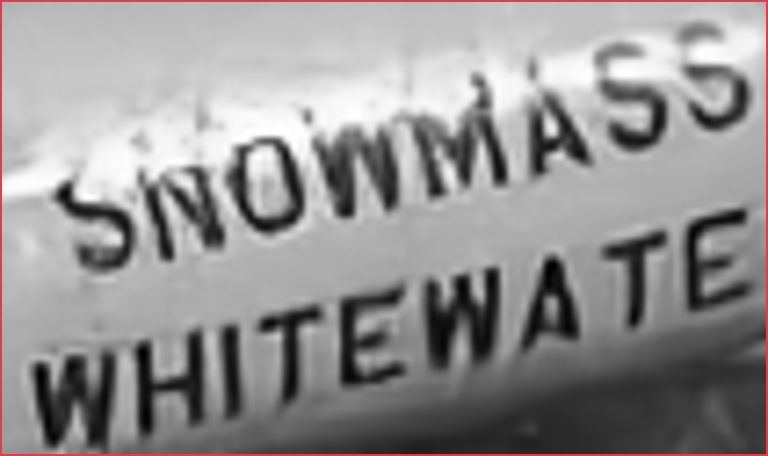}} \vspace{-0.8em}
        	\subfloat[MIMOUNet]{\includegraphics[width=1\linewidth]{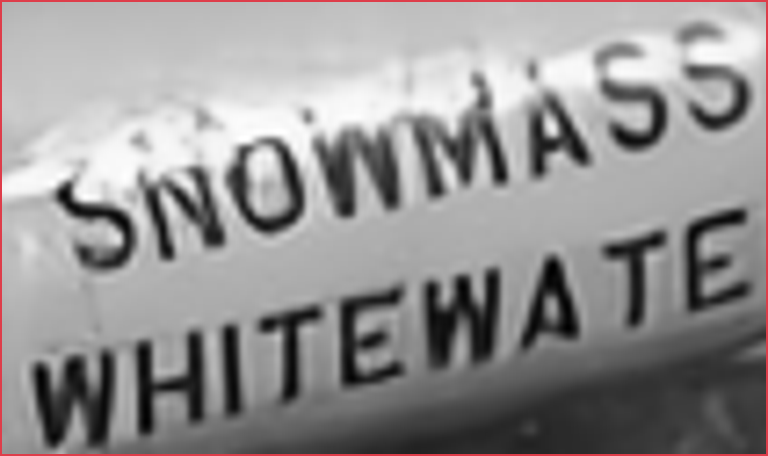}}
        \end{minipage}
        \begin{minipage}[b]{0.175\linewidth}
        	\subfloat[EDSR]{\includegraphics[width=1\linewidth]{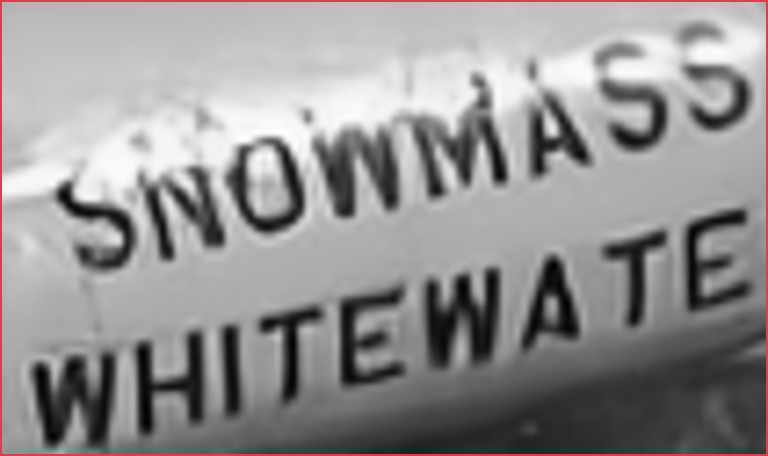}} \vspace{-0.8em}
        	\subfloat[PRL-dt (ours)]{\includegraphics[width=1\linewidth]{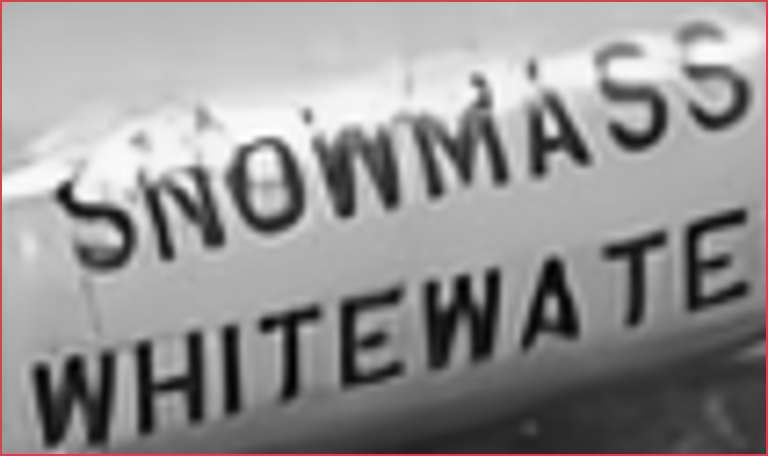}}
        \end{minipage}
        \begin{minipage}[b]{0.175\linewidth}
        	\subfloat[PRL]{\includegraphics[width=1\linewidth]{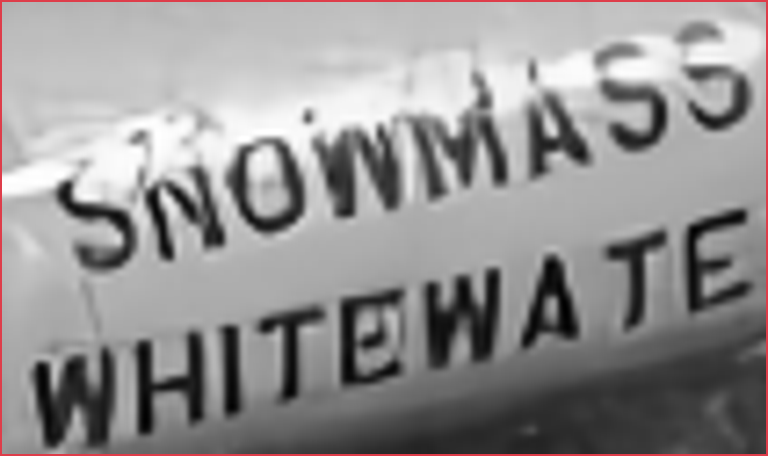}} \vspace{-0.8em}
        	\subfloat[MSPRL (ours)]{\includegraphics[width=1\linewidth]{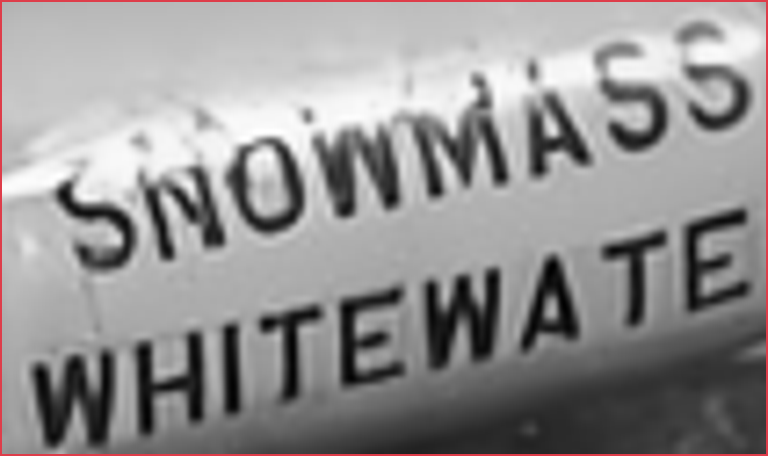}}
        \end{minipage}
        \\ \vspace{0.5em}
        % img_060
        \begin{minipage}[b]{0.255\linewidth}
        	\subfloat[Urban100: img\_060]{\includegraphics[width=1\linewidth]{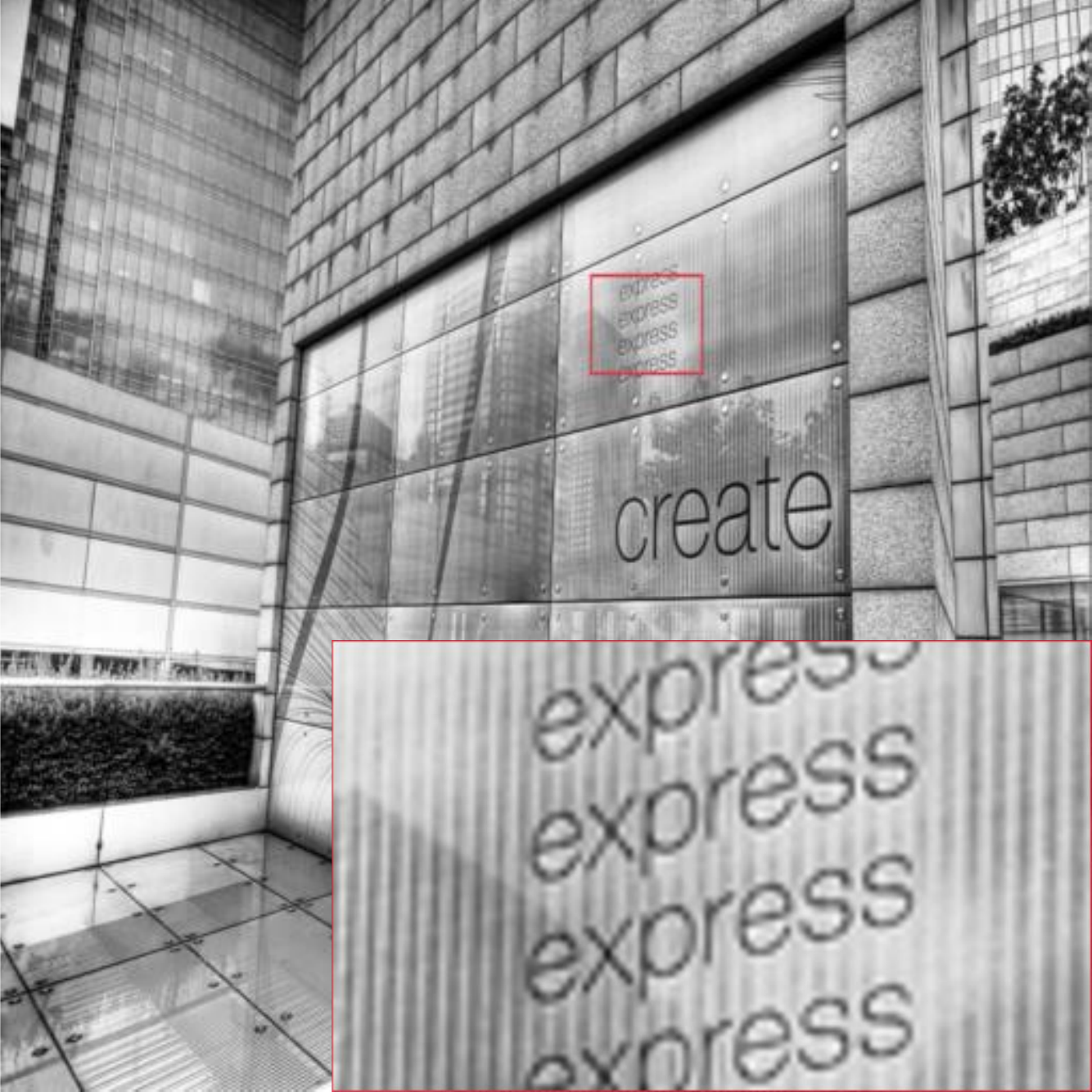}}
        \end{minipage}
        \begin{minipage}[b]{0.175\linewidth}
        	\subfloat[Halftone]{\includegraphics[width=1\linewidth]{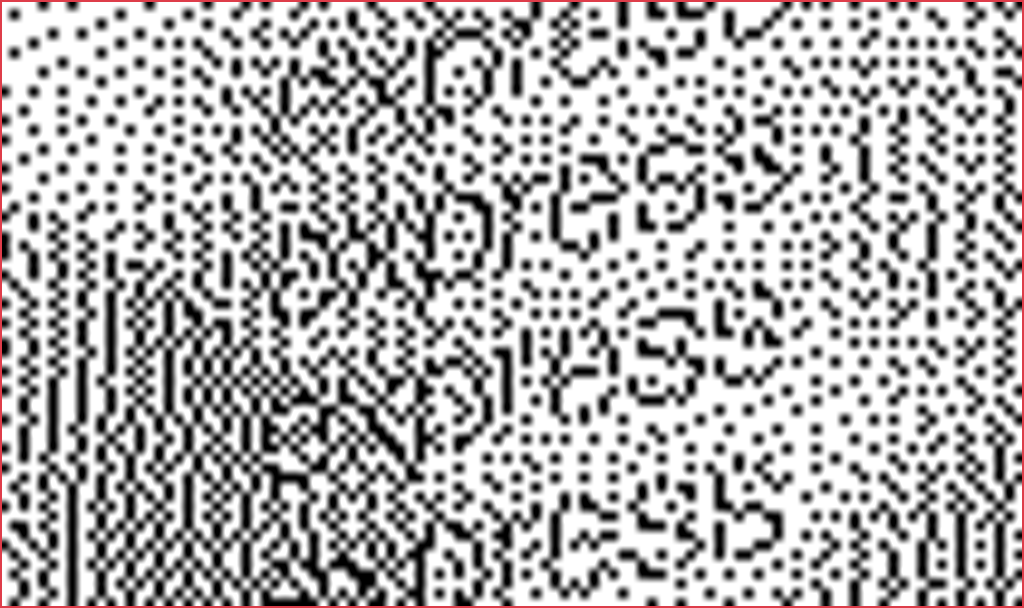}} \vspace{-0.8em}
        	\subfloat[PRL]{\includegraphics[width=1\linewidth]{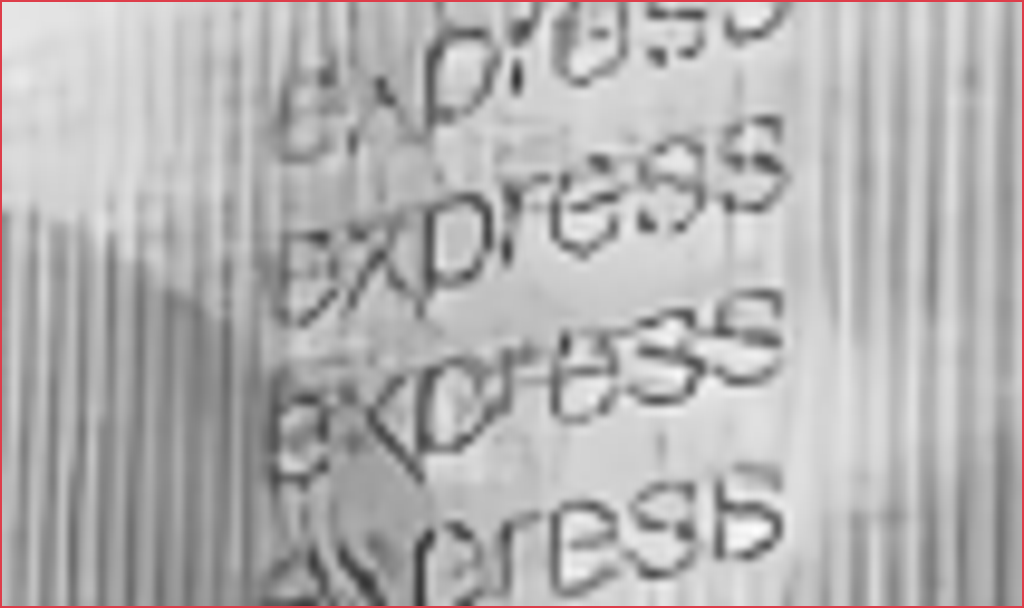}}
        \end{minipage}
        \begin{minipage}[b]{0.175\linewidth}
        	\subfloat[DnCNN]{\includegraphics[width=1\linewidth]{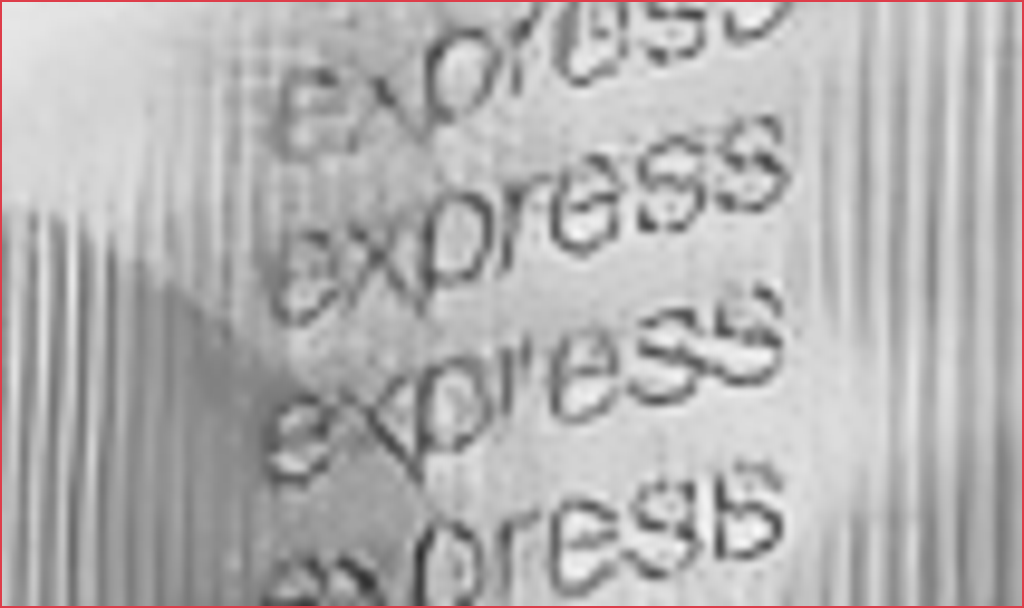}} \vspace{-0.8em}
        	\subfloat[MIMOUNet]{\includegraphics[width=1\linewidth]{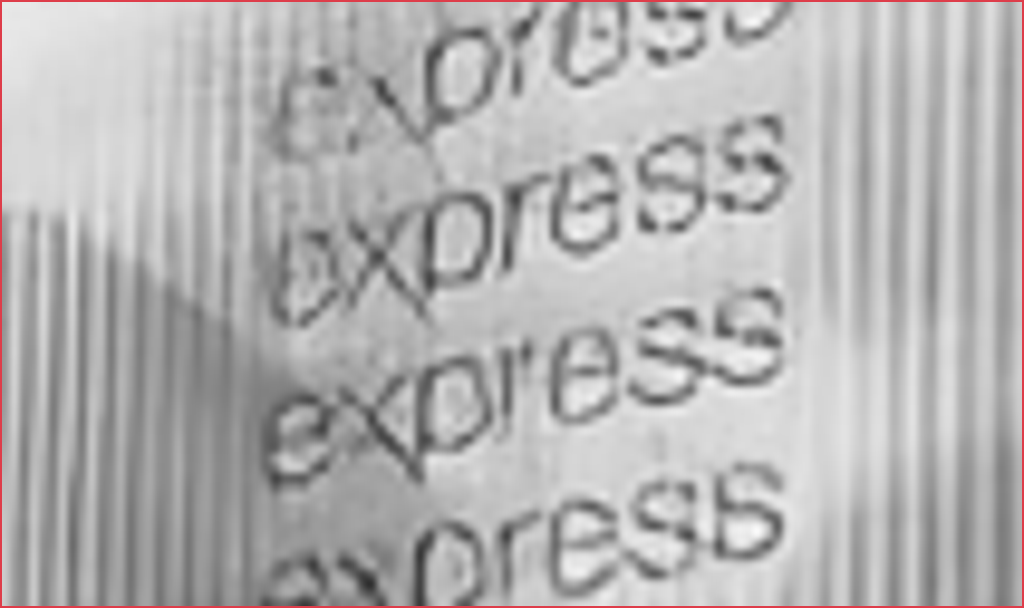}}
        \end{minipage}
        \begin{minipage}[b]{0.175\linewidth}
        	\subfloat[VDSR]{\includegraphics[width=1\linewidth]{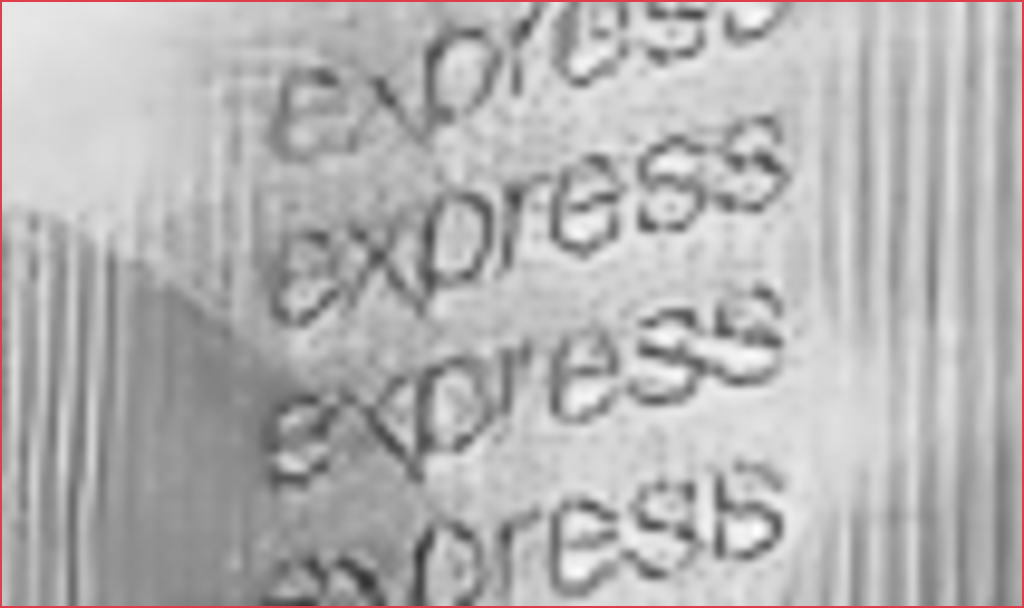}} \vspace{-0.8em}
        	\subfloat[PRL-dt (ours)]{\includegraphics[width=1\linewidth]{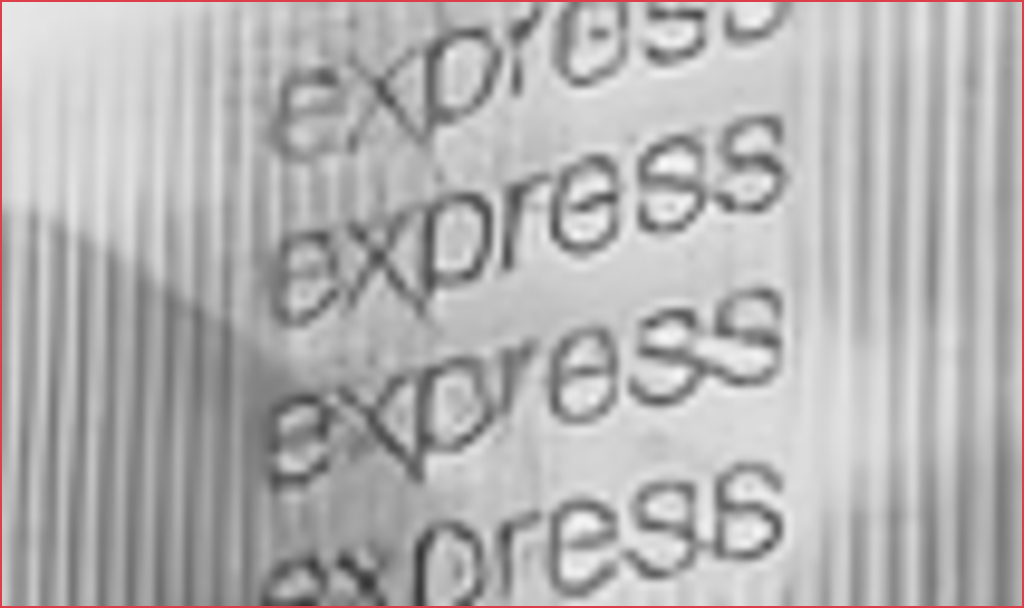}}
        \end{minipage}
        \begin{minipage}[b]{0.175\linewidth}
        	\subfloat[EDSR]{\includegraphics[width=1\linewidth]{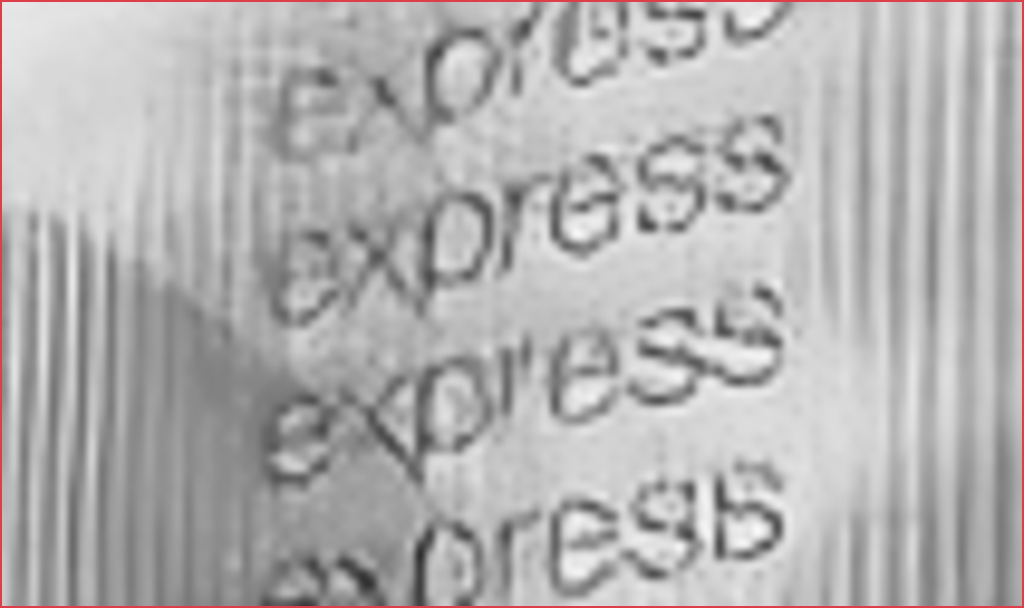}} \vspace{-0.8em}
        	\subfloat[MSPRL (ours)]{\includegraphics[width=1\linewidth]{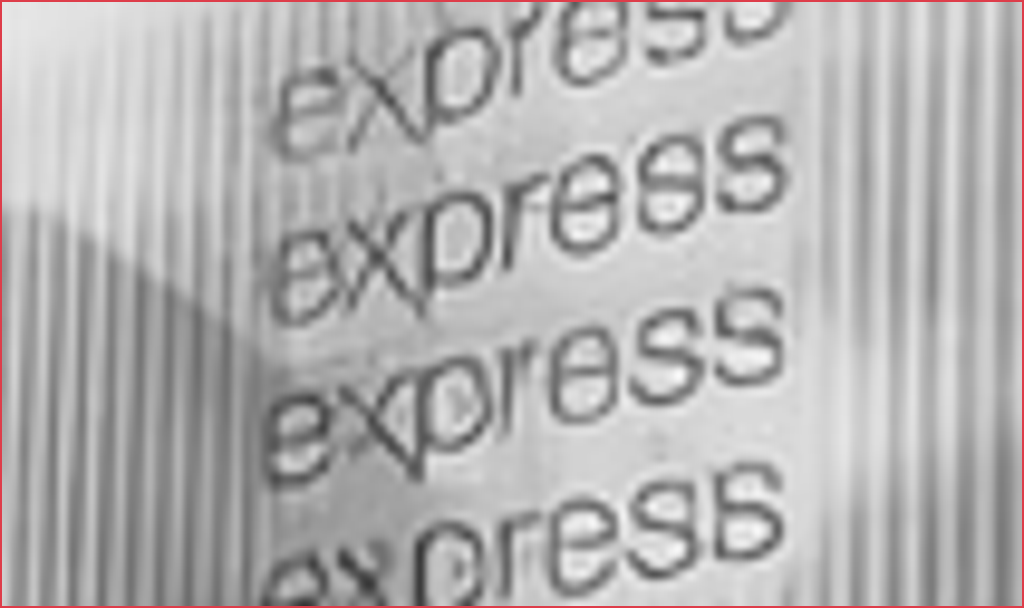}}
        \end{minipage}
        \\ \vspace{0.5em}
        % ARMS
        \begin{minipage}[b]{0.255\linewidth}
        	\subfloat[Manga109: ARMS]{\includegraphics[width=1\linewidth]{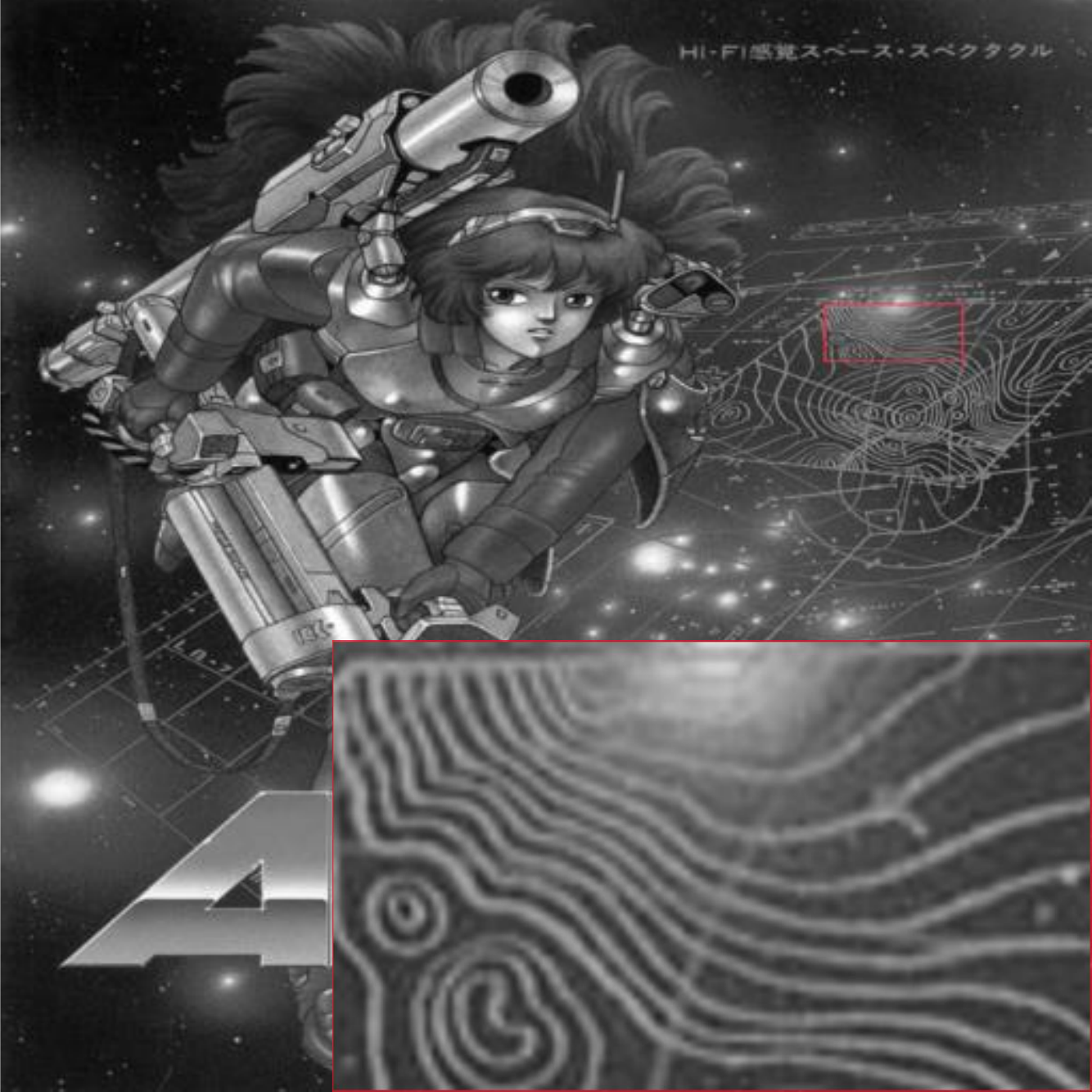}}
        \end{minipage}
        \begin{minipage}[b]{0.175\linewidth}
        	\subfloat[Halftone]{\includegraphics[width=1\linewidth]{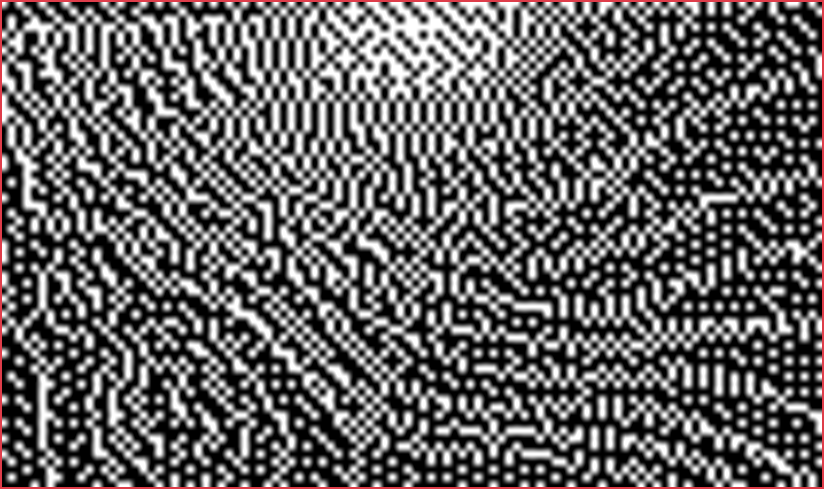}} \vspace{-0.8em}
        	\subfloat[PRL]{\includegraphics[width=1\linewidth]{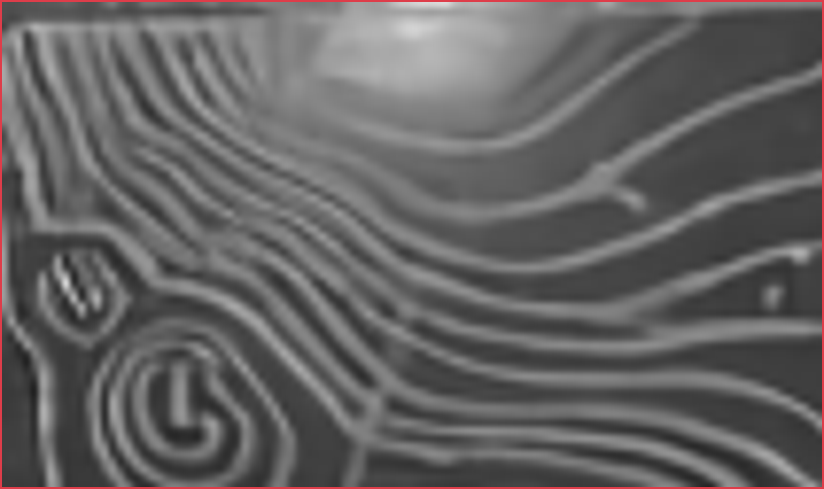}}
        \end{minipage}
        \begin{minipage}[b]{0.175\linewidth}
        	\subfloat[DnCNN]{\includegraphics[width=1\linewidth]{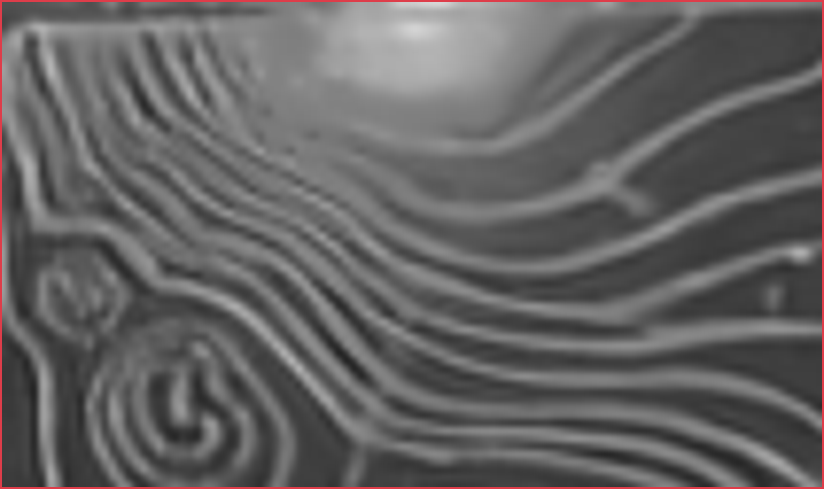}} \vspace{-0.8em}
        	\subfloat[MIMOUNet]{\includegraphics[width=1\linewidth]{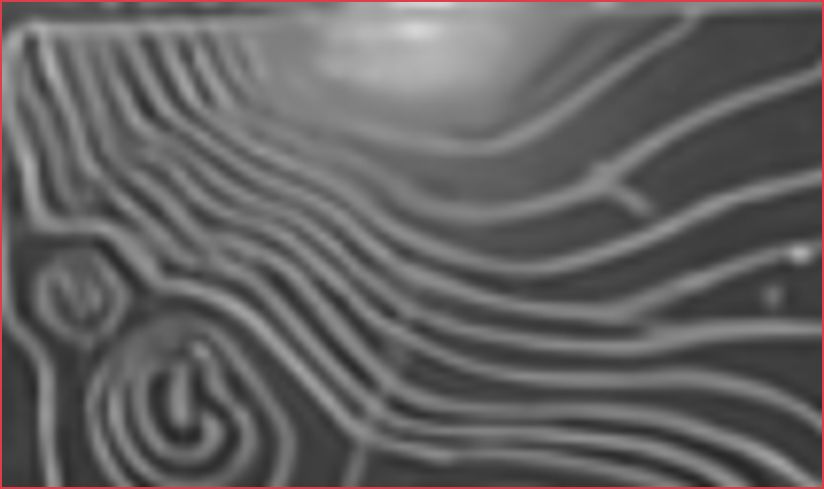}}
        \end{minipage}
        \begin{minipage}[b]{0.175\linewidth}
        	\subfloat[VDSR]{\includegraphics[width=1\linewidth]{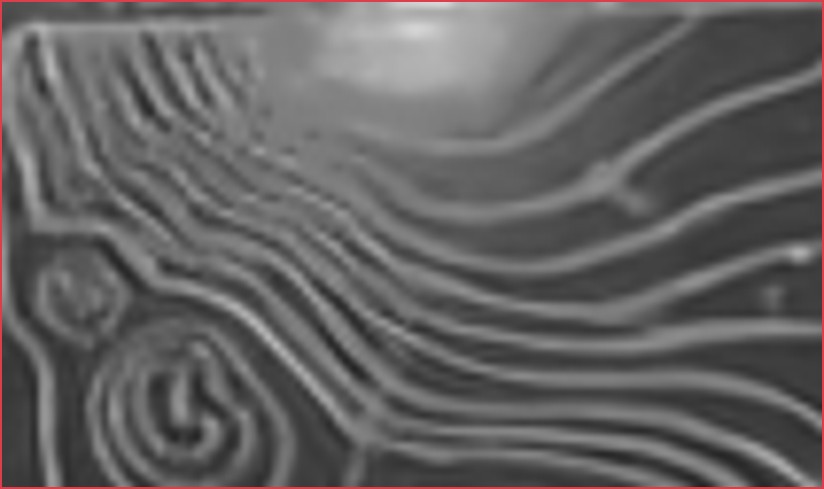}} \vspace{-0.8em}
        	\subfloat[PRL-dt (ours)]{\includegraphics[width=1\linewidth]{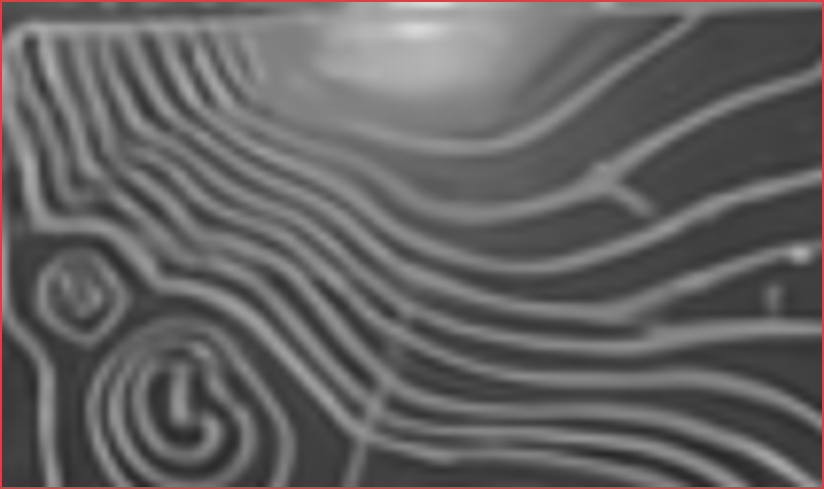}}
        \end{minipage}
        \begin{minipage}[b]{0.175\linewidth}
        	\subfloat[EDSR]{\includegraphics[width=1\linewidth]{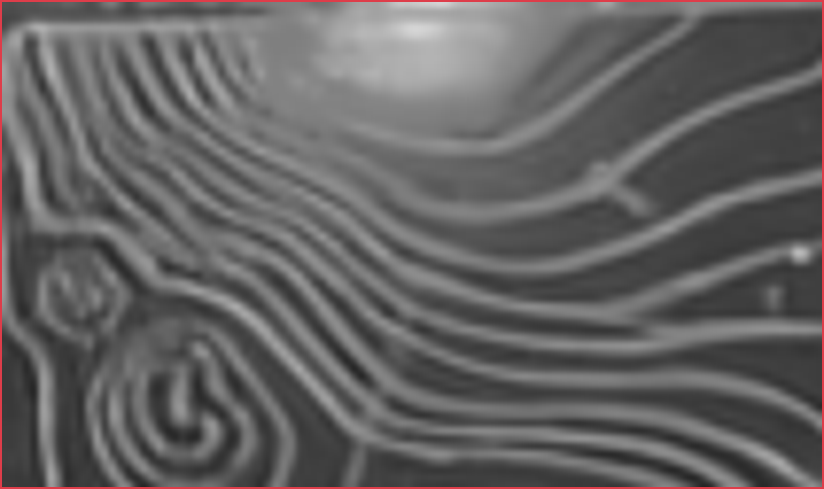}} \vspace{-0.8em}
        	\subfloat[MSPRL (ours)]{\includegraphics[width=1\linewidth]{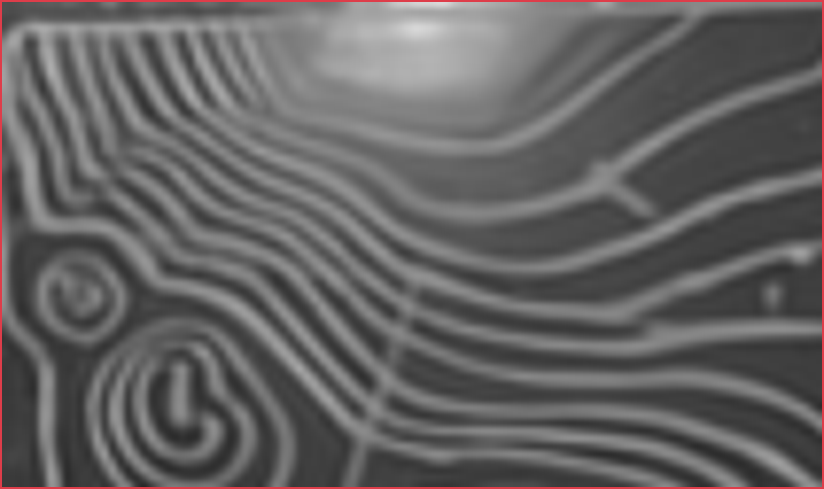}}
        \end{minipage}
        \\ \vspace{0.5em}
        % MoeruOnisan_vol19
        \begin{minipage}[b]{0.255\linewidth}
        	\subfloat[Manga109: MoeruOnisan\_vol19]{\includegraphics[width=1\linewidth]{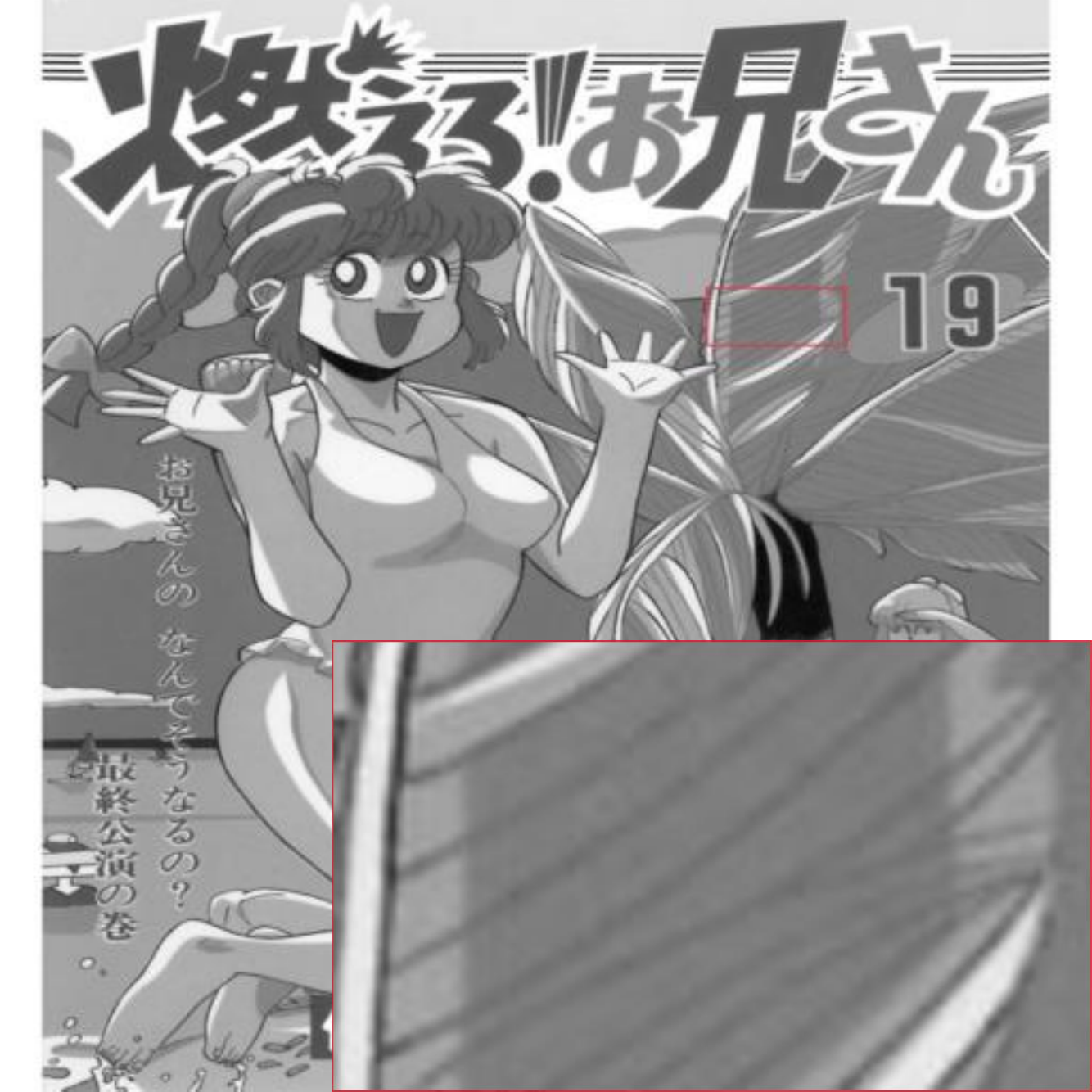}}
        \end{minipage}
        \begin{minipage}[b]{0.175\linewidth}
        	\subfloat[Halftone]{\includegraphics[width=1\linewidth]{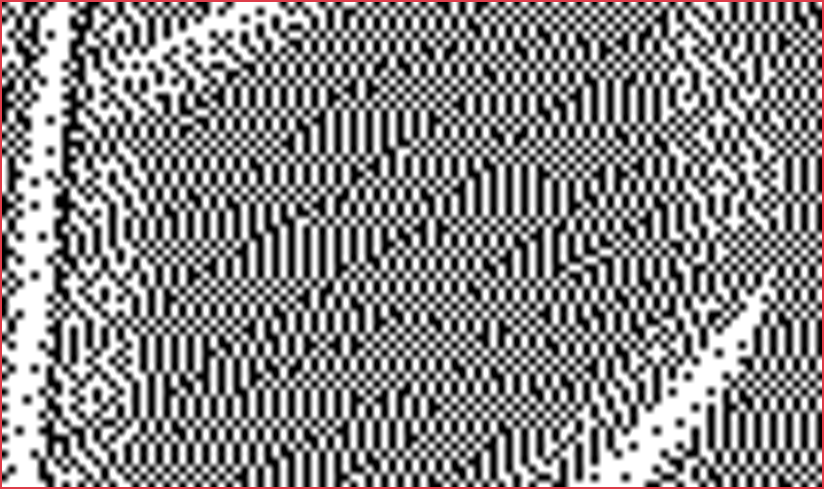}} \vspace{-0.8em}
        	\subfloat[PRL]{\includegraphics[width=1\linewidth]{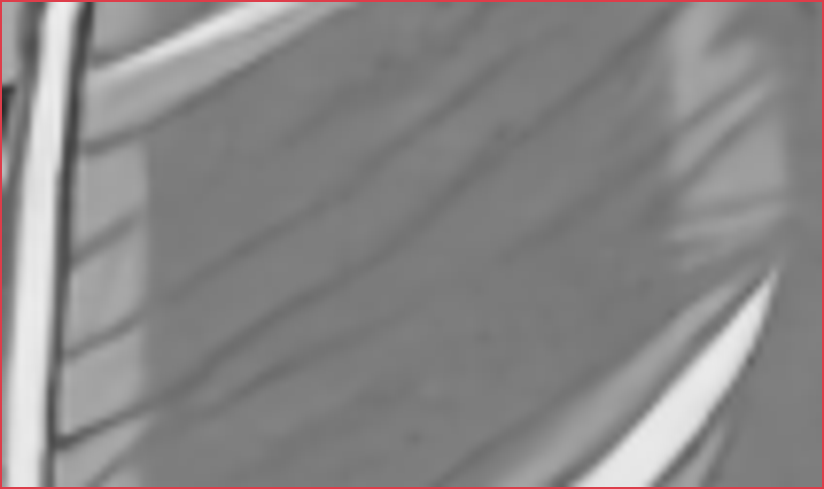}}
        \end{minipage}
        \begin{minipage}[b]{0.175\linewidth}
        	\subfloat[DnCNN]{\includegraphics[width=1\linewidth]{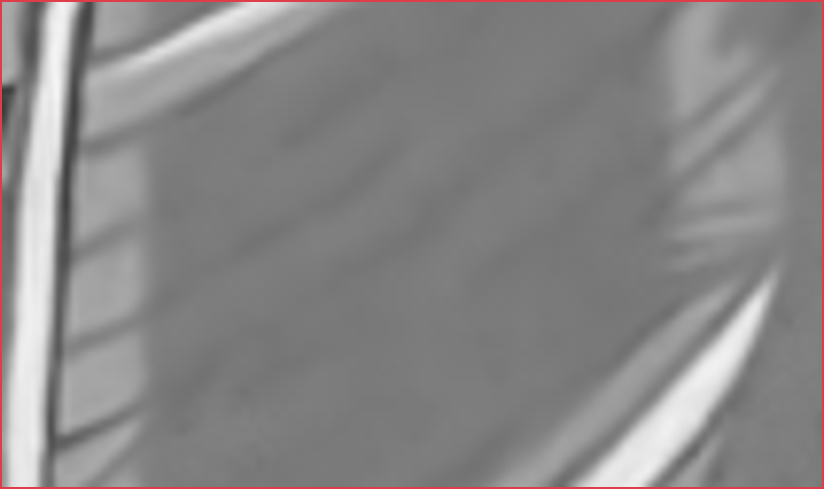}} \vspace{-0.8em}
        	\subfloat[MIMOUNet]{\includegraphics[width=1\linewidth]{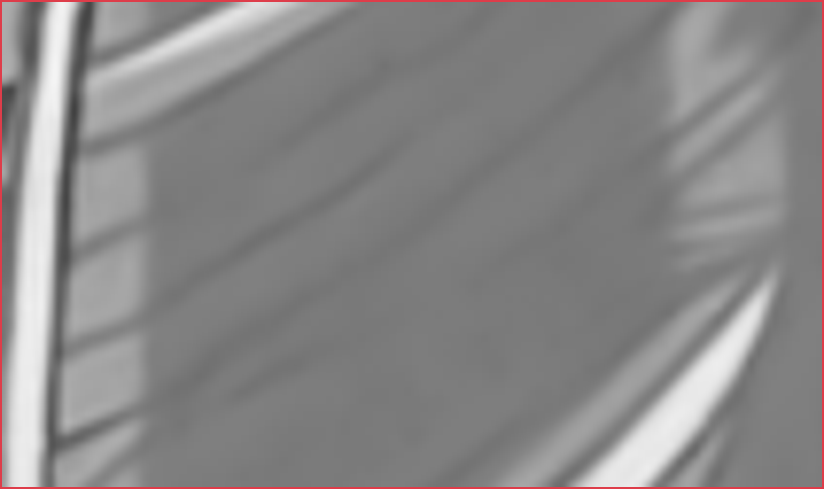}}
        \end{minipage}
        \begin{minipage}[b]{0.175\linewidth}
        	\subfloat[VDSR]{\includegraphics[width=1\linewidth]{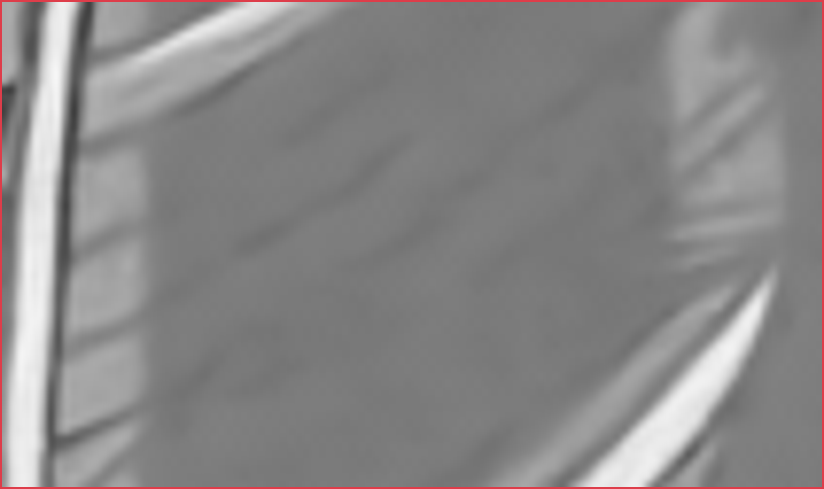}} \vspace{-0.8em}
        	\subfloat[PRL-dt (ours)]{\includegraphics[width=1\linewidth]{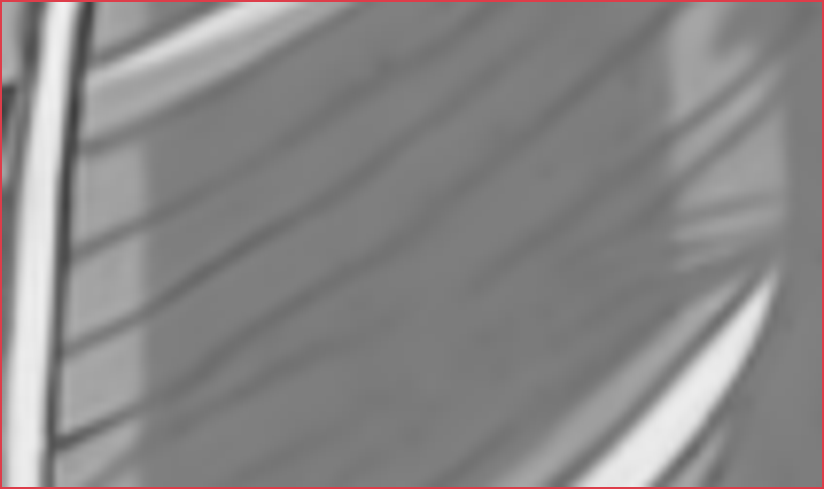}}
        \end{minipage}
        \begin{minipage}[b]{0.175\linewidth}
        	\subfloat[EDSR]{\includegraphics[width=1\linewidth]{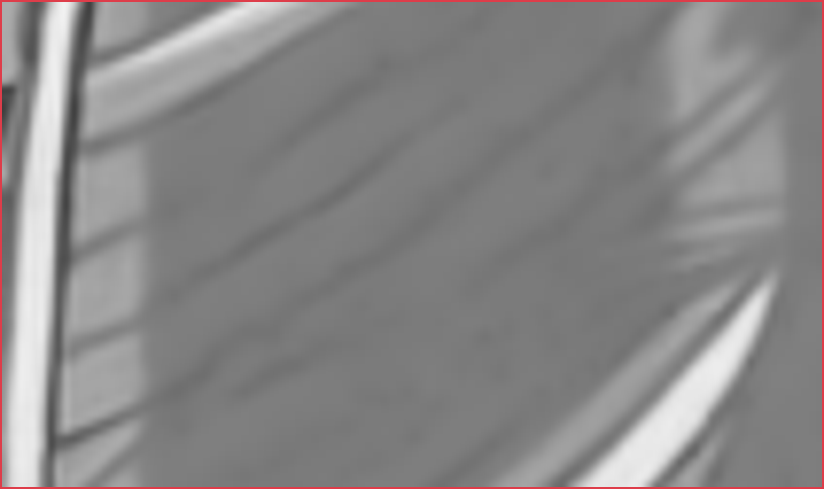}} \vspace{-0.8em}
        	\subfloat[MSPRL (ours)]{\includegraphics[width=1\linewidth]{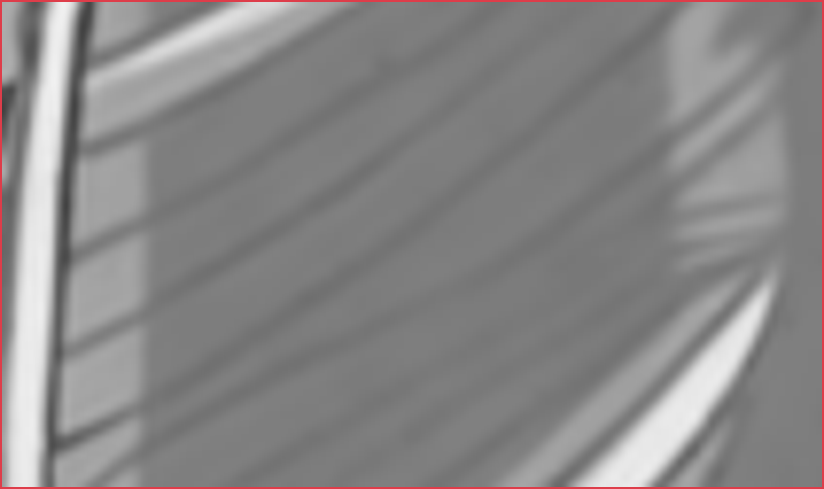}}
        \end{minipage}
        \end{center}
        \vspace{0.5em}
        \caption{Compared with the other approaches, our MSPRL more effectively restores the image details.}
    \label{fig:c5}
\end{figure*}

\begin{figure*}[ht]
\begin{center}
    \captionsetup[subfloat]{labelsep=none,format=plain,labelformat=empty,font={scriptsize}}
        % 78004
        \begin{minipage}[b]{0.18\linewidth}
            \subfloat[Ground Truth]{\includegraphics[width=1\linewidth]{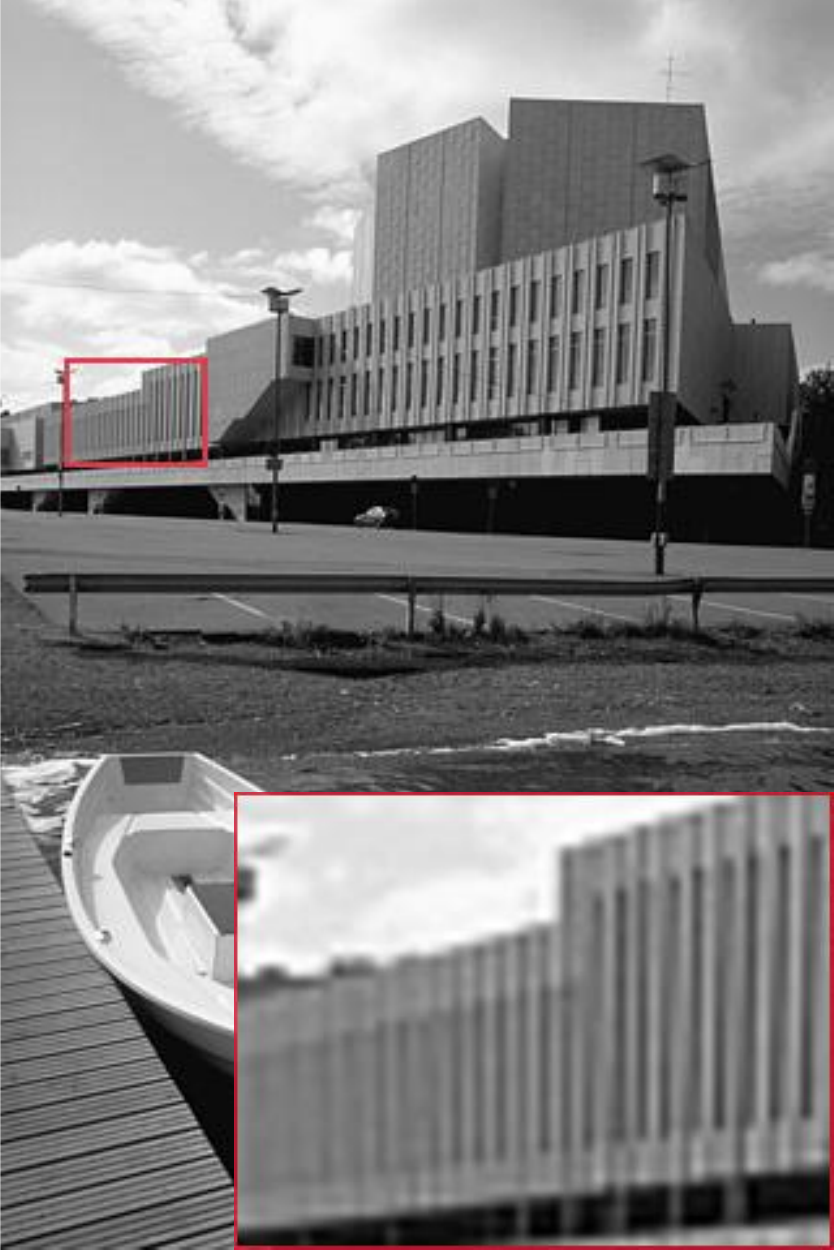}} \vspace{-0.9em}
            \subfloat[PRL]{\includegraphics[width=1\linewidth]{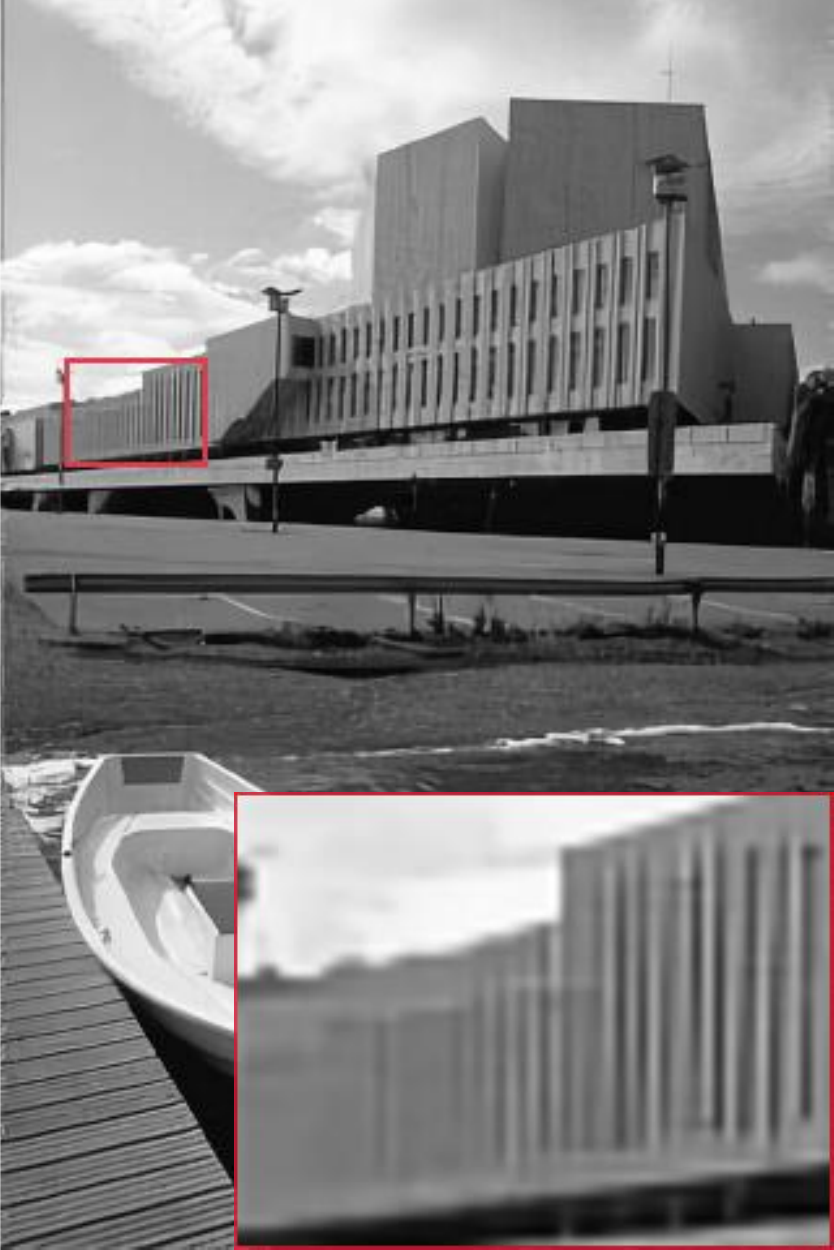}}
        \end{minipage}
        \begin{minipage}[b]{0.18\linewidth}
            \subfloat[Halftone]{\includegraphics[width=1\linewidth]{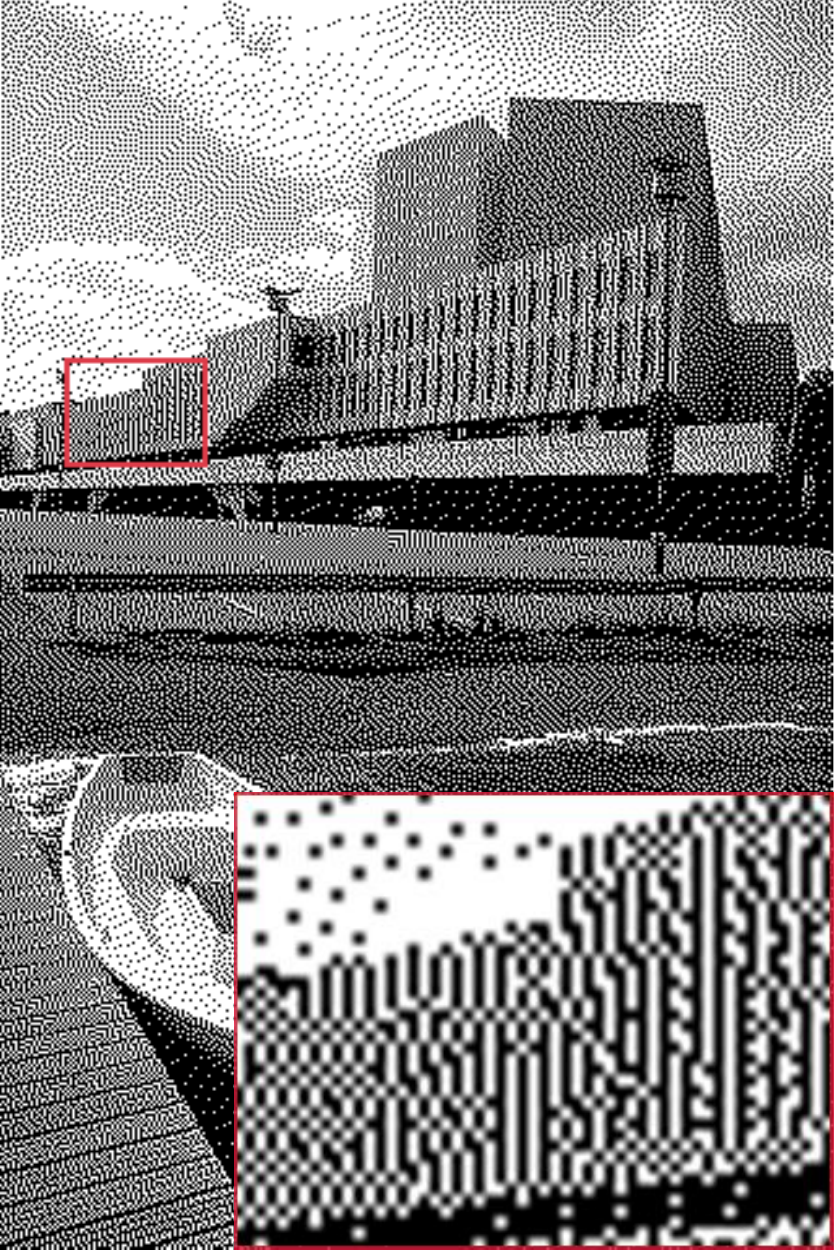}} \vspace{-0.9em}
            \subfloat[GGRL]{\includegraphics[width=1\linewidth]{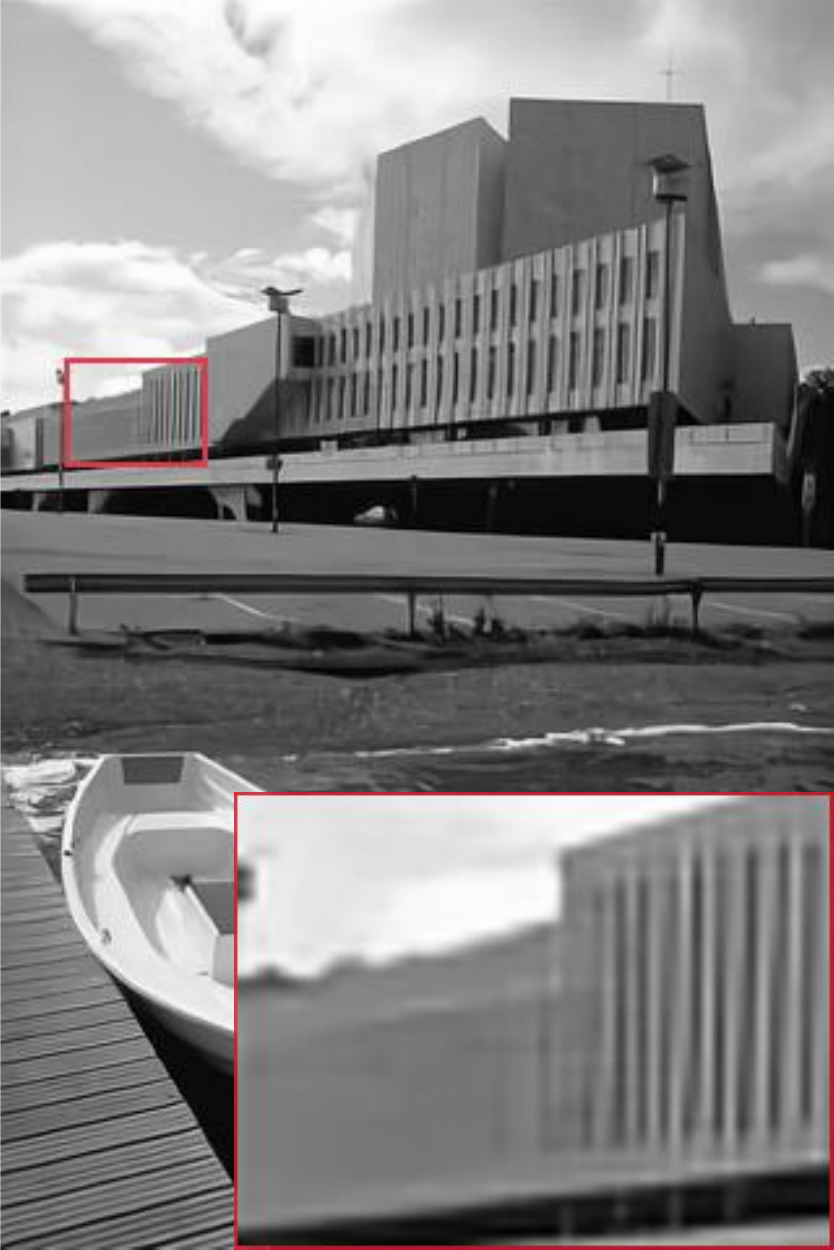}}
        \end{minipage}
        \begin{minipage}[b]{0.18\linewidth}
            \subfloat[DnCNN]{\includegraphics[width=1\linewidth]{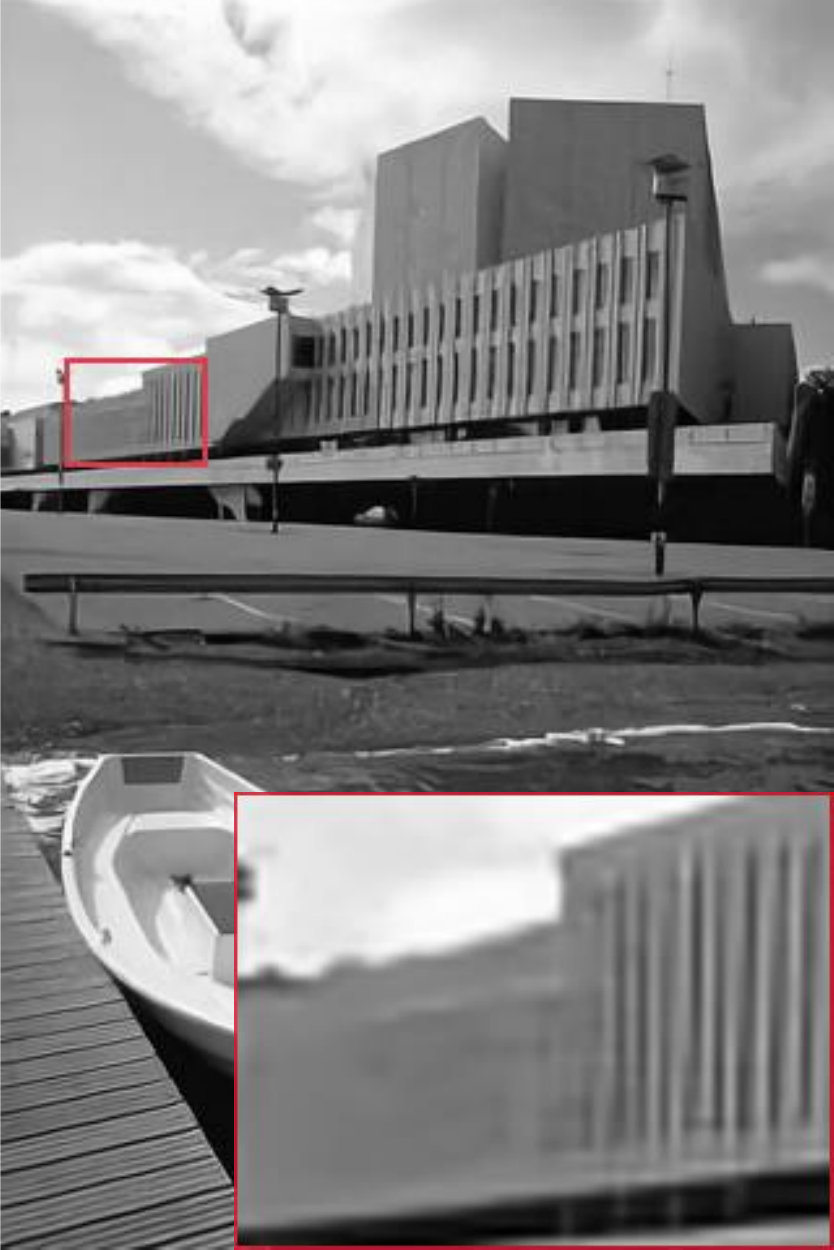}} \vspace{-0.9em}
            \subfloat[MIMOUNet]{\includegraphics[width=1\linewidth]{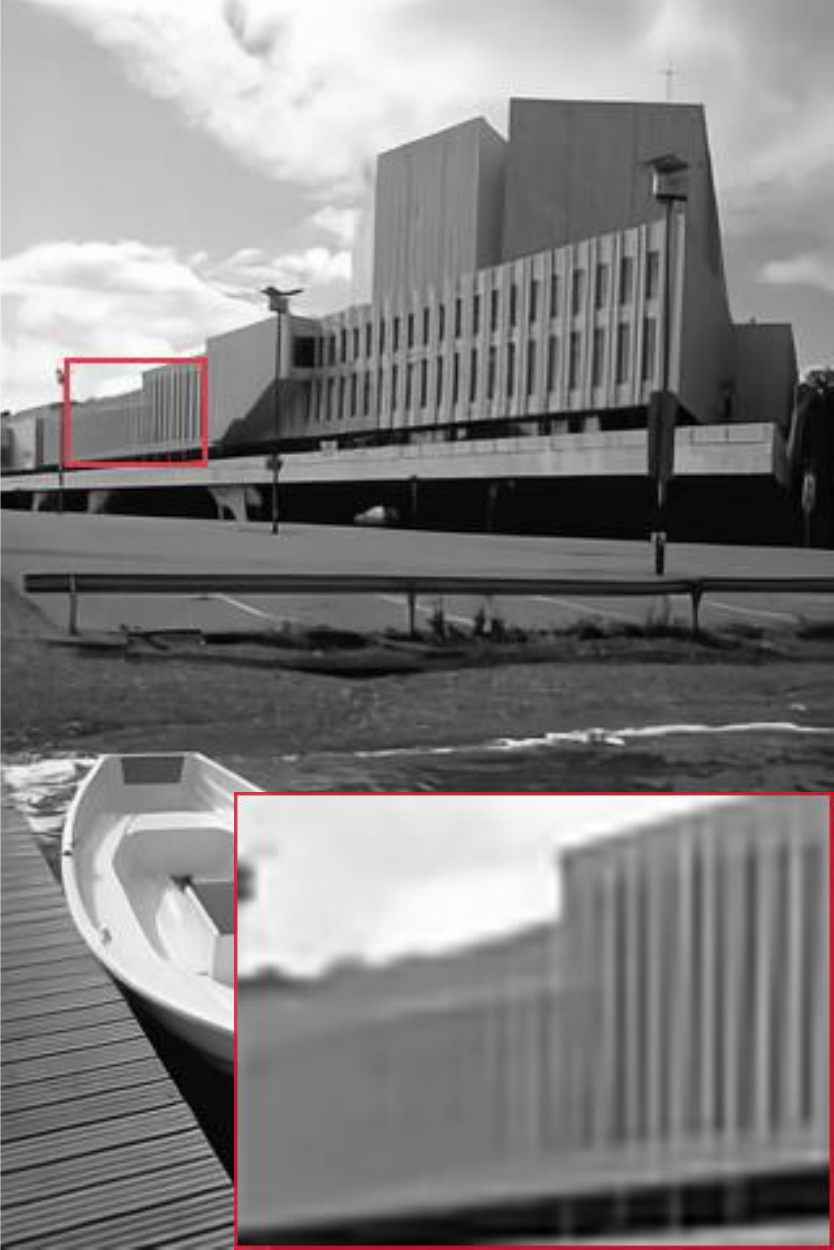}}
        \end{minipage}
        \begin{minipage}[b]{0.18\linewidth}
            \subfloat[VDSR]{\includegraphics[width=1\linewidth]{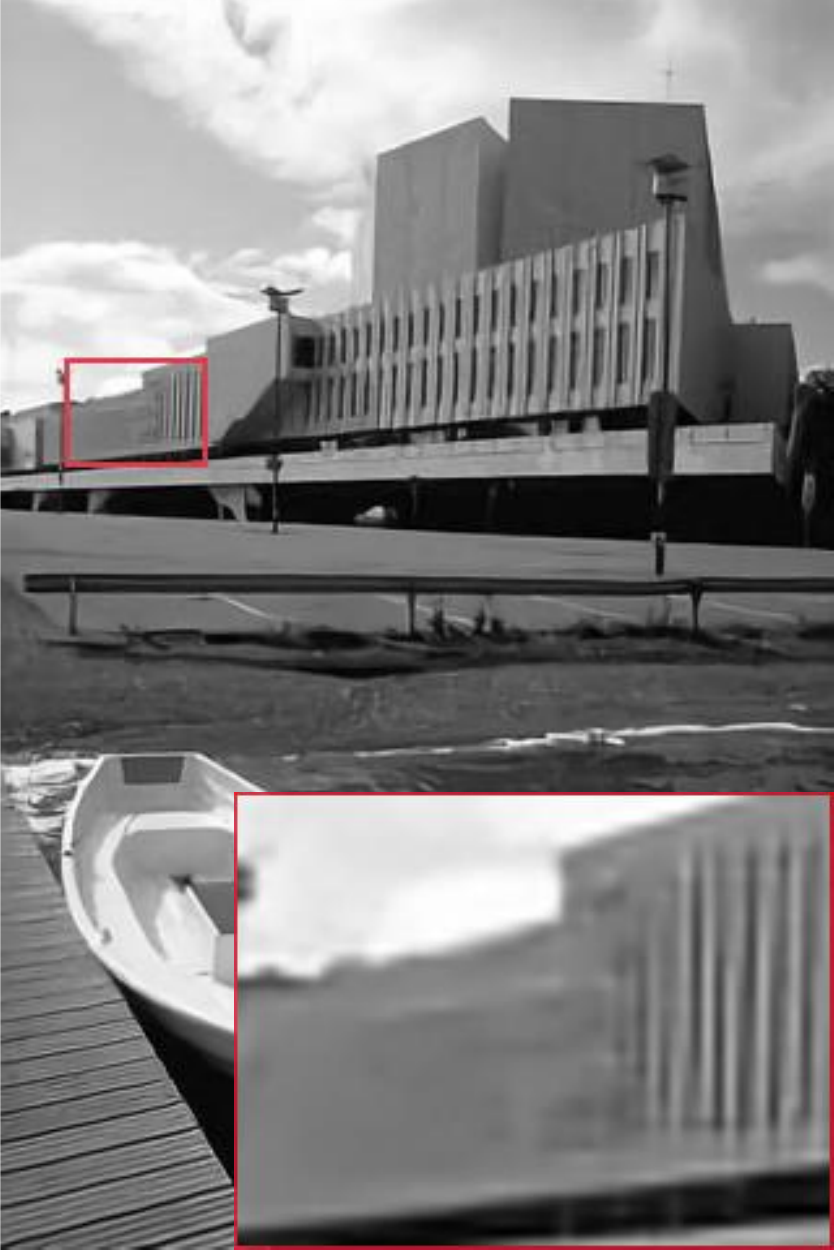}} \vspace{-0.9em}
            \subfloat[PRL-dt (ours)]{\includegraphics[width=1\linewidth]{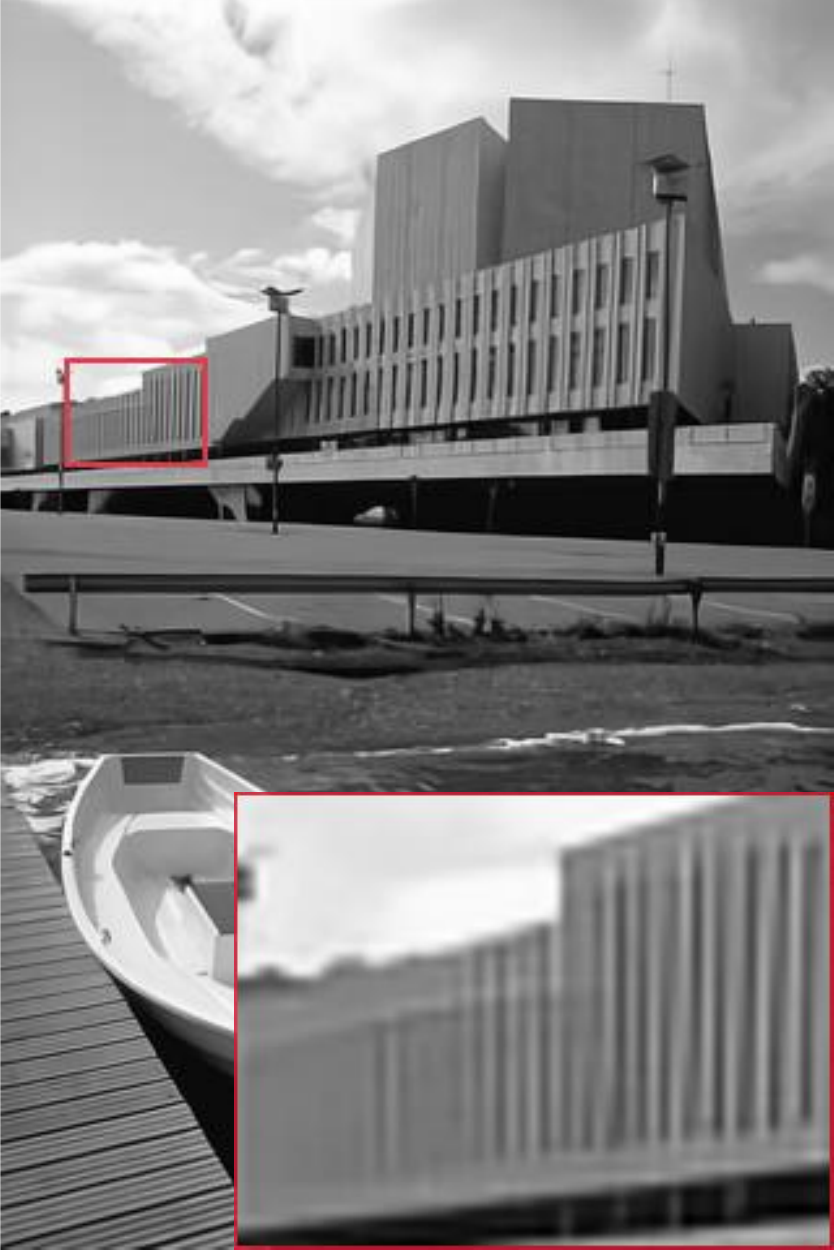}}
        \end{minipage}
        \begin{minipage}[b]{0.18\linewidth}
            \subfloat[EDSR]{\includegraphics[width=1\linewidth]{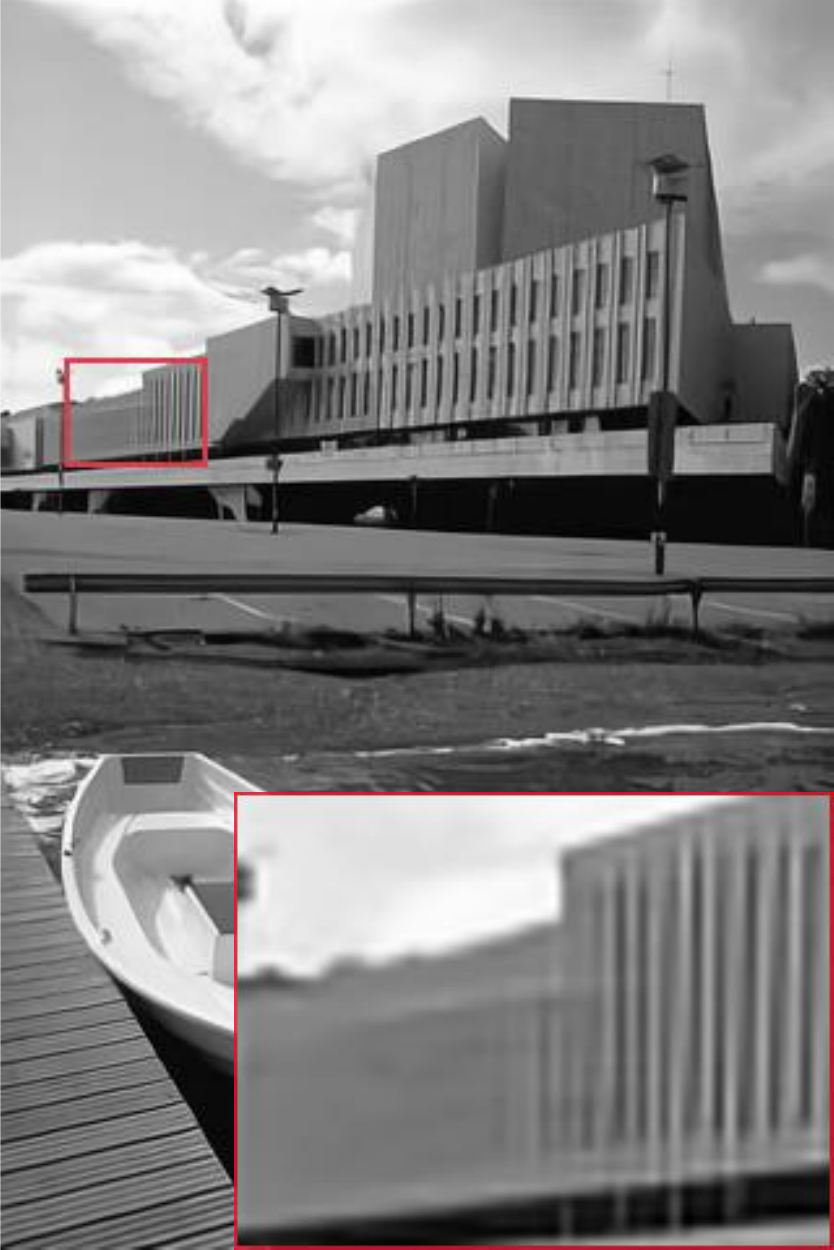}} \vspace{-0.9em}
            \subfloat[MSPRL (ours)]{\includegraphics[width=1\linewidth]{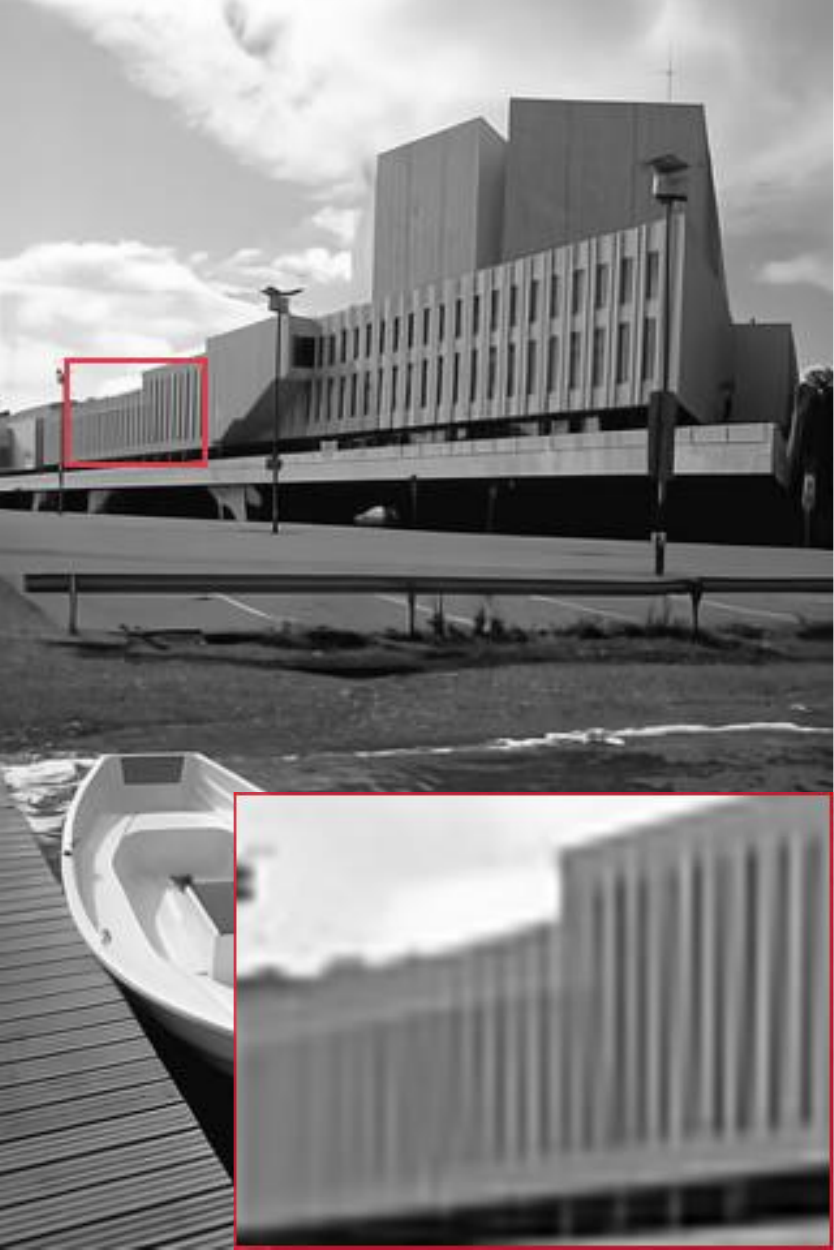}}
        \end{minipage}
        \end{center}
        \vspace{0.5em}
        \caption{Compared with the other approaches, our MSPRL more effectively restores the image details.}
    \label{fig:c6}
\end{figure*}

\begin{figure*}[ht]
\begin{center}
    \captionsetup[subfloat]{labelsep=none,format=plain,labelformat=empty,font={scriptsize}}
        % 86000
        \begin{minipage}[b]{0.18\linewidth}
            \subfloat[Ground Truth]{\includegraphics[width=1\linewidth]{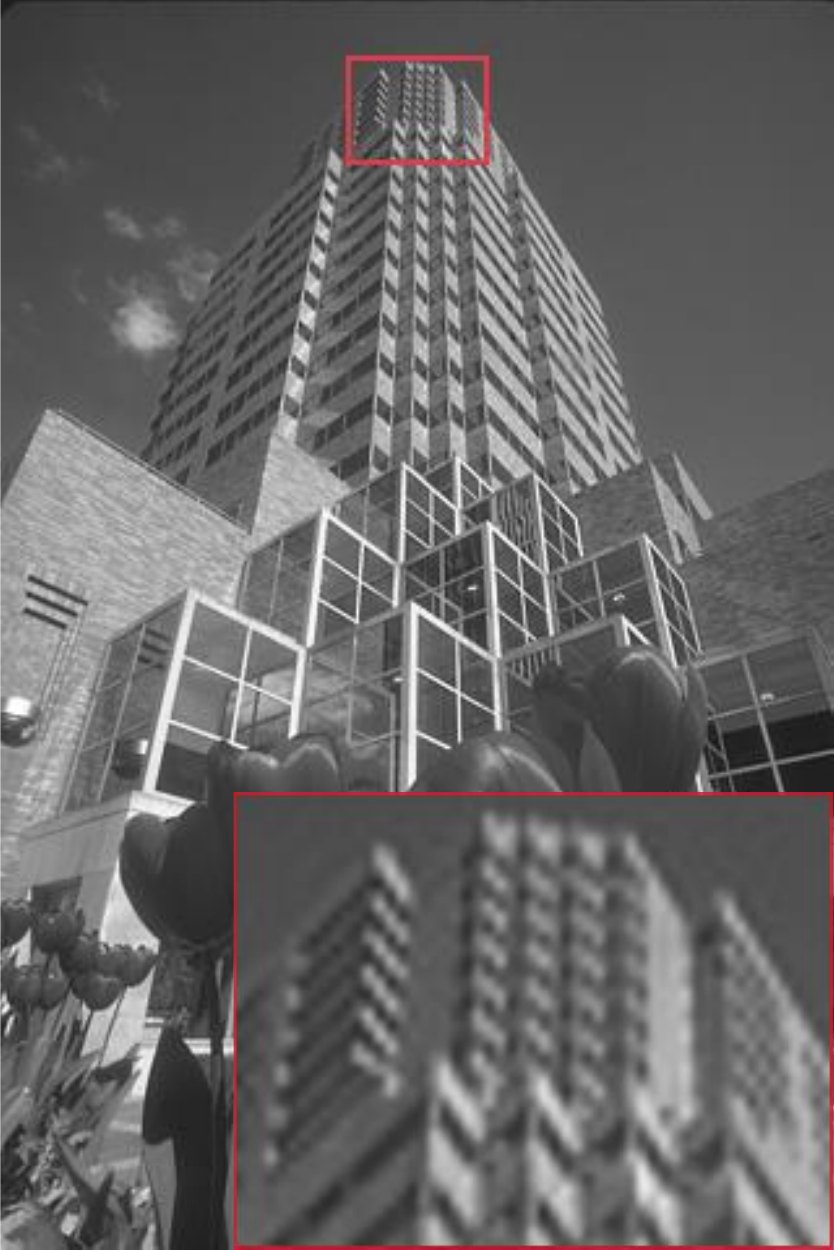}} \vspace{-0.9em}
            \subfloat[PRL]{\includegraphics[width=1\linewidth]{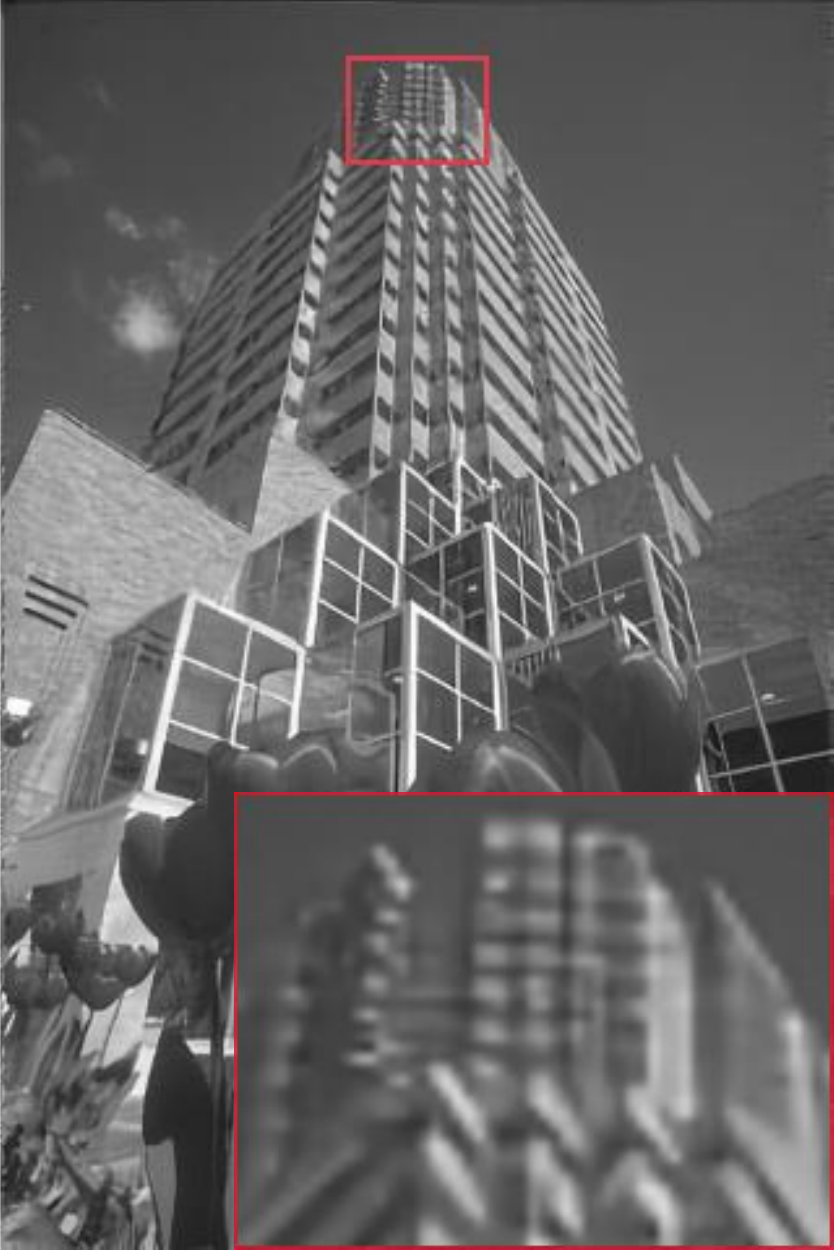}}
        \end{minipage}
        \begin{minipage}[b]{0.18\linewidth}
            \subfloat[Halftone]{\includegraphics[width=1\linewidth]{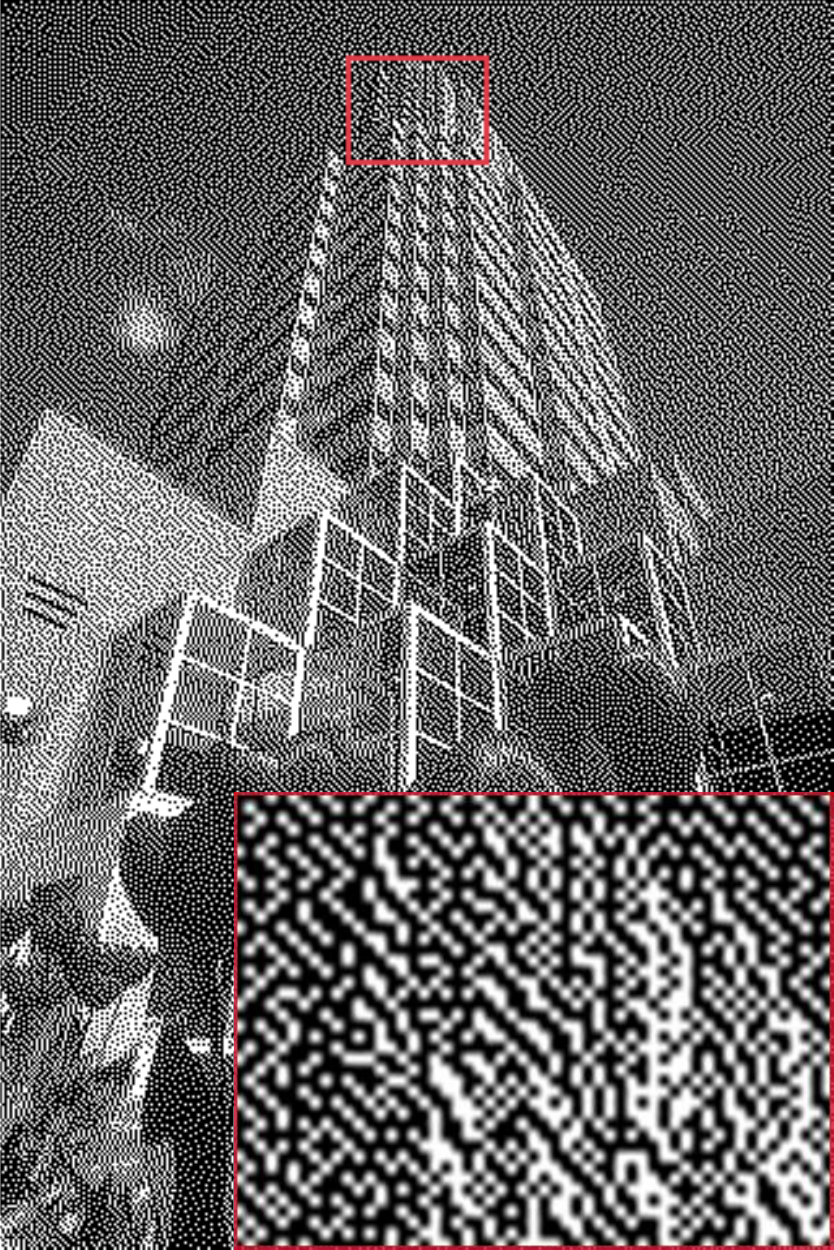}} \vspace{-0.9em}
            \subfloat[GGRL]{\includegraphics[width=1\linewidth]{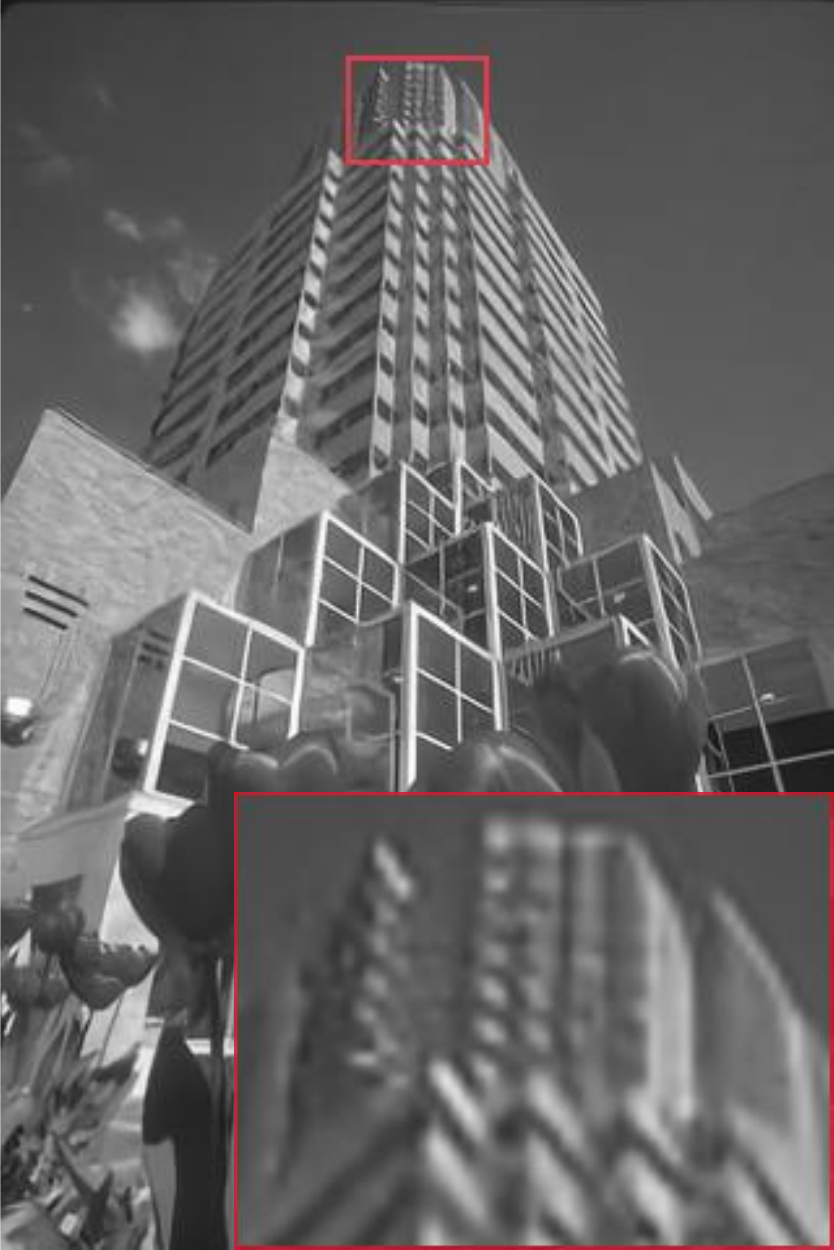}}
        \end{minipage}
        \begin{minipage}[b]{0.18\linewidth}
            \subfloat[DnCNN]{\includegraphics[width=1\linewidth]{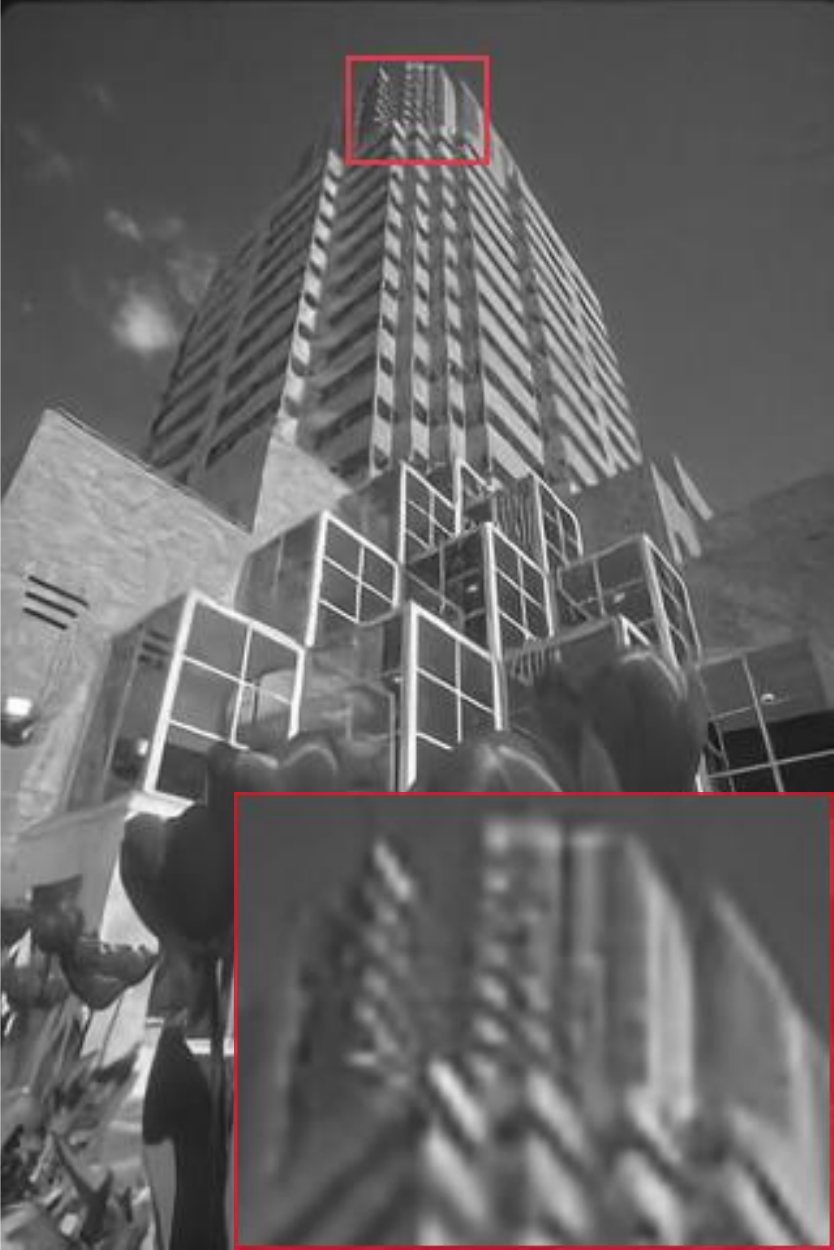}} \vspace{-0.9em}
            \subfloat[MIMOUNet]{\includegraphics[width=1\linewidth]{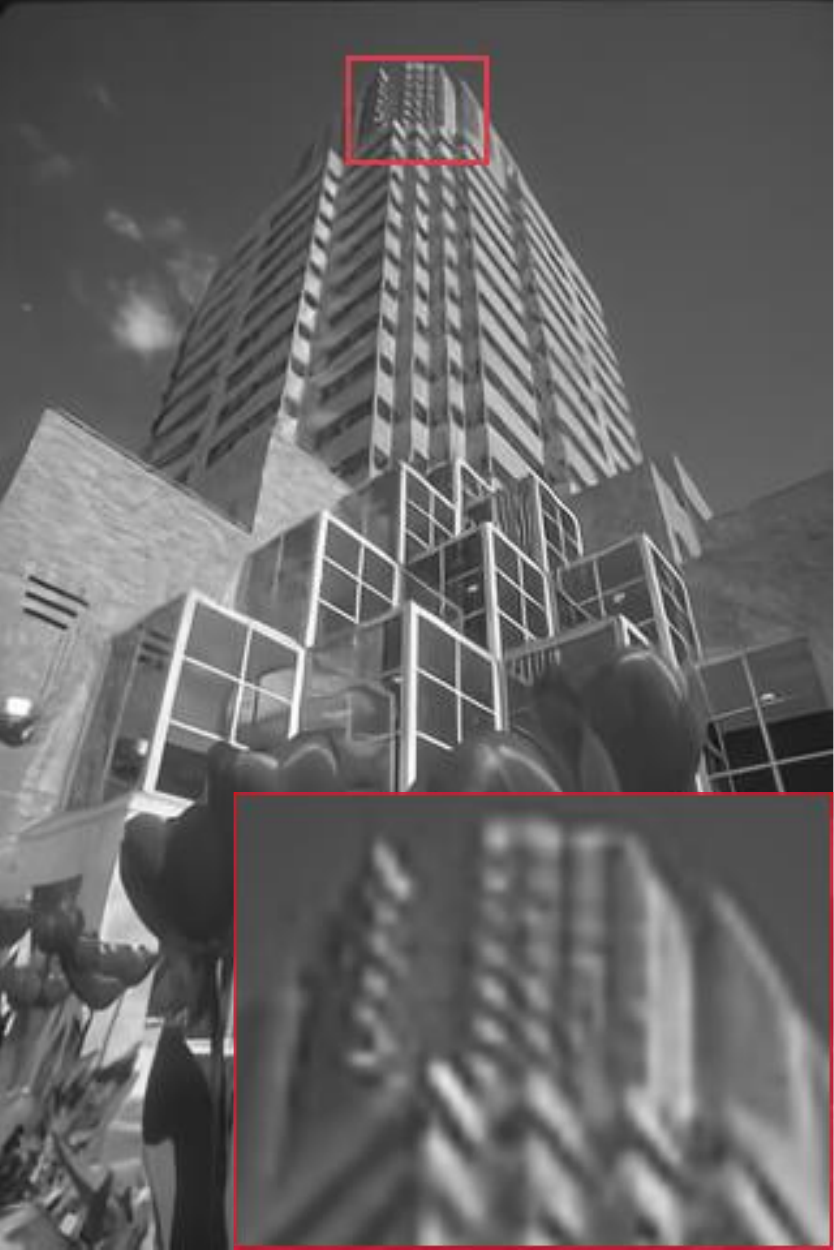}}
        \end{minipage}
        \begin{minipage}[b]{0.18\linewidth}
            \subfloat[VDSR]{\includegraphics[width=1\linewidth]{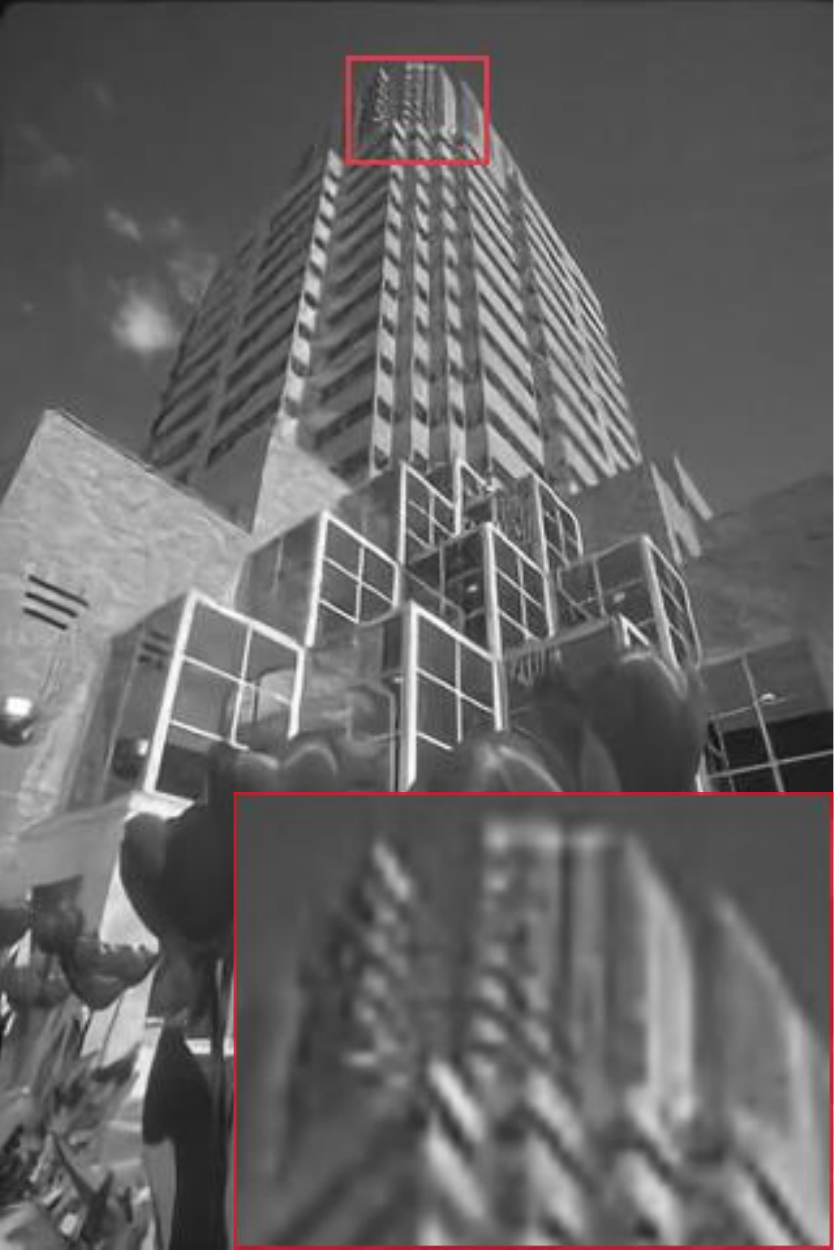}} \vspace{-0.9em}
            \subfloat[PRL-dt (ours)]{\includegraphics[width=1\linewidth]{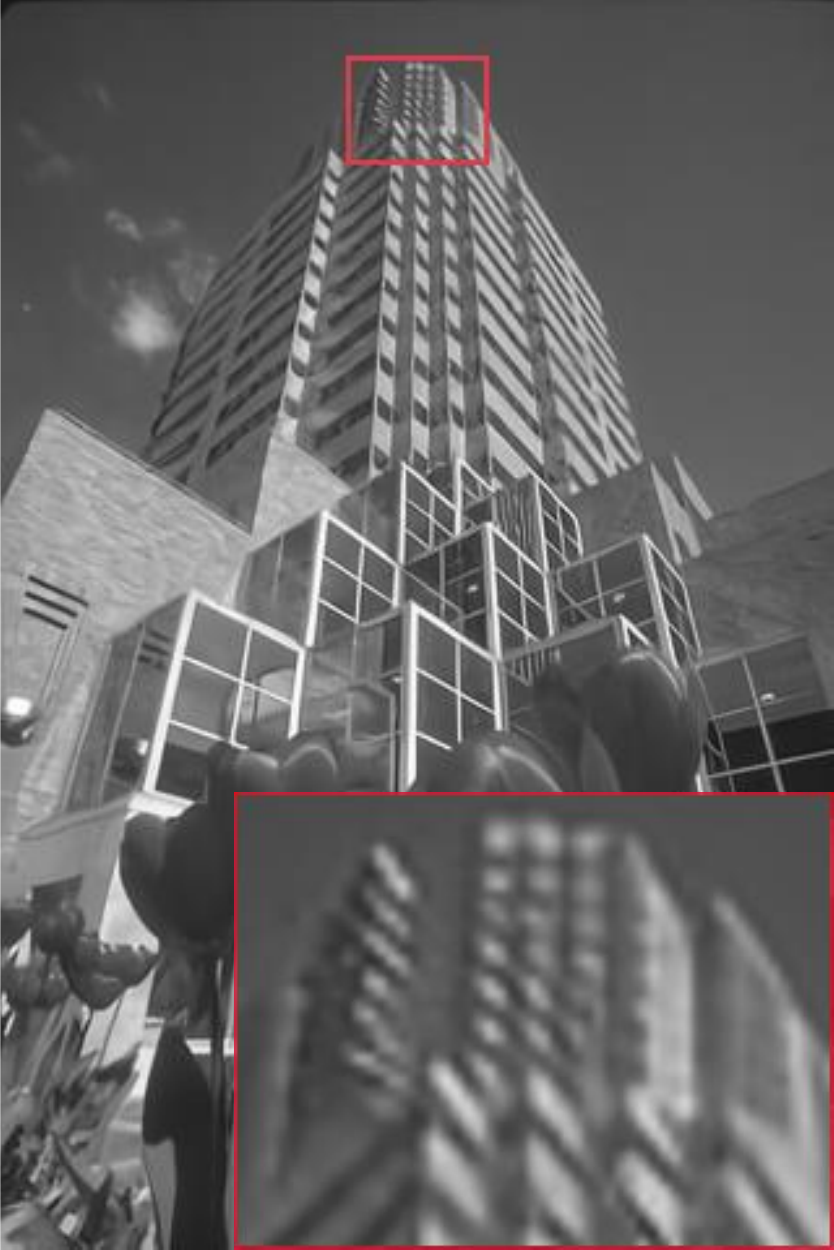}}
        \end{minipage}
        \begin{minipage}[b]{0.18\linewidth}
            \subfloat[EDSR]{\includegraphics[width=1\linewidth]{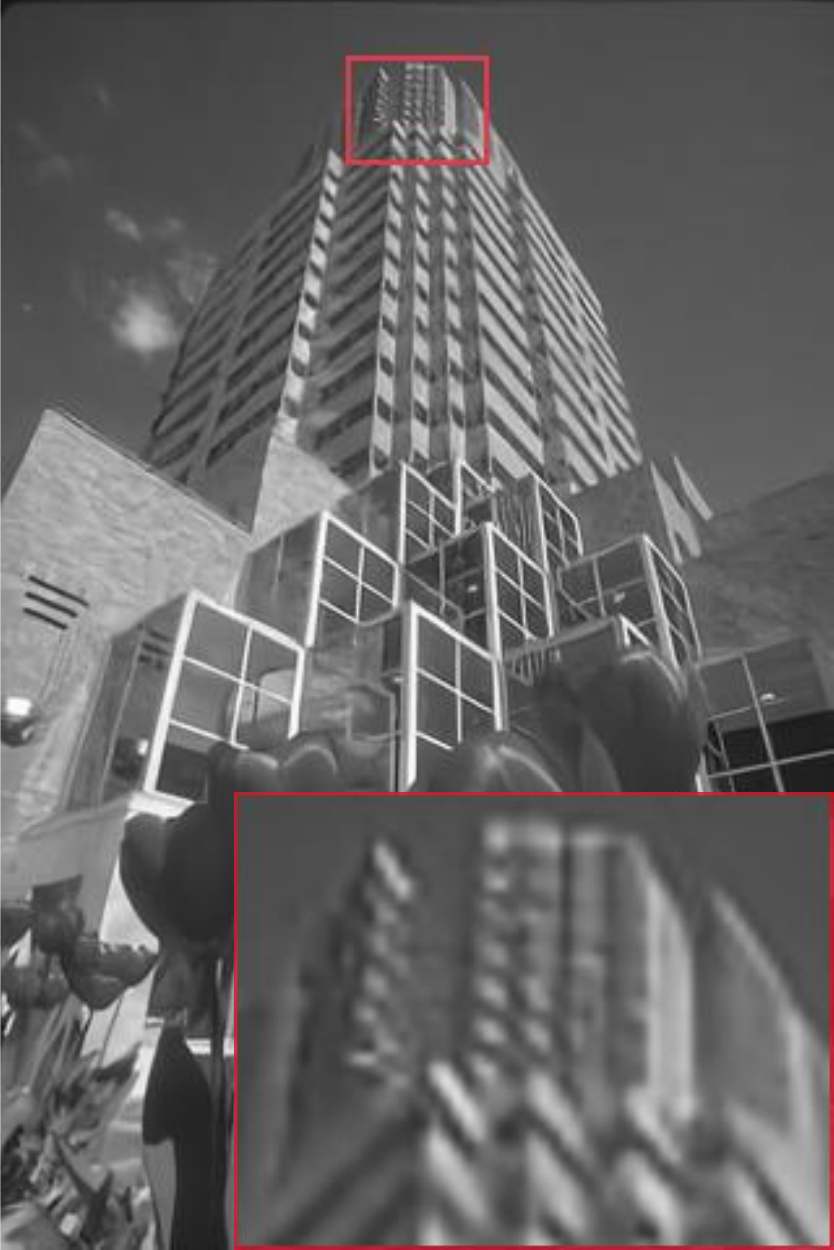}} \vspace{-0.9em}
            \subfloat[MSPRL (ours)]{\includegraphics[width=1\linewidth]{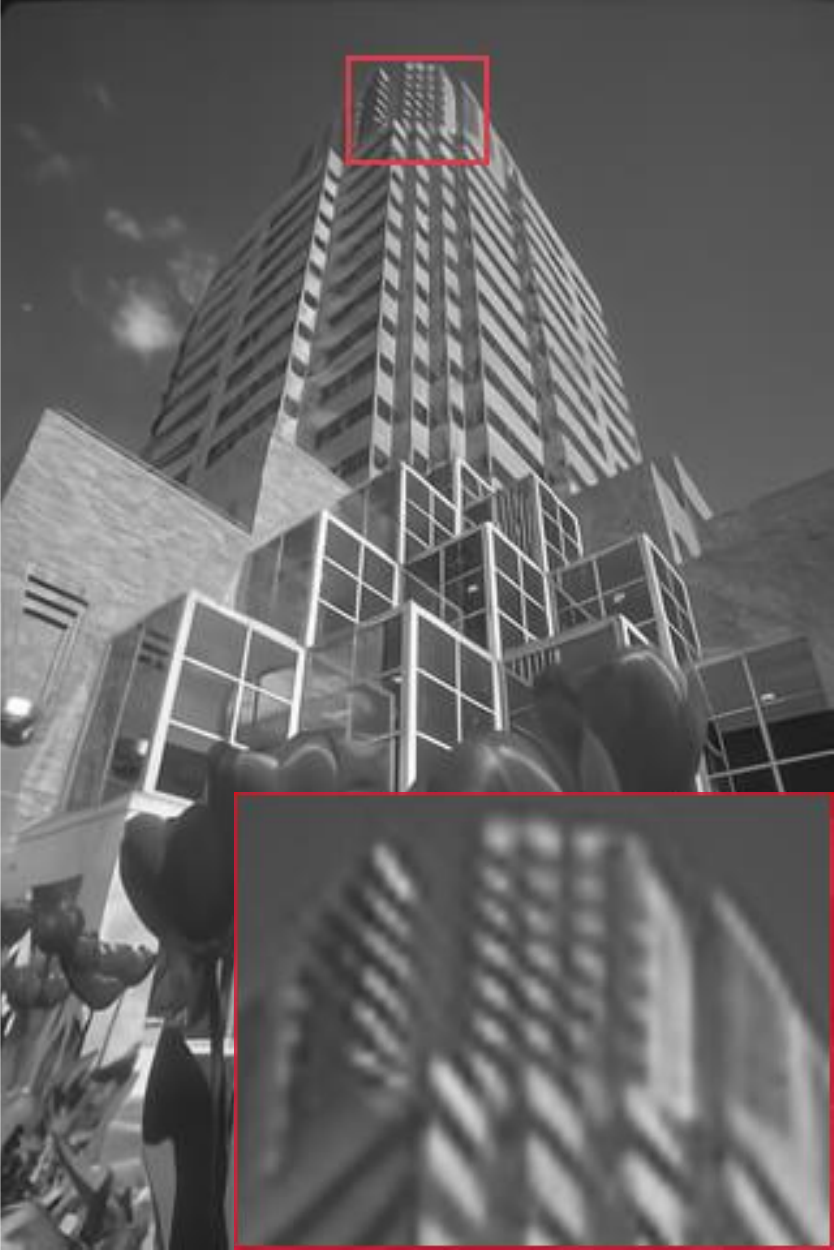}}
        \end{minipage}
        \end{center}
        \vspace{0.5em}
        \caption{Compared with the other approaches, our MSPRL more effectively restores the image details.}
    \label{fig:c7}
\end{figure*}

We show the visual comparisons in \figref{fig:c1}. Our MSPRL can obtain more obvious texture and structure information than PRL-dt, which effectively restores the image details. In the Lena image, the hat texture can be well restored using MSPRL, which is closer to the original real image. And, in the Barbara image, the cloth texture restoration of the other models shows more bending phenomena. In \figref{fig:c2} (Row 2, 3 and 4), other models cannot restore the dense circle and point shape of the image, showing the line shape in different directions, while MSPRL can avoid this problem and reconstruct the image. Compared with other models, although the netted information loss of halftone image is very serious, MSPRL is still able to recover the main details, as shown in \figref{fig:c3}. In addition, the restoration visual effects of MSPRL in the architectures, letters and lines are more smooth and refined, which are shown in \figref{fig:c4}, \figref{fig:c5}, \figref{fig:c6} and \figref{fig:c7}. Lastly, we also compare the restoration performance on the classic images shown in \tabref{tab:6}.

\begin{table}[ht]
% \vspace{-1em}
\footnotesize
\renewcommand\arraystretch{1}
\begin{center}
\caption{Performance comparison of different inverse halftoning methods on some classic images with size of 512 (PSNR). The best results are marked in \textbf{Bold}.}
% \vspace{-0.5em}
\label{tab:6}
\resizebox{\linewidth}{!}{
    \begin{tabular}{lcccccccc}
    \toprule
     Model & DnCNN
     & VDSR & EDSR
     & PRL & GGRL
     & MIMOUNet & PRL-dt
     & MSPRL
    \\
     \midrule
     Baboon & 24.73 & 24.59 & 24.85 & 24.50 & 24.83 & 24.98 & 25.03 & \textbf{25.12} \\
     Barbara & 29.35 & 28.08 & 29.95 & 29.44 & 30.19 & 30.58 & 30.79 & \textbf{31.59} \\
     Boat & 31.77 & 31.54 & 31.95 & 31.21 & 31.92 & 32.00 & 32.14 & \textbf{32.25} \\
     Couple & 31.55 & 31.36 & 31.79 & 30.91 & 31.77 & 31.87 & 31.95 & \textbf{32.07} \\
     Goldhill & 31.71 & 31.51 & 31.86 & 31.01 & 31.87 & 31.90 & 32.06 & \textbf{32.15} \\
     House & 38.90 & 38.55 & 39.38 & 36.21 & 39.39 & 39.42 & 39.75 & \textbf{39.95} \\
     Lena  & 34.51 & 34.32 & 34.78 & 33.34 & 34.77 & 34.84 & 35.00 & \textbf{35.09} \\
     Man & 31.86 & 31.68 & 31.97 & 30.96 & 31.97 & 32.00 & 32.08 & \textbf{32.15} \\
     Peppers & 34.32 & 34.09 & 34.42 & 33.11 & 34.39 & 34.43 & 34.49 & \textbf{34.55} \\
     \botrule
    \end{tabular}
}
\end{center}
\vspace{-2em}
\end{table}

\section{Conclusion}
In this paper, we present a multiscale progressively residual learning architecture network (MSPRL) for the inverse halftoning task. The encoder restores content information from different scale images and the decoder collects encoder features to extract deep features. The feature maps of the entire model are progressively learned. Our MSPRL is a simple and efficient model that can learn information of different scale images. In addition, we use suitable training strategies compared with many previous CNN-based inverse halftoning methods. We also explored the performance of the model between different settings and feature blocks. The experimental results demonstrate that our method outperforms the other model methods. Recently, many researchers have added colorization tasks to inverse halftoning; we will follow up research to restore better visual perception of color continuous-tone images in the future.

\backmatter

% \section*{Declarations}

% \begin{itemize}
% \item Funding: This work was supported in part by the Zhejiang Public Welfare Technology Research Project Fund of China under Grant LGG22F020021, and the City Public Welfare Technology Application Research Project of Jiaxing Science and Technology Bureau of China under Grant 2021AY10071.
% \item Conflict of interest: The authors declare that they have no conflict of interest regarding the publication of this article.
% \item Ethics approval: This article does not contain any studies with human participants/animals performed by any of the authors.
% \item Consent to participate: Not applicable.
% \item Consent for publication: The work has not been submitted elsewhere for publication, in whole or in part, and all the authors have approved the publication.
% \item Availability of data and materials: Not applicable.
% \item Code availability: Our codes are available at \url{https://github.com/FeiyuLi-cs/MSPRL}.
% \item Authors' contributions: \textbf{Feiyu Li}: Writing original draft, Conceptualization, Methodology,  Formal analysis. \textbf{Jun Yang}: Supervision, Writing review and editing, Investigation, Funding acquisition.
% \end{itemize}

\bibliography{sn-article}% common bib file

\end{document}